\documentclass[11pt]{article}

\usepackage[T1]{fontenc}
\usepackage{lmodern}
\usepackage[letterpaper, top=0.85in, bottom=0.85in, left=0.75in, right=0.75in]{geometry}
\usepackage{amsmath}
\usepackage{microtype}
\usepackage{graphicx}
\usepackage{array}
\usepackage{booktabs}
\usepackage{longtable}
\usepackage{caption}
\usepackage{subcaption}
\usepackage{float}
\usepackage{natbib}
\usepackage{xcolor}
\definecolor{fturlblue}{HTML}{1966FF}
\usepackage{hyperref}
\usepackage{tikz}
\usepackage{listings}
\lstnewenvironment{promptverbatim}{%
  \lstset{%
    basicstyle=\ttfamily\scriptsize,%
    breaklines=true,%
    breakatwhitespace=false,%
    columns=fullflexible,%
    keepspaces=true,%
    frame=none,%
    rulesep=0pt,%
    tabsize=2,%
    showtabs=false,%
    upquote=true,%
    aboveskip=0.6em,%
    belowskip=0.6em,%
  }%
}{}

\hypersetup{
  colorlinks=true,
  linkcolor=black,
  citecolor=black,
  urlcolor=fturlblue,
}

\graphicspath{{figures/}}

\title{\textbf{How exposed is work to robotics? Occupational, industry, and geographic measures}}

\author{%
  \begin{tabular}[t]{@{}c@{\hspace{2.5em}}c@{\hspace{2.5em}}c@{}}
    \multicolumn{3}{c}{\large Working Paper} \\[1.2em]
    {\large Sebastian Sartor$^{1,3}$} &
    {\large Robert Seamans$^{2,3}$} &
    {\large Neil C. Thompson$^{1,3}$} \\[0.65em]
    {\small \texttt{sartor97@mit.edu}} &
    {\small \texttt{rseamans@stern.nyu.edu}} &
    {\small \texttt{neil\_t@mit.edu}} \\[0.65em]
    \multicolumn{3}{c}{\small $^{1}$MIT \qquad $^{2}$NYU \qquad $^{3}$MIT FutureTech}
  \end{tabular}%
}

\date{}

\begin{document}

\maketitle

\begin{abstract}
Exposure indices that map technical capabilities onto occupational abilities have become a standard tool for anticipating how AI may reshape work, but existing measures either cover software AI or proxy robotics through patents and installation data that reflect the industrial robots of past decades.
We construct the first abilities-based exposure measures for robotics and physical AI, and the first to span occupations, industries, and geographies.
Following the framework of \citet{felten2018method,felten2021occupational}, we build a robotics capability taxonomy of 16 capabilities in eight domains from 78 academic, industry, and standards sources, use a large language model to score the relatedness of each capability to the 52 O{*}NET work abilities, and aggregate to Robotics Occupational Exposure (ROE) for 774 occupations, Robotics Industry Exposure (RIE) for 250 NAICS industries, and county-level Robotics Geographic Exposure (RGE).
We find that robotics exposure concentrates in physically intensive work such as construction, farming, transportation, and production, and in industries such as logging, mining, trucking, and warehousing.
Unlike generative-AI exposure, ROE is higher in occupations with lower wages, lower education requirements, and less creative-ability content.
Physical collaboration and assistance, terrain-based locomotion, grasping and handling, and task and functional reasoning move the exposure ranking most, yet no single capability carries the index.
Because the measures capture technical relatedness rather than readiness, cost, or adoption, they should be read as an upper bound on where physical automation could affect work.
\end{abstract}

\vspace{0.75em}

\section{Introduction}

Exposure studies have become a standard tool for understanding the economic and societal implications of AI. They map technical capabilities onto occupational tasks and identify which tasks, occupations, industries, and geographic areas are likely to change the most, and in what way. This is particularly useful for disruptive technologies because they provide predictive insight on the potential implications. Besides being a central foundation for economic studies done by researchers, policy makers rely on them for policy measures around automation and augmentation, workforce development, and upskilling.

A growing literature measures exposure to AI. \citet{frey2017future} estimate the probability of computerisation for US occupations, \citet{brynjolfsson2018can} score tasks on their suitability for machine learning, and \citet{felten2021occupational} build an occupation-level AI exposure index using a method that links AI capabilities to O{*}NET abilities \citet{felten2018method}. More recent work turns to large language models \citep{eloundou2024gpts}.

Robots appear in this literature in two limited forms. \citet{webb2019impact} constructs a robot exposure score from patents, which reflects the robotics of past decades: programmed machines for structured, repetitive tasks, mostly in manufacturing. A second strand studies the effects of robot adoption using installation data, which measures where robots already are rather than where they could go \citep{graetz2015robots,acemoglu2020robots}. Concurrent with this work, \citet{legateyang2026robots} score whether robots demonstrated today can perform each physical O{*}NET task and aggregate those task ratings to an occupation-level index. That measure is anchored to the currently demonstrated frontier, is defined for occupations only, and does not decompose into the underlying robot capabilities. Each of these measures is anchored to robots that already exist: patents filed, machines installed, or demonstrations recorded. None describes exposure to the capabilities that current work on physical AI aims to deliver, and none extends beyond occupations to industries and places.

Robots are not new. The word “robot” is about 35 years older than the term “artificial intelligence” and robots have performed a wide range of tasks across industries for decades. What has changed is the promise of physical AI and its ability to generalize across tasks, objects, and environments. If this generalization materializes, it widens the set of tasks that can be automated, and it also lowers the barriers that limit adoption today. Deployment cost and integration time remain major barriers relative to hardware cost \citep{berger2022puzzle}, as does the skill required to program and maintain a system. Physical AI promises to reduce all three.

Physical AI is also not tied to one form factor. Robot foundation models are trained to transfer across embodiments, including humanoids \citep{o2024open}. Humanoids have existed for decades, mainly in research settings and more recently for entertainment purposes. Yet, they have yet to demonstrate the capability to contribute with meaningful manipulation tasks. Recent investment in humanoid platforms, large-scale robot learning datasets, and robot foundation models suggest that manipulation capabilities may improve substantially in short time.

These developments matter for the existing exposure picture. LLM exposure concentrates in cognitive, often higher-wage work and physical tasks are largely outside these measures by construction. If new embodiments and foundation-model control have large effects on labor and industries, it is open which occupations and sectors are most exposed.

Therefore, this paper provides the first abilities-based estimate of exposure to robotics, and the first that spans occupations, industries, and geographies. We apply the same abilities-based pipeline: define a robot domain and capability taxonomy, score how each domain and capability relates to work abilities, and aggregate to exposure indices for occupations (ROE), industries (RIE), and geographies (RGE). We then describe how occupational exposure correlates with median wages, required education, creative-ability requirements, and demographic composition. Next we compare these patterns to published generative-AI exposure measures \citep{felten2024generative}. We also study which capabilities drive the exposure score the most. We find that physical collaboration and assistance, terrain-based locomotion, grasping and handling, and task and functional reasoning matter most, but dropping any one of them leaves the occupation ranking largely intact. This changes on the domain level. There Manipulation and Locomotion \& Mobility dominate under both aggregation rules we consider, and manipulation alone contributes 2.5 times what its four capabilities contribute separately. The results are meant to inform both academic researchers and policy makers about where physical automation could affect work.

Most importantly, our measure does not account for current hardware and model capabilities, the economic attractiveness of automation (e.g., as studied by \citet{svanberg2024beyond} for computer vision), or regulatory and safety constraints. It describes what could be automated if physical AI delivers on its promise of generalization, and should be read as an upper bound. Previous work shows that technical feasibility and adoption diverge substantially: \citet{mertens2026crashing} show that models could complete about 50\% of LLM-accessible tasks at least at average human quality, while \citet{yu2026ai} find that AI is deeply integrated in only 11\% of S\&P~500 firms. \citet{svanberg2024beyond} find that only a minority of vision tasks exposed to AI would be cost-effective to automate at current costs.

The remainder of the paper is organized as follows. Section~\ref{sec:method} presents the method with a focus on the relation to prior AI exposure indices, the robotics capability taxonomy, the LLM-scored relatedness matrix, exposure index formulas for ROE, RIE, and RGE, and labor-market correlates. Section~\ref{sec:results} reports occupational, industry, and geographic exposure levels; descriptive correlates with wages, education, creative ability, and demographic composition; and a comparison of robotics exposure to generative-AI exposure on rankings and correlates, and to observed LLM task success. It closes with the capability decomposition. Sections~\ref{sec:discussion} and~\ref{sec:conclusion} discuss interpretation, limitations, and extensions, and conclude. The appendix provides taxonomy sources and prompts, relatedness-matrix evaluation details, and validation of the LLM scores.

\section{Method}
\label{sec:method}

\subsection{Overview}
\label{sec:method-overview}

We follow the methodology of \citet{felten2018method} and \citet{felten2021occupational} to estimate exposure to robotics. At a high level, the methodology links (i) technology domain areas with (ii) occupational work abilities using a matrix for pairwise correlation estimation. \citet{felten2021occupational} begin from a pre-defined list of AI application domains from the Electronic Frontier Foundation (EFF), take the relevant work abilities from O{*}NET, use human input to score how each domain relates to work abilities and aggregate the results to estimate exposure indices for occupations, industries, and places.

Two key differences apply to our methodology. First, we construct our own robotics capability taxonomy, built inductively from literature and standards, to better capture the breadth of robotics and physical AI. Second, we score the relatedness between each capability and each O{*}NET ability using an LLM rather than a human survey. LLMs have become an established tool in economics research because of their cost effectiveness and ability to scale \citep{eloundou2024gpts,mertens2026crashing}. We validate the LLM scores against a human survey in the appendix.

\subsection{Robotics capability taxonomy}
\label{sec:method-taxonomy}

Prior AI exposure work begins from a fixed list of application domains; for robotics we derive the domains from the source literature instead. We use Google Gemini to run a two-stage LLM pipeline on 78 academic, industry, and government or standards documents. We first process each pdf in isolation and then condensate the output of the combined extractions into a structured capability hierarchy.

We assembled sources in three passes (initial coverage of major robot classes, gap analysis by subfield, and targeted additions). Extraction kept only explicitly stated categories and preserved source wording. The condensation merged redundant terms into 16 capabilities grouped into eight top-level domains (Table~\ref{tab:capabilities}); the 16 capabilities are the rows of the relatedness matrix in Section~\ref{sec:method-matrix}. Capabilities play the same role as the published AI application list for linking technology to work, but are grounded in robotics-specific sources rather than software-AI benchmarks.\footnote{Source list and extraction and condensation prompts are in Appendix~\ref{app:taxonomy-sources}.}

\newcommand{\taxonomydomainrule}{%
  \addlinespace[0.25em]%
  \noalign{\vskip 0.15em}%
  \noalign{\hfil\tikz{\draw[dashed,line width=0.35pt] (0,0) -- (0.935\linewidth,0);}\hfil}%
  \noalign{\vskip 0.15em}%
}

\begin{table}[H]
  \centering
  \small
  \caption{Robotics capability taxonomy: 16 capabilities grouped into eight domains.}
  \label{tab:capabilities}
  \begin{tabular}{@{}>{\raggedright\arraybackslash}p{0.17\linewidth}>{\raggedright\arraybackslash}p{0.26\linewidth}>{\raggedright\arraybackslash}p{0.51\linewidth}@{}}
    \toprule
    Domain & Capability & Definition \\
    \midrule
    Manipulation & Grasping \& Handling &
    Direct contact, containment, and secure physical engagement with objects. \\
    & Dexterous \& In-Hand Manipulation &
    High-degree-of-freedom interactions including reorientation and fine motor control. \\
    & Tool Use \& Process Interaction &
    Applying mechanical operations, force-controlled tasks, or specialized material processing. \\
    & Complex Object Manipulation &
    Interacting with non-rigid, soft, or high-uncertainty environments. \\
    \taxonomydomainrule
    Locomotion \& Mobility & Terrain-Based Locomotion &
    Ground-based movement using various physical mechanisms. \\
    & Aerial \& Aquatic Locomotion &
    Movement through non-terrestrial mediums. \\
    \taxonomydomainrule
    Navigation \& Spatial Awareness & Mapping \& Localization &
    Determining self-position and modeling the geometry of the environment. \\
    & Path \& Trajectory Planning &
    Calculating, tracking, and optimizing routes through space. \\
    \taxonomydomainrule
    Perception \& Sensing & Visual \& Environmental Perception &
    Detecting and identifying objects, features, or events using cameras and environmental sensors. \\
    & Tactile \& Force Sensing &
    Interpreting physical contact, material properties, or forces. \\
    \taxonomydomainrule
    Cognition \& Planning & Task \& Functional Reasoning &
    High-level cognitive processing, commonsense, and logic. \\
    \taxonomydomainrule
    Human-Robot Interaction (HRI) & Expressive \& Communicative Interaction &
    Social, verbal, or non-verbal engagement. \\
    & Physical Collaboration \& Assistance &
    Direct support, guidance, or shared physical tasks with humans. \\
    \taxonomydomainrule
    Multi-Agent \& Collective Behavior & Collective Coordination &
    Group-based operations, swarming, and decentralized decision-making. \\
    \taxonomydomainrule
    System Maintenance \& Autonomy & System Health \& Self-Management &
    Internal monitoring, repair, and configuration. \\
    & Autonomy \& Mission Execution &
    Levels of operational independence and task completion. \\
    \bottomrule
  \end{tabular}
\end{table}

\subsection{Relatedness matrix}
\label{sec:method-matrix}

The idea of linking technology categories to the 52 O{*}NET job abilities dates to \citet{felten2018method}; we estimate functional relatedness between each of the 16 capabilities in Table~\ref{tab:capabilities} and each of those 52 abilities in O{*}NET ($832$ pairs), following the same scoring and aggregation protocol but using an LLM rather than a human survey. Using GPT-OSS-20B each pair is assigned to an integer from 0 (unrelated) to 4 (strongly related) and we map each rating to a range between 0 and 1 by dividing by four. For robustness and bias purposes we prompt the LLM with four different personas (robotics engineer, occupational therapist, industrial engineer, and workplace safety and ergonomics inspector) and with five independent samples. For each pair we average 20 ratings. Table~\ref{tab:matrix-extremes} shows illustrative high- and low-scoring pairs. The scoring prompt and persona definitions are in Appendix~\ref{app:matrix-prompt}.

% Auto-generated by 100 Paper Scripts/build_matrix_extremes_table.py
\begin{table}[H]
  \centering
  \small
  \caption{Illustrative capability--ability pairs with the highest and lowest normalized relatedness scores ($X_{ca}$). Many pairs tie at 0 or 1 after averaging 20 LLM ratings; rows shown are the first five under alphabetical tie-breaking.}
  \label{tab:matrix-extremes}
  \begin{tabular}{@{}>{\raggedright\arraybackslash}p{0.38\linewidth}>{\raggedright\arraybackslash}p{0.38\linewidth}>{\raggedleft\arraybackslash}p{0.14\linewidth}@{}}
    \toprule
    Robotics capability & O{*}NET ability & $X_{ca}$ \\
    \midrule
    \multicolumn{3}{@{}l}{\textit{Highest five}} \\
    Aerial \& Aquatic Locomotion & Control Precision & 1.000 \\
    Complex Object Manipulation & Manual Dexterity & 1.000 \\
    Dexterous \& In-Hand Manipulation & Finger Dexterity & 1.000 \\
    Dexterous \& In-Hand Manipulation & Manual Dexterity & 1.000 \\
    Expressive \& Communicative Interaction & Oral Expression & 1.000 \\
    \addlinespace
    \multicolumn{3}{@{}l}{\textit{Lowest five}} \\
    Aerial \& Aquatic Locomotion & Finger Dexterity & 0.000 \\
    Aerial \& Aquatic Locomotion & Fluency of Ideas & 0.000 \\
    Aerial \& Aquatic Locomotion & Oral Expression & 0.000 \\
    Aerial \& Aquatic Locomotion & Speech Clarity & 0.000 \\
    Aerial \& Aquatic Locomotion & Written Comprehension & 0.000 \\
    \bottomrule
  \end{tabular}
\end{table}

\subsection{Exposure index formulas (ROE, RIE, RGE)}
\label{sec:method-formulas}

We aggregate the relatedness matrix (Section~\ref{sec:method-matrix}) to occupation-, industry-, and county-level indices, treating each occupation as a bundle of O{*}NET abilities in the spirit of \citet{felten2018method}.
Subscripts $c$, $a$, and $o$ refer to robotics capabilities (c) in Table~\ref{tab:capabilities}, O{*}NET work abilities (a), and occupations (o). We denote $X_{ca} \in [0,1]$ as the normalized relatedness between capability $c$ and ability $a$ (the mean LLM rating divided by four). For each ability $a$, ability-level relatedness $R_a$ sums across capabilities:
\begin{equation}
  R_a = \sum_{c=1}^{16} X_{ca}.
  \label{eq:ability-exposure}
\end{equation}

O{*}NET reports for each pair $(o,a)$ an importance rating IM on a 1--5 Likert scale and a level rating LV on a 0--7 Likert scale. We normalize IM and LV by dividing by 5 and 7. Robotics Occupational Exposure weights each $R_a$ by how much ability $a$ matters in occupation $o$, using the same importance--level weighting:
\begin{equation}
  \mathrm{ROE}^{\mathrm{raw}}_{o}
  = \frac{\sum_a R_a\,(\mathrm{IM}_{oa}/5)\,(\mathrm{LV}_{oa}/7)}
         {\sum_a (\mathrm{IM}_{oa}/5)\,(\mathrm{LV}_{oa}/7)}.
  \label{eq:roe-raw}
\end{equation}
The denominator averages over abilities so occupations remain comparable when O{*}NET reports different numbers of rated abilities. When several O{*}NET--SOC codes map to the same six-digit SOC occupation, we average $\mathrm{ROE}^{\mathrm{raw}}$ across variants. The standardized ROE is
\begin{equation}
  \mathrm{ROE}_o = \frac{\mathrm{ROE}^{\mathrm{raw}}_o - \mu}{\sigma},
  \label{eq:roe-std}
\end{equation}
where $\mu$ and $\sigma$ are the cross-occupation mean and standard deviation of $\mathrm{ROE}^{\mathrm{raw}}$.
For job-family summaries we assign each six-digit SOC occupation to a broad job family using the O{*}NET--SOC crosswalk in the Felten-style taxonomy.
We report both the unweighted mean of occupation-level standardized ROE across occupations in the family and an employment-weighted mean using BLS May OEWS national employment counts.
We compare ROE to published AIOE from \citet{felten2021occupational} on six-digit SOC codes. Both lists span 774 occupations, but 92 codes differ because the AIOE appendix uses superseded SOC identifiers that current O{*}NET no longer reports.
We keep the 682 exact matches, re-standardize ROE and AIOE on that sample, and aggregate to job families on a common scale.

Robotics Industry Exposure (RIE) employment-weights standardized ROE to detailed NAICS industries using occupation--industry counts from BLS May OES. We index $i$ for industries and $e_{o,i}$ for employment of occupation $o$ in industry $i$:
\begin{equation}
  \mathrm{RIE}^{\mathrm{raw}}_{i} = \frac{\sum_o \mathrm{ROE}_o\, e_{o,i}}{\sum_o e_{o,i}}, \qquad
  \mathrm{RIE}_i = \frac{\mathrm{RIE}^{\mathrm{raw}}_{i} - \mu}{\sigma}.
  \label{eq:rie}
\end{equation}
where $\mu$ and $\sigma$ are the cross-industry mean and standard deviation of $\mathrm{RIE}^{\mathrm{raw}}$.

Robotics Geographic Exposure (RGE) employment-weights standardized RIE to counties using 2019 county-by-four-digit-NAICS employment from BLS QCEW. We keep only county-level records, dropping QCEW statewide totals (FIPS codes ending in 000), metropolitan-area totals, statewide or unknown county placeholders (codes ending in 999), and the U.S.\ Virgin Islands (codes beginning with 78), which leaves 3{,}217 counties and county equivalents. We index $g$ for counties (five-digit FIPS) and $h_{i,g}$ for employment of industry $i$ in county $g$:
\begin{equation}
  \mathrm{RGE}^{\mathrm{raw}}_{g} = \frac{\sum_i \mathrm{RIE}_i\, h_{i,g}}{\sum_i h_{i,g}}, \qquad
  \mathrm{RGE}_g = \frac{\mathrm{RGE}^{\mathrm{raw}}_{g} - \mu}{\sigma}.
  \label{eq:rge}
\end{equation}
where $\mu$ and $\sigma$ are the cross-county mean and standard deviation of $\mathrm{RGE}^{\mathrm{raw}}$.
As in \citet{felten2018method}, ROE, RIE, and RGE measure how strongly work aligns with robotics capabilities; they do not identify whether robotics substitutes for or complements labor, and they do not reflect technical feasibility given current hardware, software, or models, the economic attractiveness of automation, safety and regulatory constraints, or broader adoption barriers.

\subsection{Labor-market correlates}
\label{sec:method-correlates}

We describe how standardized ROE covaries with occupation-level labor-market attributes in Section~\ref{sec:results-correlates}. The analysis is descriptive: we merge ROE to covariates at six-digit SOC and plot binned scatters with 20 equal-count bins on the covariate; each point reports the bin mean of the covariate (horizontal axis) and the bin mean of standardized ROE (vertical axis), following \citet{felten2024generative}. Covariates are 2021 median wage, required education,\footnote{\label{fn:required-education}Required education is the O{*}NET ``Required Level of Education'' item, coded on 12 categories: (1)~less than a high school diploma; (2)~high school diploma or equivalent; (3)~post-secondary certificate; (4)~some college courses; (5)~associate's degree; (6)~bachelor's degree; (7)~post-baccalaureate certificate; (8)~master's degree; (9)~post-master's certificate; (10)~first professional degree; (11)~doctoral degree; (12)~post-doctoral training. An occupation's value is the average category, weighted by the share of respondents reporting each category.} creative-ability weight,\footnote{\label{fn:creative-weight}Creative-ability weight is the share of an occupation's total O{*}NET ability weight accounted for by the two creativity-oriented abilities, Originality and Category Flexibility, where each of the 52 abilities is weighted by its importance and level ratings \citep{felten2024generative}.} and occupational shares by sex and race; occupation-level values are taken from the data accompanying \citet{felten2024generative}.

\section{Results}
\label{sec:results}

\subsection{Exposure levels}
\label{sec:results-exposure}

\subsubsection{Occupational exposure}
\label{sec:results-exposure-occ}

Standardized ROE ranges from $-2.38$ to $2.11$ across the 774 six-digit SOC occupations in the O{*}NET-based sample, a span of about 4.5 standard deviations.
Table~\ref{tab:roe-occ-rankings} lists the highest- and lowest-ranked occupations.

\begin{figure}[H]
  \centering
  \begin{subfigure}[t]{0.48\textwidth}
    \centering
    \includegraphics[width=\linewidth]{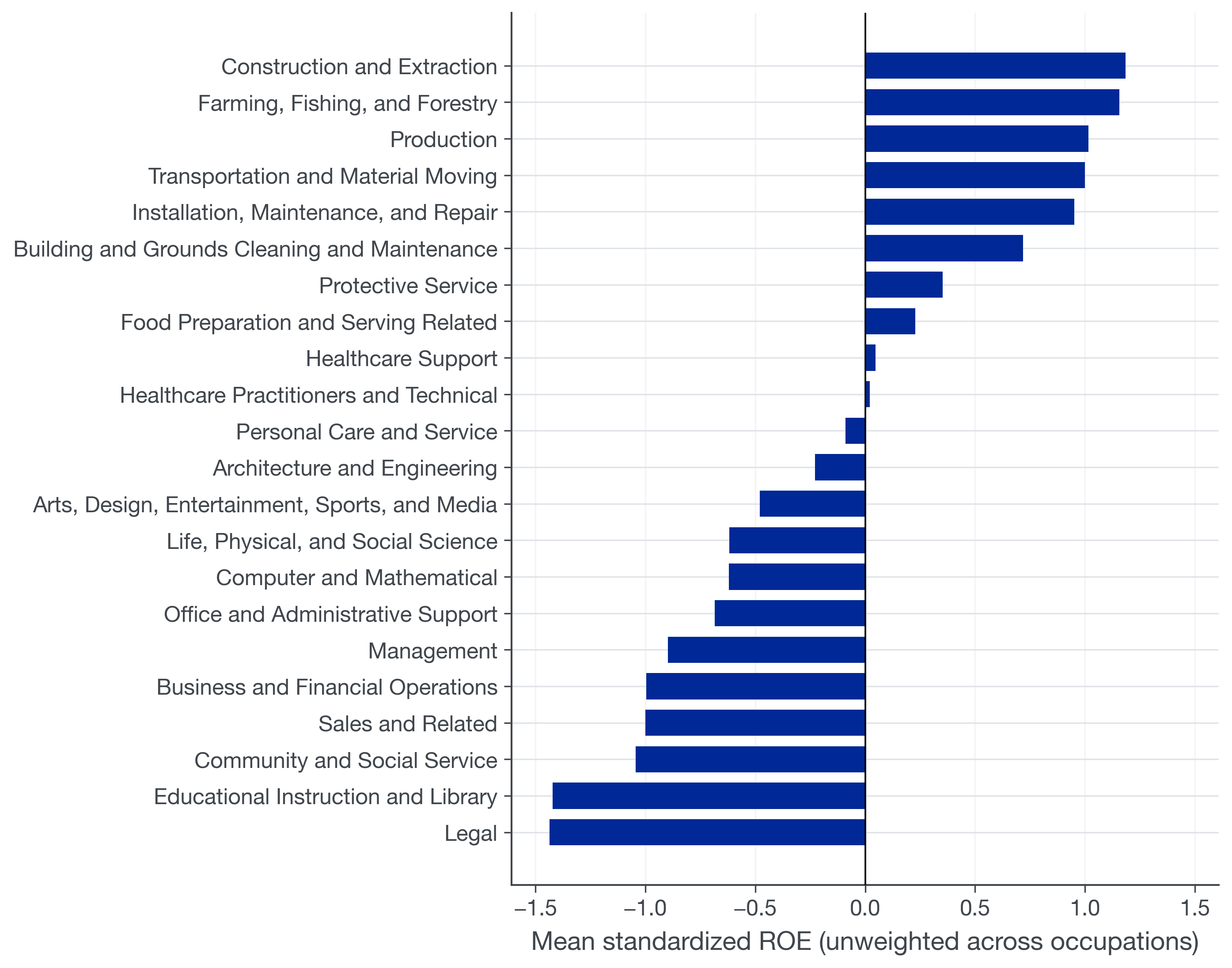}
    \caption{Mean standardized ROE (unweighted across occupations)}
    \label{fig:job-family-unweighted}
  \end{subfigure}\hfill
  \begin{subfigure}[t]{0.48\textwidth}
    \centering
    \includegraphics[width=\linewidth]{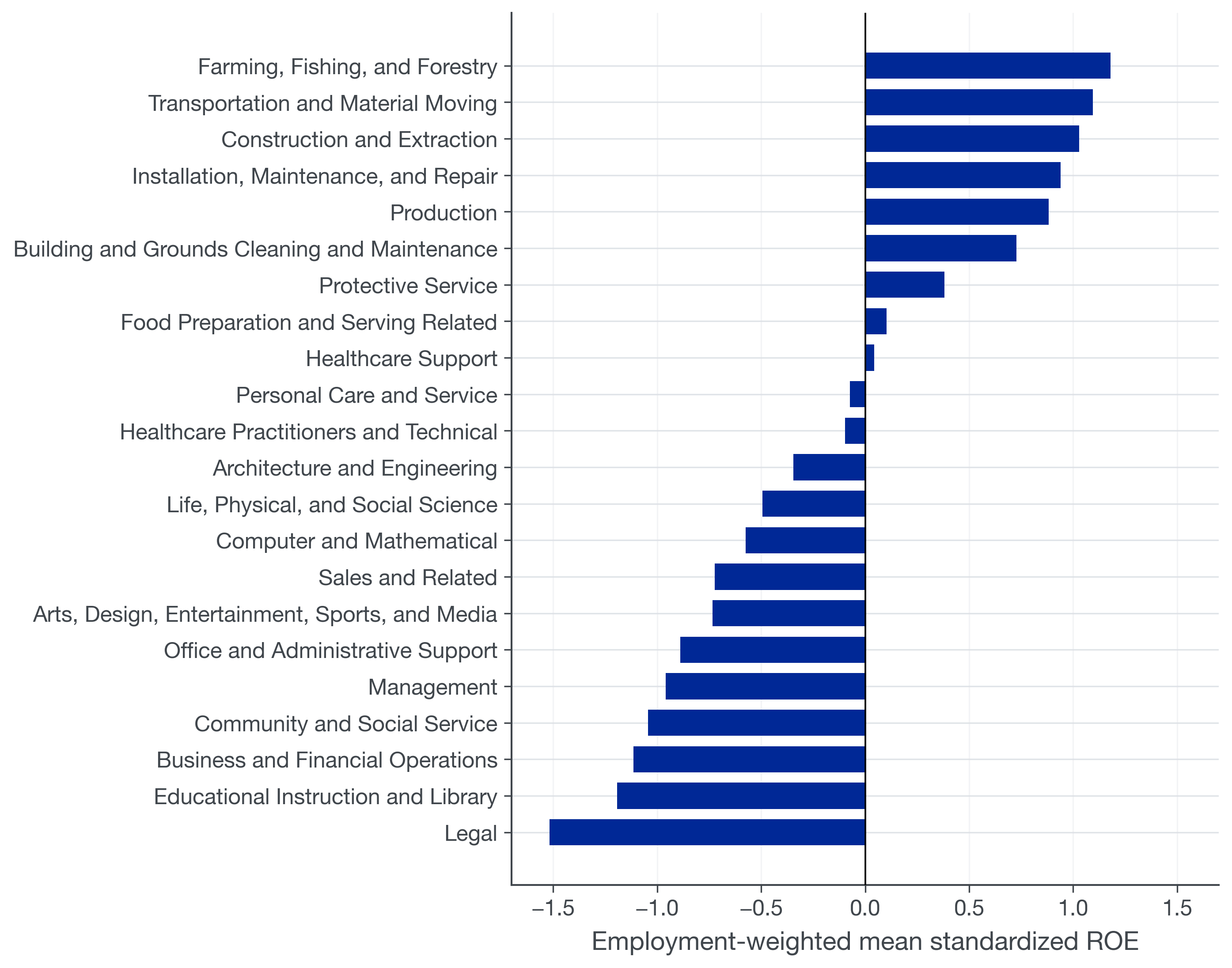}
    \caption{Mean standardized ROE (employment-weighted)}
    \label{fig:job-family-weighted}
  \end{subfigure}
  \caption{Robotics exposure by job family.
  Panel~(a) reports $\overline{\mathrm{ROE}}^{\mathrm{unw}}_{f} = \frac{1}{n_f}\sum_{o \in f} \mathrm{ROE}_{o}$, where $n_f$ is the number of six-digit occupations in family~$f$, unweighted by employment counts.
  Panel~(b) reports $\overline{\mathrm{ROE}}^{\mathrm{emp}}_{f} = \sum_{o \in f} \mathrm{ROE}_{o}\, e_{o} \big/ \sum_{o \in f} e_{o}$, where $e_{o}$ is national employment from BLS May OEWS.}
  \label{fig:job-family}
\end{figure}

We next summarize exposure by broad job family.
Each occupation maps to a single family in the Felten-style O{*}NET--SOC taxonomy.
Figure~\ref{fig:job-family} reports family-level means in two ways: an unweighted mean across occupations in the family (panel~(a)) and an employment-weighted mean using May 2025 OEWS national counts (panel~(b)).

% Auto-generated by 100 Paper Scripts/build_ranking_tables.py — do not edit by hand.
\begin{table}[H]
\centering
\caption{Occupational robotics exposure (standardized O*NET-based ROE, $n=774$).}
\label{tab:roe-occ-rankings}
\vspace{0.25em}
\begin{minipage}[t]{0.48\textwidth}
\centering\small
\textbf{Highest exposure (ranks 1--20)}\par\vspace{0.35em}
\begin{tabular}{@{}r>{\raggedright\arraybackslash}p{0.88\linewidth}@{}}
\toprule
Rank & Occupation \\
\midrule
1 & Pile-Driver Operators \\
2 & Sewing Machine Operators \\
3 & Sewers, Hand \\
4 & Logging Equipment Operators \\
5 & Pressers, Textile, Garment, and Related Materials \\
6 & Industrial Truck and Tractor Operators \\
7 & Heavy and Tractor-Trailer Truck Drivers \\
8 & Operating Engineers and Other Construction Equipment Operators \\
9 & Fishing and Hunting Workers \\
10 & Paving, Surfacing, and Tamping Equipment Operators \\
11 & Fallers \\
12 & Dredge Operators \\
13 & Cleaners of Vehicles and Equipment \\
14 & Airline Pilots, Copilots, and Flight Engineers \\
15 & Crane and Tower Operators \\
16 & Excavating and Loading Machine and Dragline Operators, Surface Mining \\
17 & Continuous Mining Machine Operators \\
18 & Tree Trimmers and Pruners \\
19 & Rail-Track Laying and Maintenance Equipment Operators \\
20 & Bus Drivers, Transit and Intercity \\
\bottomrule
\end{tabular}
\end{minipage}
\hfill
\begin{minipage}[t]{0.48\textwidth}
\centering\small
\textbf{Lowest exposure (ranks 755--774)}\par\vspace{0.35em}
\begin{tabular}{@{}r>{\raggedright\arraybackslash}p{0.88\linewidth}@{}}
\toprule
Rank & Occupation \\
\midrule
755 & Home Economics Teachers, Postsecondary \\
756 & Environmental Science Teachers, Postsecondary \\
757 & Mathematical Science Teachers, Postsecondary \\
758 & Anthropology and Archeology Teachers, Postsecondary \\
759 & Criminal Justice and Law Enforcement Teachers, Postsecondary \\
760 & Social Work Teachers, Postsecondary \\
761 & Business Teachers, Postsecondary \\
762 & Geography Teachers, Postsecondary \\
763 & Door-To-Door Sales Workers, News and Street Vendors, and Related Workers \\
764 & Law Teachers, Postsecondary \\
765 & Psychology Teachers, Postsecondary \\
766 & Communications Teachers, Postsecondary \\
767 & Sociology Teachers, Postsecondary \\
768 & History Teachers, Postsecondary \\
769 & Philosophy and Religion Teachers, Postsecondary \\
770 & Area, Ethnic, and Cultural Studies Teachers, Postsecondary \\
771 & English Language and Literature Teachers, Postsecondary \\
772 & Telemarketers \\
773 & Political Science Teachers, Postsecondary \\
774 & Foreign Language and Literature Teachers, Postsecondary \\
\bottomrule
\end{tabular}
\end{minipage}
\end{table}

\subsubsection{Industry exposure (RIE)}
\label{sec:results-exposure-ind}

Table~\ref{tab:rie-ind-rankings} reports the industries with the highest and lowest employment-weighted RIE.

% Auto-generated by 100 Paper Scripts/build_ranking_tables.py — do not edit by hand.
\begin{table}[H]
\centering
\caption{Industry robotics exposure (standardized RIE, $n=250$ NAICS industries).}
\label{tab:rie-ind-rankings}
\vspace{0.25em}
\begin{minipage}[t]{0.48\textwidth}
\centering\small
\textbf{Highest exposure (ranks 1--20)}\par\vspace{0.35em}
\begin{tabular}{@{}r>{\raggedright\arraybackslash}p{0.88\linewidth}@{}}
\toprule
Rank & Industry \\
\midrule
1 & Logging \\
2 & Truck Transportation \\
3 & Coal Mining \\
4 & Urban Transit Systems \\
5 & Charter Bus Industry \\
6 & Support Activities for Crop Production \\
7 & Fiber, Yarn, and Thread Mills \\
8 & Apparel Accessories and Other Apparel Manufacturing \\
9 & Sawmills and Wood Preservation \\
10 & Nonmetallic Mineral Mining and Quarrying \\
11 & Highway, Street, and Bridge Construction \\
12 & Interurban and Rural Bus Transportation \\
13 & Textile Furnishings Mills \\
14 & Other Specialty Trade Contractors \\
15 & Couriers and Express Delivery Services \\
16 & Support Activities for Rail Transportation \\
17 & Support Activities for Road Transportation \\
18 & Waste Collection \\
19 & Support Activities for Water Transportation \\
20 & Warehousing and Storage \\
\bottomrule
\end{tabular}
\end{minipage}
\hfill
\begin{minipage}[t]{0.48\textwidth}
\centering\small
\textbf{Lowest exposure (ranks 231--250)}\par\vspace{0.35em}
\begin{tabular}{@{}r>{\raggedright\arraybackslash}p{0.88\linewidth}@{}}
\toprule
Rank & Industry \\
\midrule
231 & Other Information Services \\
232 & Credit Intermediation and Related Activities (5221 And 5223 only) \\
233 & Accounting, Tax Preparation, Bookkeeping, and Payroll Services \\
234 & Elementary and Secondary Schools \\
235 & Insurance Carriers \\
236 & Educational Support Services \\
237 & Lessors of Nonfinancial Intangible Assets (except Copyrighted Works) \\
238 & Colleges, Universities, and Professional Schools \\
239 & Insurance and Employee Benefit Funds \\
240 & Business Support Services \\
241 & Business Schools and Computer and Management Training \\
242 & Nondepository Credit Intermediation \\
243 & Grantmaking and Giving Services \\
244 & Other Investment Pools and Funds \\
245 & Securities, Commodity Contracts, and Other Financial Investments and Related Activities \\
246 & Travel Arrangement and Reservation Services \\
247 & Agencies, Brokerages, and Other Insurance Related Activities \\
248 & Junior Colleges \\
249 & Agents and Managers for Artists, Athletes, Entertainers, and Other Public Figures \\
250 & Legal Services \\
\bottomrule
\end{tabular}
\end{minipage}
\end{table}

\subsubsection{Geographic exposure (RGE)}
\label{sec:results-exposure-geo}

\begin{figure}[H]
  \centering
  \begin{subfigure}[t]{0.49\textwidth}
    \centering
    \includegraphics[width=\linewidth]{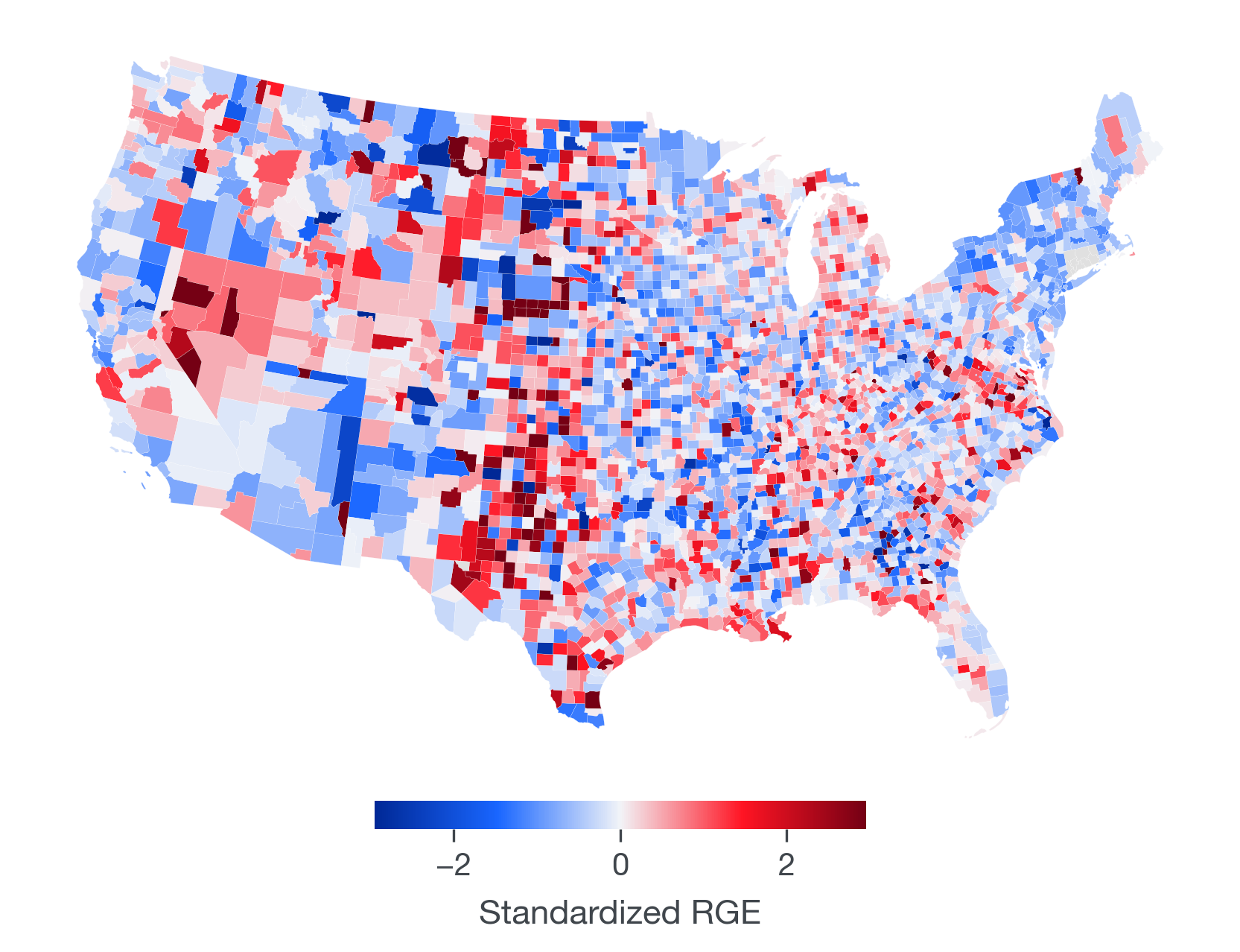}
    \caption{County-level RGE}
    \label{fig:rge-county}
  \end{subfigure}\hfill
  \begin{subfigure}[t]{0.49\textwidth}
    \centering
    \includegraphics[width=\linewidth]{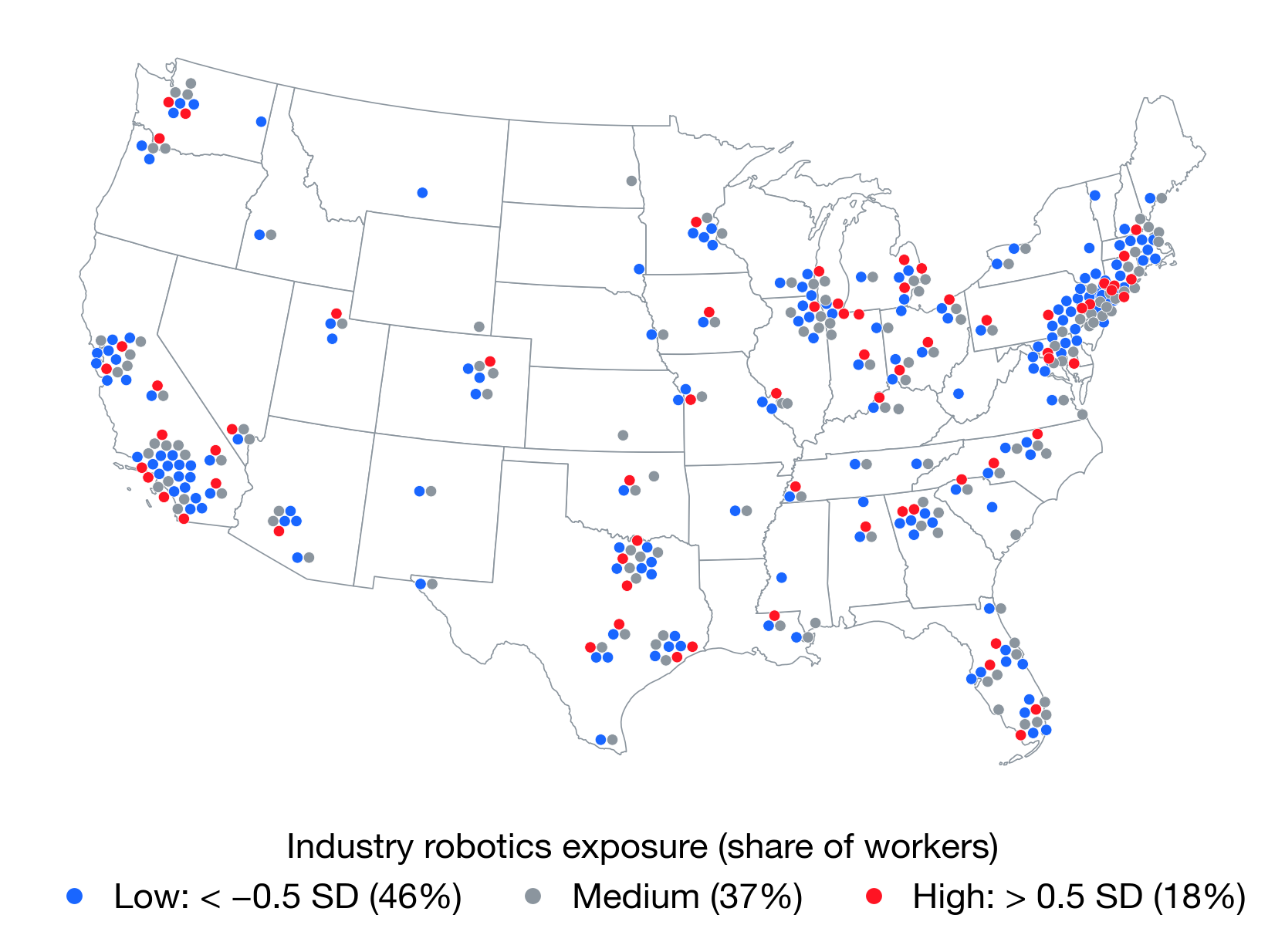}
    \caption{Workers by industry exposure class}
    \label{fig:rge-dots}
  \end{subfigure}
  \caption{Geographic robotics exposure, contiguous U.S.
  Panel~(a): standardized RGE by county; the color scale is capped at the 1st and 99th percentiles, and gray counties have no matched employment.
  Panel~(b): workers by industry exposure class, one dot per 250,000 workers.
  A state with only one or two dots, such as Wyoming, can show only its largest classes.}
  \label{fig:rge}
\end{figure}

Figure~\ref{fig:rge-county} maps county-level RGE, with each county having equal weight.
In sparsely populated counties, few industries have reported and matched employment, so a single industry can dominate the score; the most extreme values come from small rural counties.
Because every county is weighted equally, the map reflects land area more than where workers live.
Therefore, we also present a dot-density version (Figure~\ref{fig:rge-dots}) that shows where workers in low-, medium-, and high-exposure industries are employed.
We classify four-digit NAICS industries as low (RIE below $-0.5$ SD), medium, or high exposure (RIE above $0.5$ SD) and count the workers in each class.
Because large industries such as schools and hospitals have low RIE, 46\% of matched workers are in low-exposure industries, 37\% in medium, and 18\% in high.
Each dot represents 250,000 workers in 2019 QCEW county-by-industry employment.
A state's number of dots is its employment divided by 250,000, rounded to a whole number; Texas, for example, has 8.5 million matched workers and receives 34 dots.
These dots are then colored in proportion to the state's workers in each class.
Within a state, dots are drawn near the counties where the workers are employed, so their positions are approximate.
Across the 48 contiguous states and the District of Columbia, the share of workers in high-exposure industries ranges from 7\% in the District of Columbia and 12\% in New York to 23\% in Indiana and 24\% in Kentucky; it is highest in North Dakota (28\%) and Wyoming (31\%), which have one dot each.

\subsection{Labor-market correlates}
\label{sec:results-correlates}

\subsubsection{Wage, education, and creative ability}
\label{sec:results-correlates-wage}

\begin{figure}[H]
  \centering
  \begin{subfigure}[t]{0.32\textwidth}
    \centering
    \includegraphics[width=\linewidth]{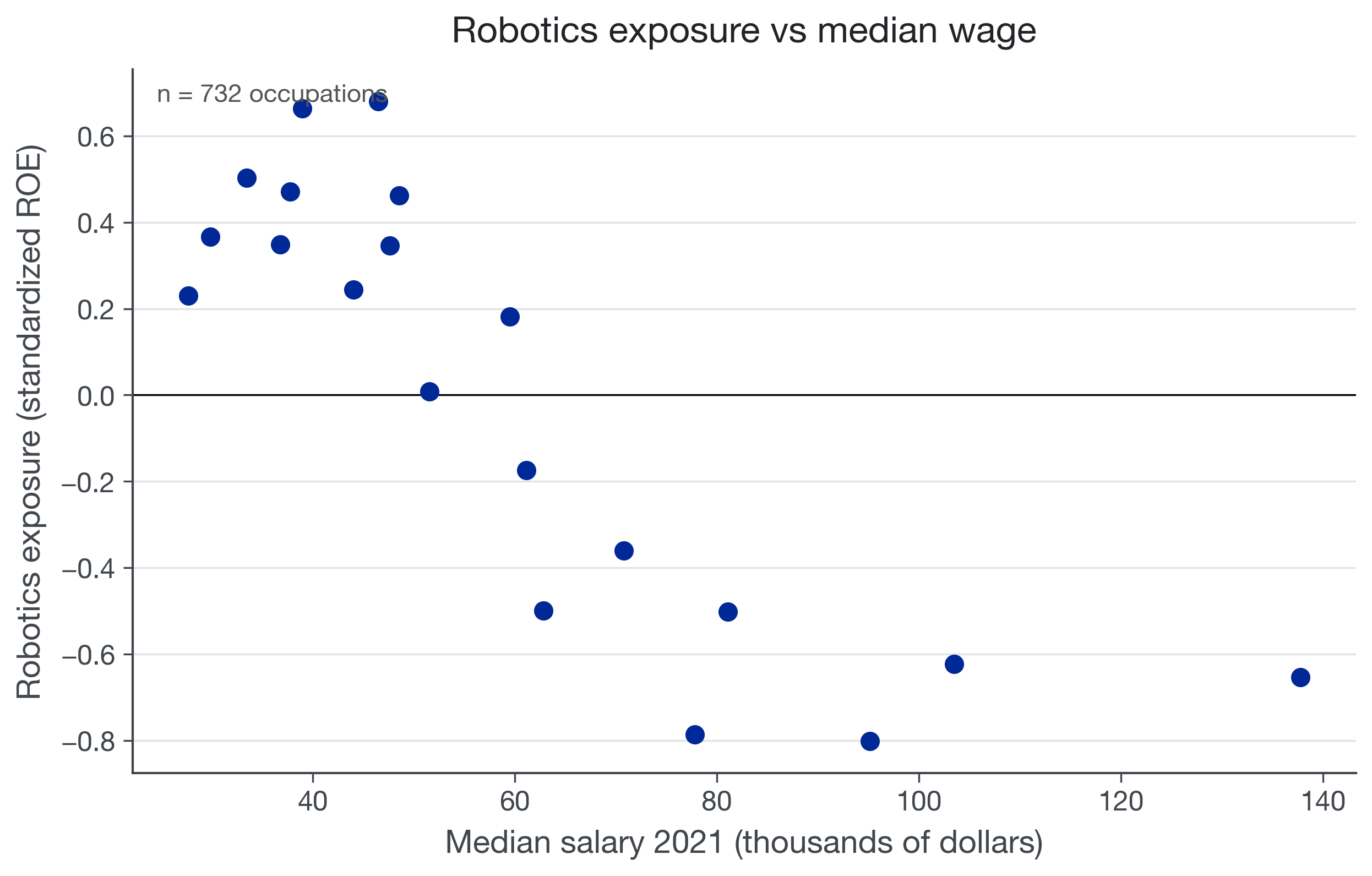}
    \caption{Median wage (2021)}
    \label{fig:roe-wage}
  \end{subfigure}\hfill
  \begin{subfigure}[t]{0.32\textwidth}
    \centering
    \includegraphics[width=\linewidth]{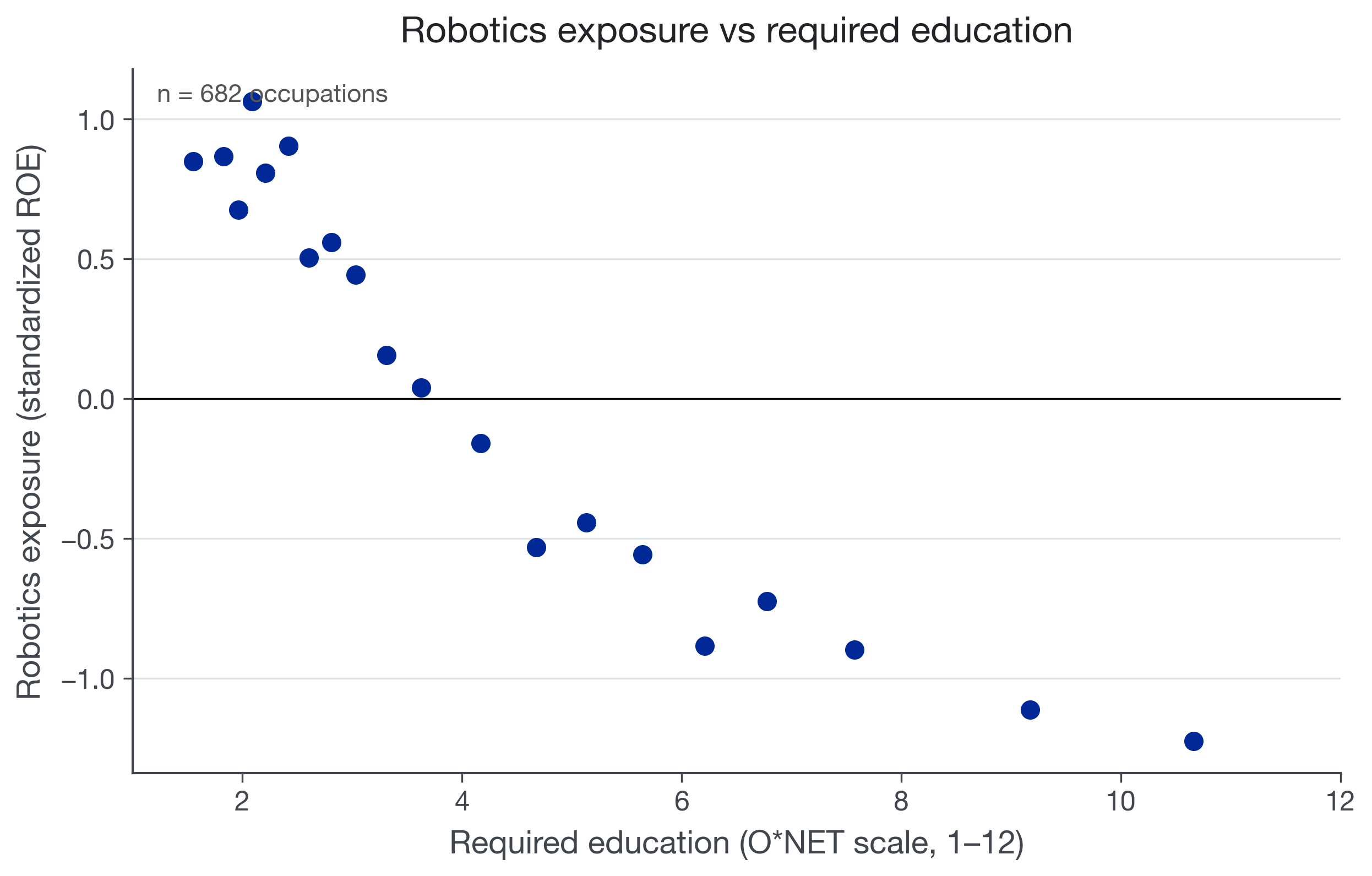}
    \caption{Required education}
    \label{fig:roe-education}
  \end{subfigure}\hfill
  \begin{subfigure}[t]{0.32\textwidth}
    \centering
    \includegraphics[width=\linewidth]{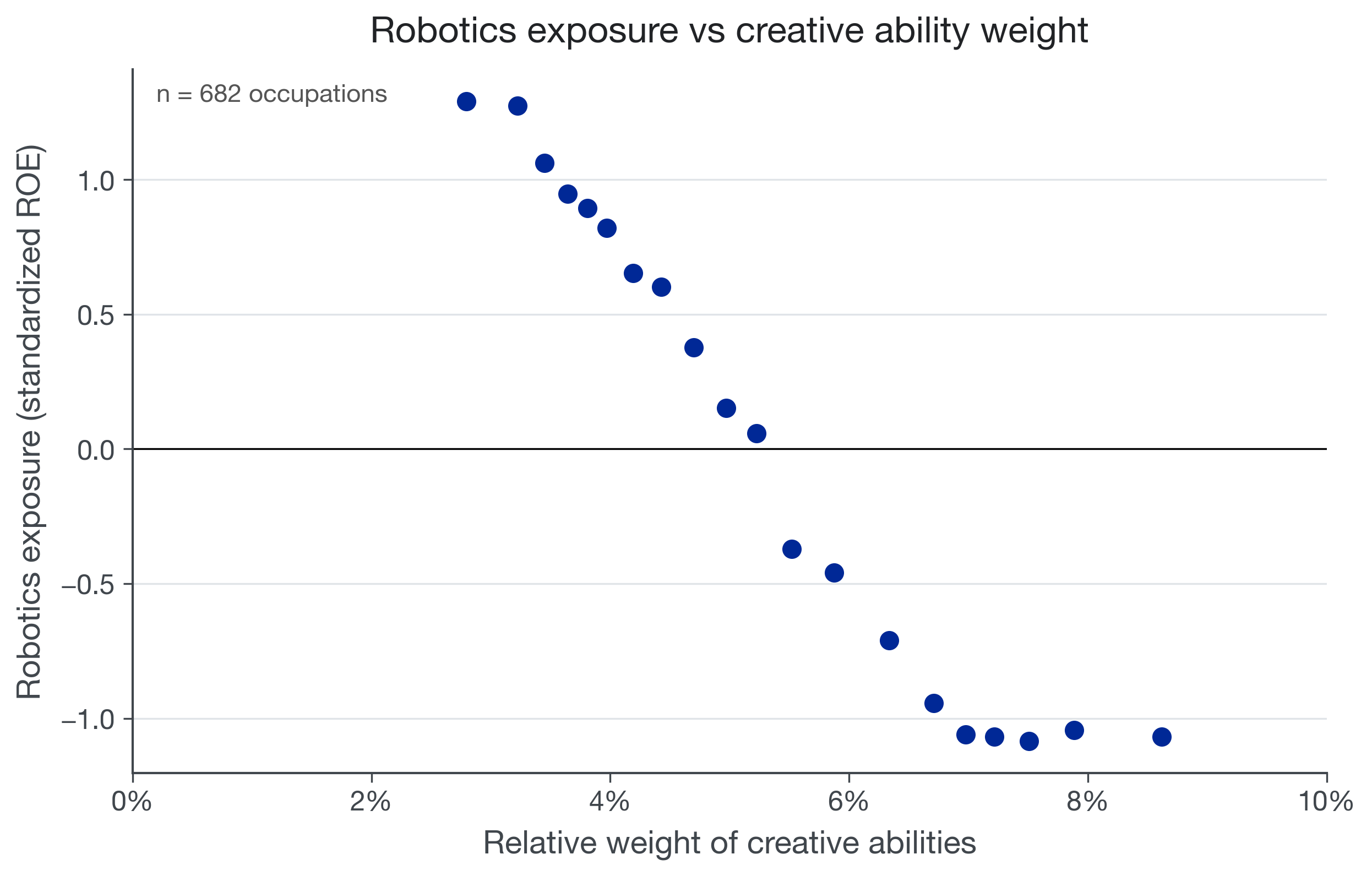}
    \caption{Creative-ability weight}
    \label{fig:roe-creative}
  \end{subfigure}
  \caption{Robotics exposure and labor-market correlates (O{*}NET ROE). Each point is the mean of standardized ROE within an equal-count bin on the horizontal axis.
  Required education is on the O{*}NET 1--12 scale (2 = high school diploma, 4 = some college, 6 = bachelor's, 8 = master's, 11 = doctorate; full list in footnote~\ref{fn:required-education}).
  Creative-ability weight is defined in footnote~\ref{fn:creative-weight}.}
  \label{fig:roe-correlates-main}
\end{figure}

Figure~\ref{fig:roe-correlates-main} plots binned relationships between standardized ROE and labor-market attributes, following the descriptive approach in \citet{felten2024generative}.
Each point is the mean of standardized ROE within an equal-count bin on the covariate shown on the horizontal axis.
Mean standardized ROE falls from $+0.23$ in the lowest-wage bin to $-0.65$ in the highest, so robotics exposure concentrates in lower-paying work (Spearman $\rho = -0.42$).
The education gradient is steeper: ROE falls from $+0.85$ in the bin with the lowest required education to $-1.22$ in the highest ($\rho = -0.71$).
Creative-ability weight separates occupations most sharply of the three, from $+1.29$ in the least creative bin to $-1.07$ in the most ($\rho = -0.85$).

\subsubsection{Demographic composition}
\label{sec:results-correlates-demo}

\begin{figure}[H]
  \centering
  \begin{subfigure}[t]{0.48\textwidth}
    \centering
    \includegraphics[width=\linewidth]{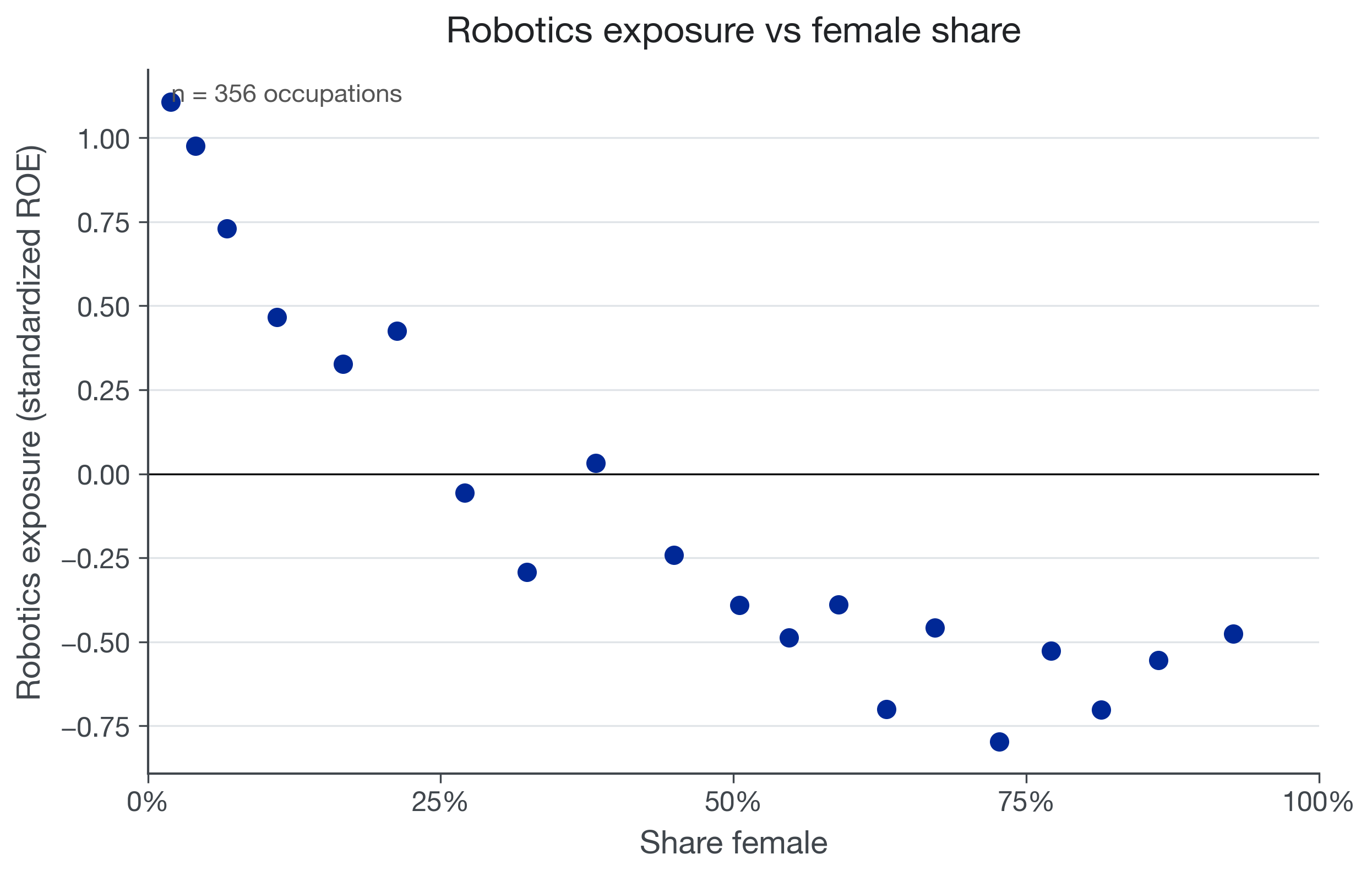}
    \caption{Female share}
    \label{fig:roe-female-share}
  \end{subfigure}\hfill
  \begin{subfigure}[t]{0.48\textwidth}
    \centering
    \includegraphics[width=\linewidth]{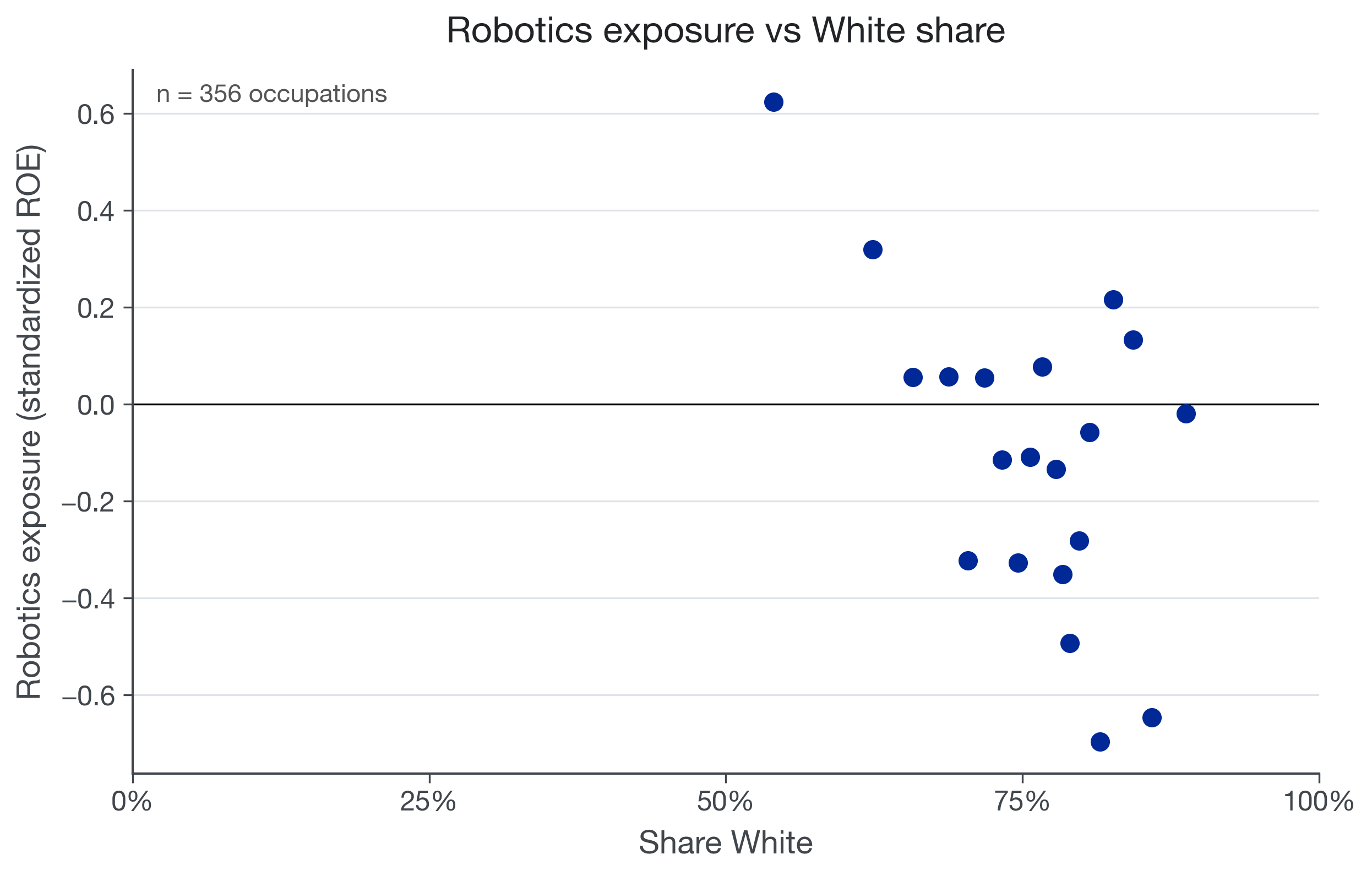}
    \caption{White share}
    \label{fig:roe-white-share}
  \end{subfigure}\\[0.75em]
  \begin{subfigure}[t]{0.48\textwidth}
    \centering
    \includegraphics[width=\linewidth]{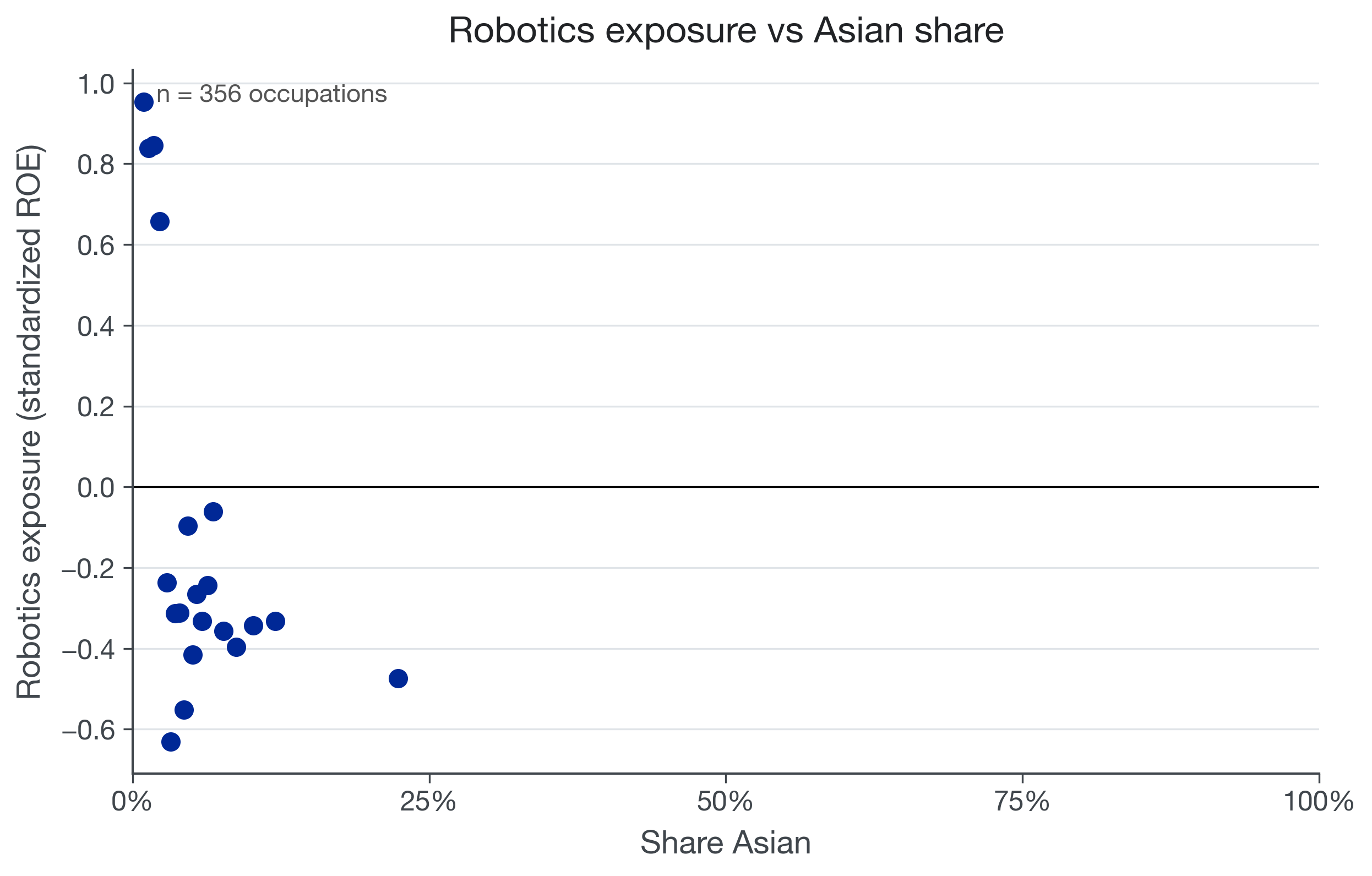}
    \caption{Asian share}
    \label{fig:roe-asian-share}
  \end{subfigure}\hfill
  \begin{subfigure}[t]{0.48\textwidth}
    \centering
    \includegraphics[width=\linewidth]{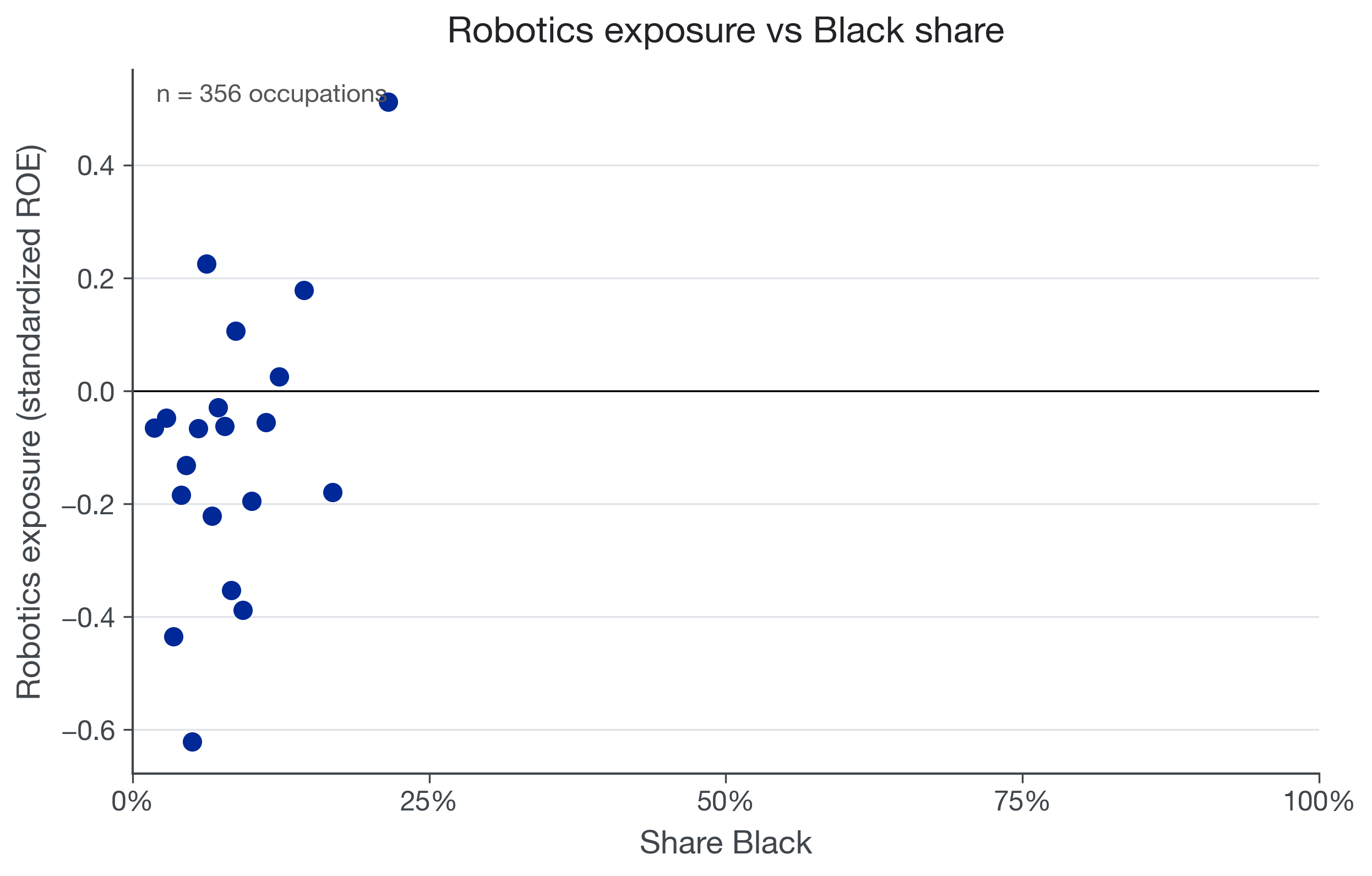}
    \caption{Black share}
    \label{fig:roe-black-share}
  \end{subfigure}
  \caption{Robotics exposure and demographic composition of occupations.}
  \label{fig:roe-correlates-demo}
\end{figure}

Figure~\ref{fig:roe-correlates-demo} plots bin means of standardized ROE against occupational shares by sex and race.
Sample sizes for demographic covariates are smaller than for wage and education because not every occupation reports full breakdowns in the underlying data.
Exposure is highest in occupations with few women, falling from $+1.11$ in the lowest female-share bin to $-0.48$ in the highest ($\rho = -0.55$). The gradients by race are weaker: exposure declines modestly with Asian share, and no clear trend is visible by white or Black share.

\subsection{Robotics exposure versus AI exposure}
\label{sec:results-ai}

\subsubsection{Comparison of exposure}
\label{sec:results-ai-exposure}

\begin{figure}[H]
  \centering
  \begin{subfigure}[t]{0.48\textwidth}
    \centering
    \includegraphics[width=\linewidth]{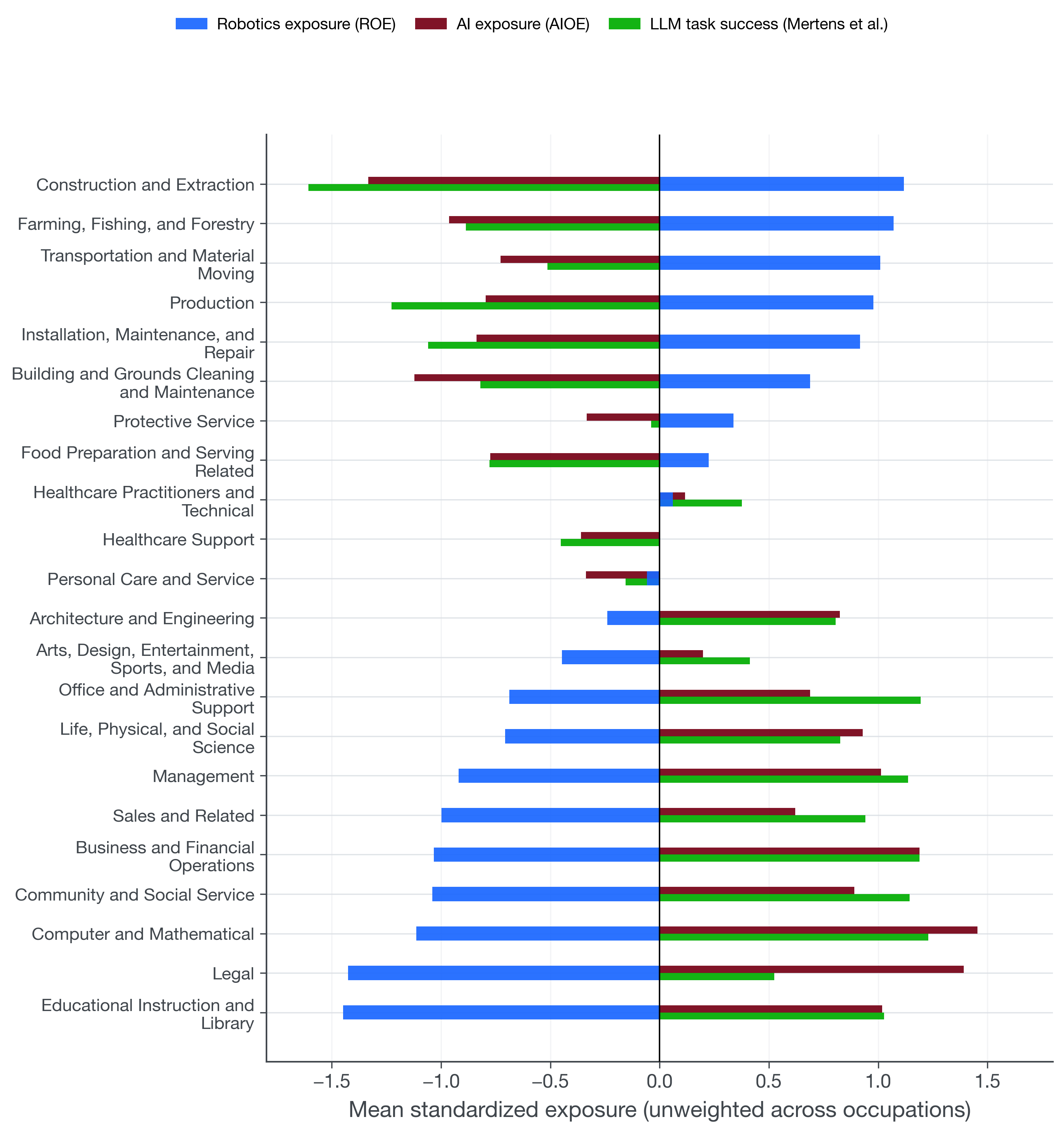}
    \caption{Mean standardized exposure (unweighted across occupations)}
    \label{fig:roe-aioe-job-family-bars-unw}
  \end{subfigure}\hfill
  \begin{subfigure}[t]{0.48\textwidth}
    \centering
    \includegraphics[width=\linewidth]{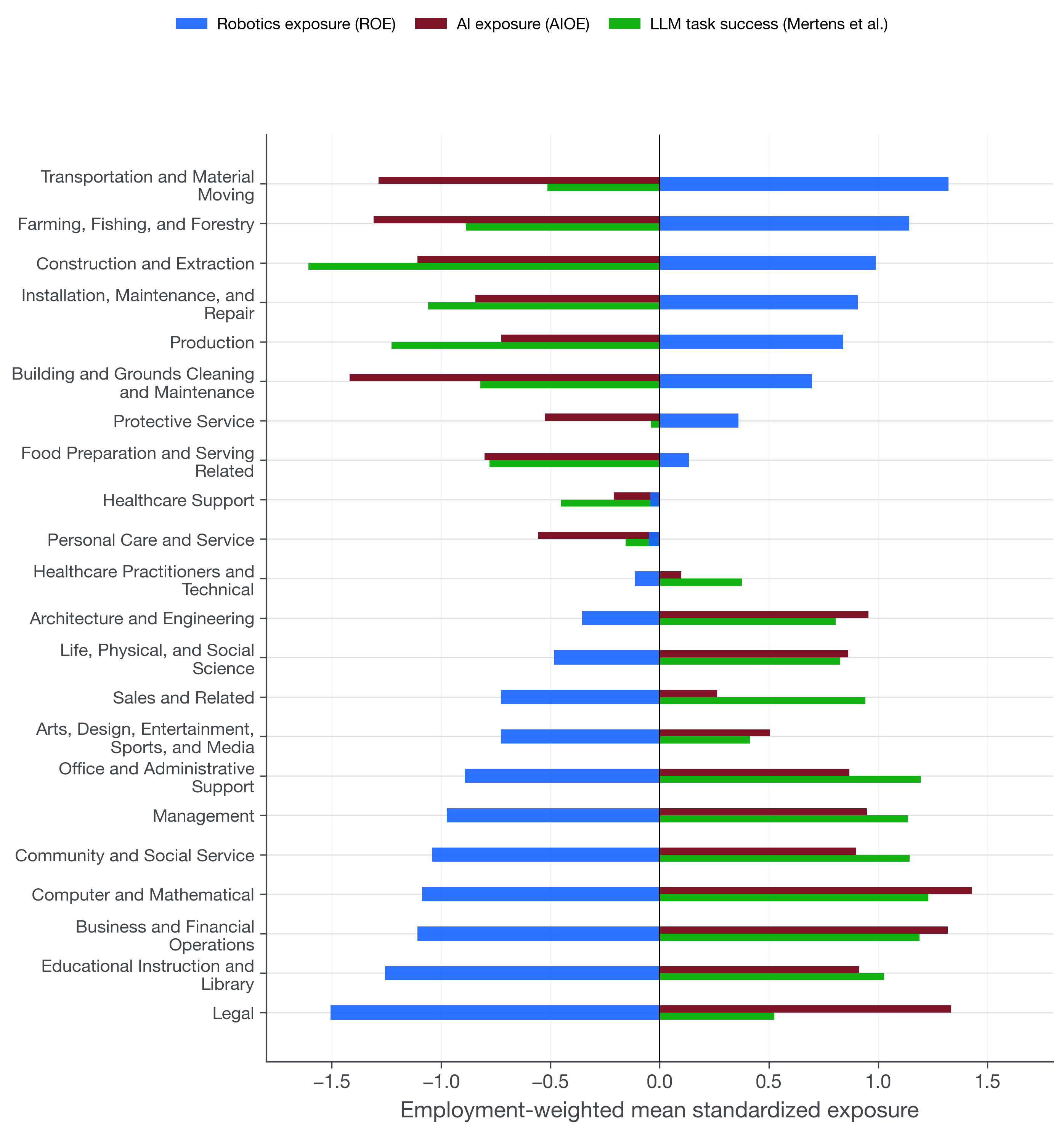}
    \caption{Mean standardized exposure (employment-weighted)}
    \label{fig:roe-aioe-job-family-bars-emp}
  \end{subfigure}
  \caption{Robotics exposure (ROE), AI exposure (AIOE), and LLM task success by job family (matched sample, $n=682$ occupations).
  After z-scoring each measure on the matched sample, panel~(a) averages within family with equal occupation weights; panel~(b) uses OEWS employment weights.
  LLM task success is the share of a family's O{*}NET tasks with at least 10\% LLM time-saving potential times the acceptance rate on the surveyed subset of those tasks, pooled over 41 models \citep[Figure~2a and Table~2]{mertens2026crashing}.}
  \label{fig:roe-aioe-job-family-bars}
\end{figure}

\begin{figure}[H]
  \centering
  \noindent
  \begin{minipage}[c]{0.57\textwidth}
    \centering
    \includegraphics[width=\linewidth]{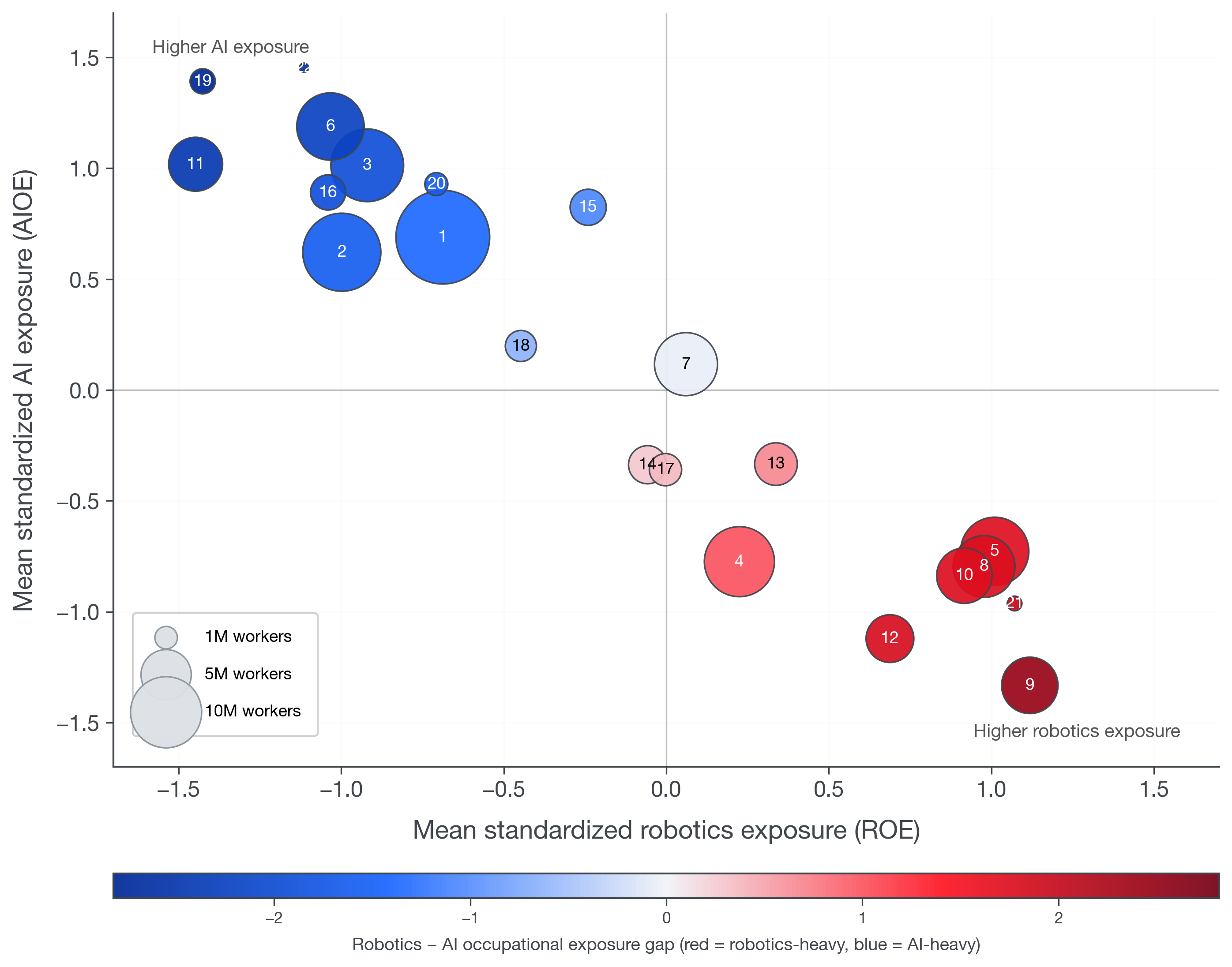}
  \end{minipage}%
  \hfill
  \begin{minipage}[c]{0.41\textwidth}
    \centering
    \resizebox{\linewidth}{!}{%
      \scriptsize
      % Auto-generated by 100 Paper Scripts/build_job_family_bubble_table.py
\begin{tabular}{@{}r>{\raggedright\arraybackslash}p{0.42\linewidth}rrr@{}}
\toprule
\# & Job family & Workers (M) & ROE & AIOE \\
\midrule
1 & Office \& admin.\ support & 17.4 & -0.69 & 0.69 \\
2 & Sales & 12.1 & -1.00 & 0.62 \\
3 & Management & 10.5 & -0.92 & 1.01 \\
4 & Food prep.\ \& serving & 9.7 & 0.22 & -0.77 \\
5 & Transportation \& material moving & 9.2 & 1.01 & -0.73 \\
6 & Business \& financial & 9.0 & -1.03 & 1.19 \\
7 & Healthcare practitioners & 7.9 & 0.06 & 0.12 \\
8 & Production & 7.6 & 0.98 & -0.80 \\
9 & Construction \& extraction & 6.3 & 1.12 & -1.33 \\
10 & Installation \& repair & 6.1 & 0.92 & -0.84 \\
11 & Education \& library & 5.8 & -1.45 & 1.02 \\
12 & Building \& grounds maint. & 4.5 & 0.69 & -1.12 \\
13 & Protective Service & 3.6 & 0.34 & -0.33 \\
14 & Personal care \& service & 2.9 & -0.06 & -0.34 \\
15 & Architecture \& engineering & 2.6 & -0.24 & 0.82 \\
16 & Community \& social service & 2.5 & -1.04 & 0.89 \\
17 & Healthcare Support & 2.1 & -0.00 & -0.36 \\
18 & Arts \& media & 1.9 & -0.45 & 0.20 \\
19 & Legal & 1.3 & -1.43 & 1.39 \\
20 & Life, Physical, and Social Science & 1.1 & -0.71 & 0.93 \\
21 & Farming, Fishing, and Forestry & 0.4 & 1.07 & -0.96 \\
22 & Computer and Mathematical & 0.2 & -1.11 & 1.45 \\
\bottomrule
\end{tabular}
    }
  \end{minipage}
  \caption{Robotics versus AI exposure by job family (matched ROE--AIOE sample).
  Axes show mean standardized ROE and AIOE.
  Bubble area is proportional to BLS May OEWS national employment summed over matched occupations in each family.
  Numbers inside bubbles match rank by workforce size in the table.
  Bubble color shows the robotics-minus-AI gap (red: robotics-heavy; blue: AI-heavy).}
  \label{fig:roe-aioe-job-family-bubble}
\end{figure}

We compare O{*}NET-based ROE to language-modeling AIOE from \citet{felten2024generative} on the 682 six-digit SOC codes that appear in both indices (Section~\ref{sec:method-formulas}).
ROE and published AIOE each cover 774 occupations, but they rely on different O{*}NET--SOC occupational classification vintages, so 92 codes do not line up exactly; we do not impute AIOE across renumbered codes.
Job-family comparisons apply the same matched sample.
For each measure we z-score occupations on that sample, then aggregate within each family.
Figure~\ref{fig:roe-aioe-job-family-bars} reports unweighted occupation means (panel~(a)) and employment-weighted means using May 2025 OEWS counts (panel~(b)).
The two indices order the 22 families almost inversely (Spearman $\rho = -0.93$): construction and extraction is at once the most robotics-exposed family and the least exposed to AI, with computer and mathematical occupations at the opposite corner.
Figure~\ref{fig:roe-aioe-job-family-bubble} plots the same family means; bubble area is proportional to OEWS employment in the matched occupations, and bubble color encodes mean standardized ROE minus mean standardized AIOE.

Figure~\ref{fig:roe-aioe-job-family-bars} also adds a measure of realized language-model performance from the worker evaluations of \citet{mertens2026crashing}.
Whereas AIOE measures how strongly an occupation's abilities relate to AI applications, \citet{mertens2026crashing} ask whether LLM outputs on O{*}NET tasks would be accepted by workers with relevant experience.
They report two quantities by job family: the share of a family's O{*}NET tasks that an LLM screen judges to offer at least 10\% potential time savings \citep[Figure~2a]{mertens2026crashing}, and the share of model responses rated at least minimally sufficient \citep[Table~2]{mertens2026crashing}.
The second is measured only on tasks that cleared the first screen, so the two multiply to the share of all of a family's tasks that LLMs complete without human edits.
That product is what we plot.
Because it is published by job family only, each occupation takes its family's value, and we standardize across the 682 matched occupations; the bars are therefore relative, and identical in both panels.
Acceptance rates pool 41 models with 2023--2025 release dates, so the measure describes an average model from that pool and understates frontier capability, which \citet{mertens2026crashing} put at roughly 70\% success in 2025-Q3 against 60\% pooled.
The ranking follows AIOE (Spearman $\rho = 0.91$ across the 22 families).
The clearest disagreement is legal occupations, which AIOE ranks second most exposed but the evaluations place tenth: legal tasks screen as highly addressable (85\%) yet draw the lowest acceptance rate in the data (49.8\%), so 42\% are completed at acceptable quality.
On its absolute scale the product runs from 12\% of tasks in construction and extraction to 52\% in computer and mathematical occupations. The spread comes almost entirely from addressability. It varies by a factor of five across families, whereas the acceptance rate stays between 49.8\% and 69.0\%.


\begin{figure}[H]
  \centering
  \includegraphics[width=\textwidth]{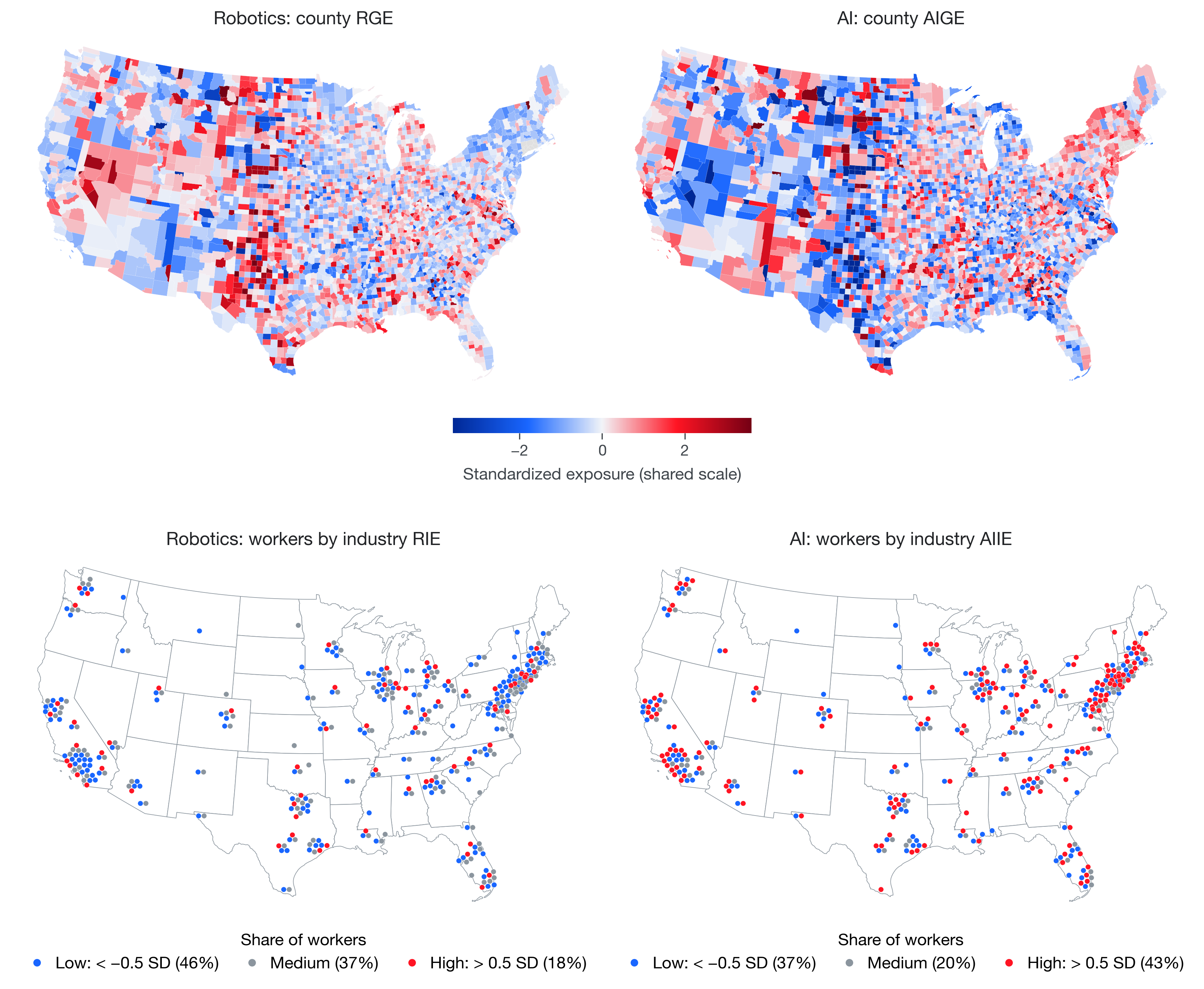}
  \caption{Geographic robotics and generative-AI exposure.
  Top: county-level robotics exposure (RGE) and generative-AI geographic exposure (AIGE) on a shared color scale capped at the 1st and 99th percentiles, with equal weight per county.
  Bottom: each dot represents 250,000 workers (2019 QCEW), colored by whether the worker's industry has low (below $-0.5$ SD), medium, or high (above $0.5$ SD) robotics exposure (RIE, left) or AI exposure (AIIE, right); legend shares are the fraction of matched workers in each class.
  Dots are allocated and placed as in Figure~\ref{fig:rge-dots}.}
  \label{fig:rge-aige}
\end{figure}

Figure~\ref{fig:rge-aige} maps county-level RGE and AIGE from \citet{felten2024generative} in the top row.
The bottom row shows the corresponding dot-density maps, built as in Figure~\ref{fig:rge-dots}.
Only 18\% of workers are in high robotics-exposure industries, compared with 43\% in high AI-exposure industries.
The two measures diverge most in large coastal cities: in New York City, 62\% of workers are in low robotics-exposure industries and 58\% in high AI-exposure industries, with similar patterns in Washington, D.C., Boston, and the San Francisco Bay Area.

\subsubsection{Comparison of correlates}
\label{sec:results-ai-correlates}

\begin{figure}[H]
  \centering
  \begin{subfigure}[t]{0.32\textwidth}
    \centering
    \includegraphics[width=\linewidth]{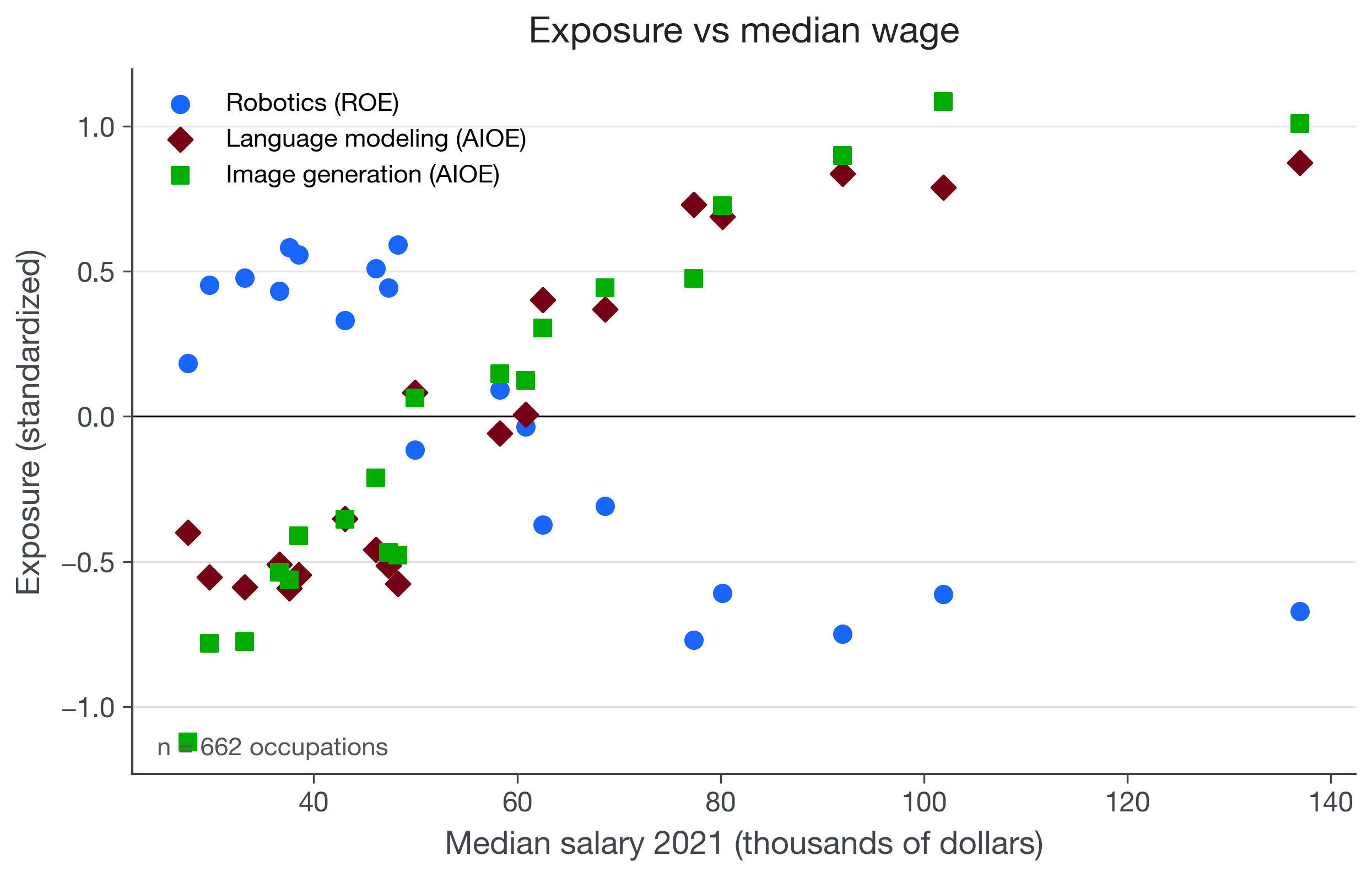}
    \caption{Median wage}
    \label{fig:roe-aioe-wage}
  \end{subfigure}\hfill
  \begin{subfigure}[t]{0.32\textwidth}
    \centering
    \includegraphics[width=\linewidth]{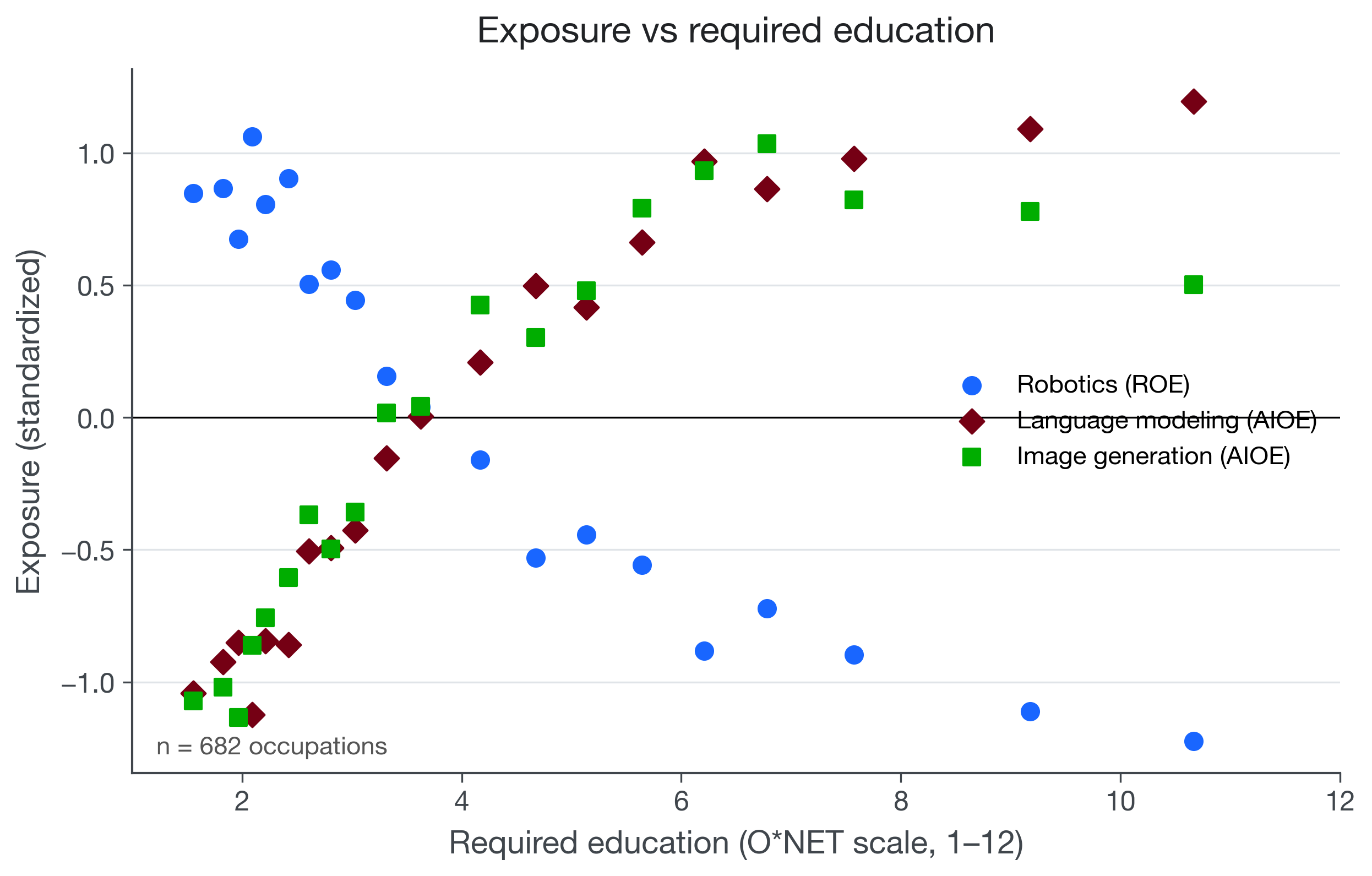}
    \caption{Required education}
    \label{fig:roe-aioe-education}
  \end{subfigure}\hfill
  \begin{subfigure}[t]{0.32\textwidth}
    \centering
    \includegraphics[width=\linewidth]{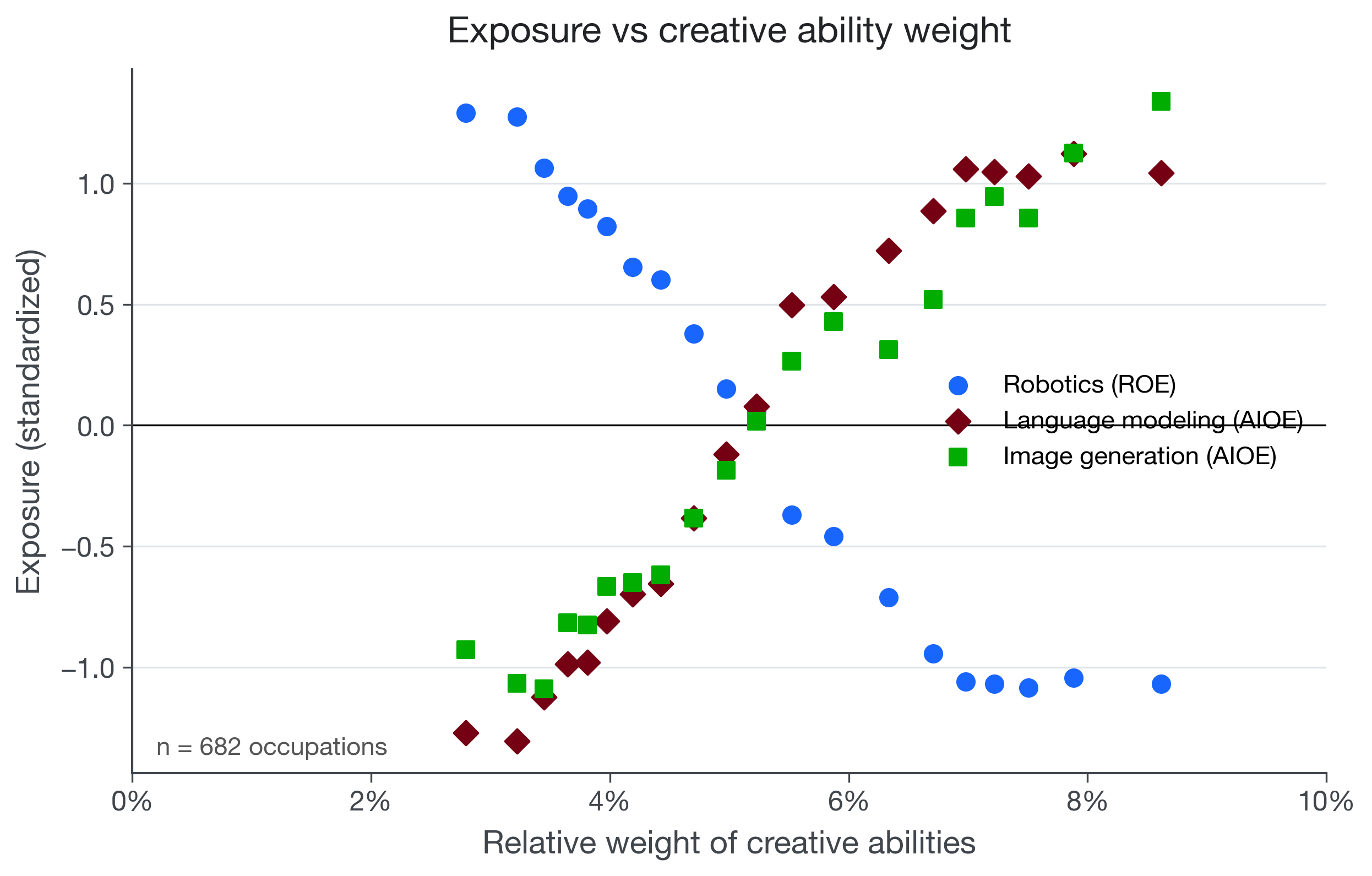}
    \caption{Creative-ability weight}
    \label{fig:roe-aioe-creative}
  \end{subfigure}
  \caption{Robotics ROE, language-modeling AIOE, and image-generation AIOE versus wage, education, and creative ability.
  Required education is on the O{*}NET 1--12 scale (2 = high school diploma, 4 = some college, 6 = bachelor's, 8 = master's, 11 = doctorate; full list in footnote~\ref{fn:required-education}).
  Creative-ability weight is defined in footnote~\ref{fn:creative-weight}.}
  \label{fig:roe-aioe-correlates-main}
\end{figure}

Figure~\ref{fig:roe-aioe-correlates-main} overlays robotics ROE with language-modeling and image-generation AIOE on the same covariate bins.
Mean standardized ROE and mean AIOE move in opposite directions on median salary and required education across bins; creative-ability weight aligns with lower mean ROE and higher mean AIOE.

\begin{figure}[H]
  \centering
  \begin{subfigure}[t]{0.48\textwidth}
    \centering
    \includegraphics[width=\linewidth]{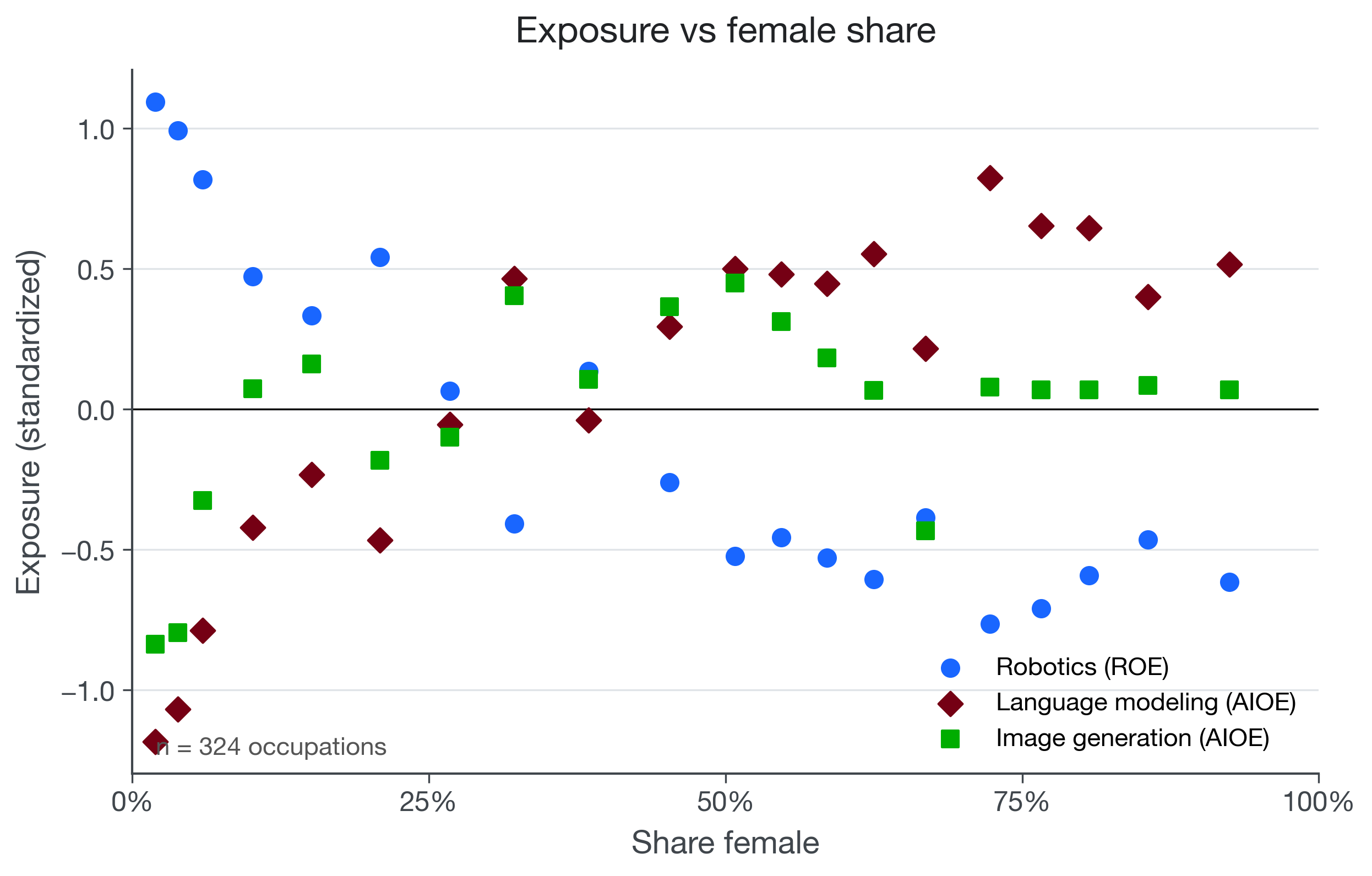}
    \caption{Female share}
    \label{fig:roe-aioe-female-share}
  \end{subfigure}\hfill
  \begin{subfigure}[t]{0.48\textwidth}
    \centering
    \includegraphics[width=\linewidth]{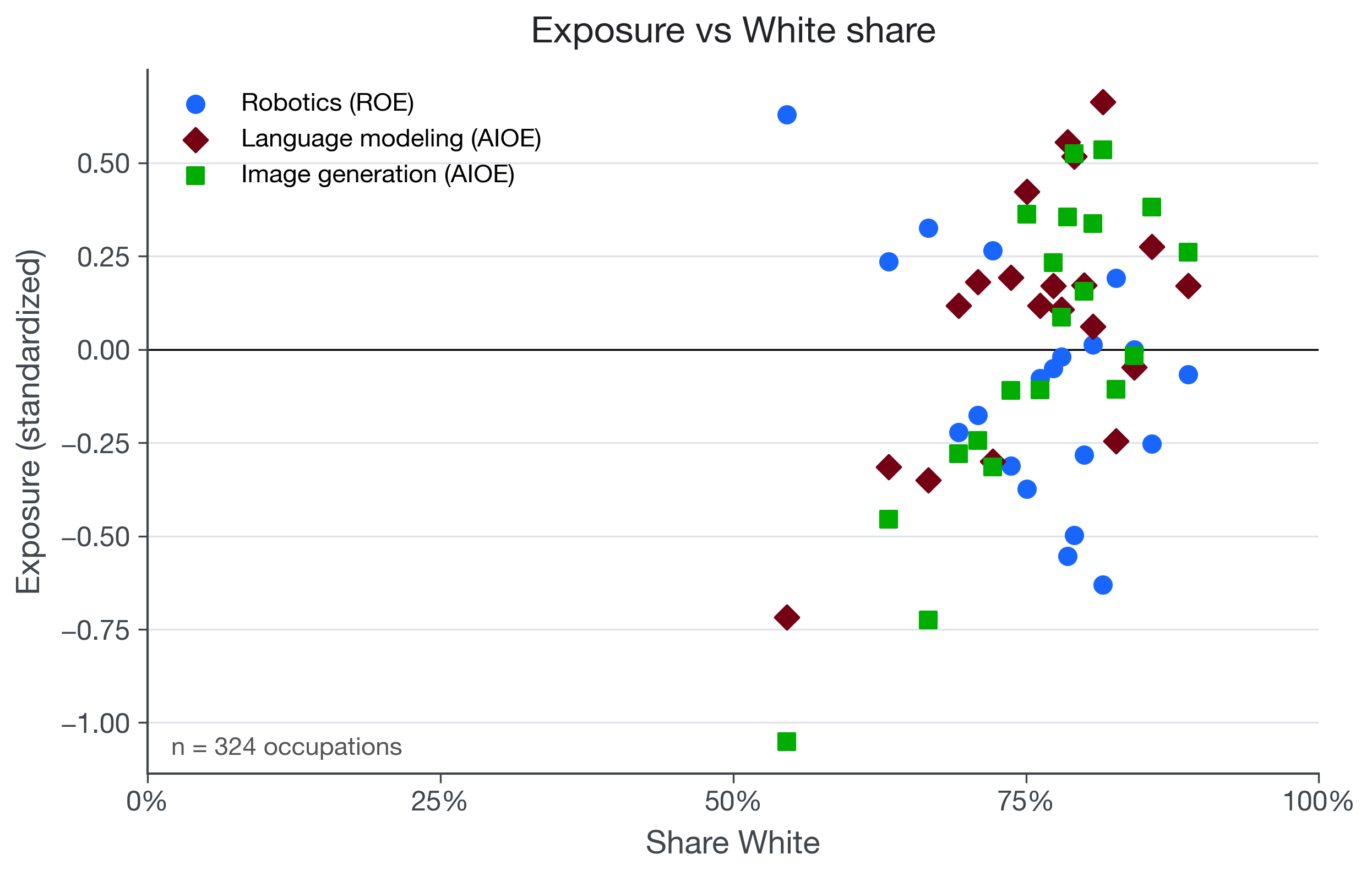}
    \caption{White share}
    \label{fig:roe-aioe-white-share}
  \end{subfigure}\\[0.75em]
  \begin{subfigure}[t]{0.48\textwidth}
    \centering
    \includegraphics[width=\linewidth]{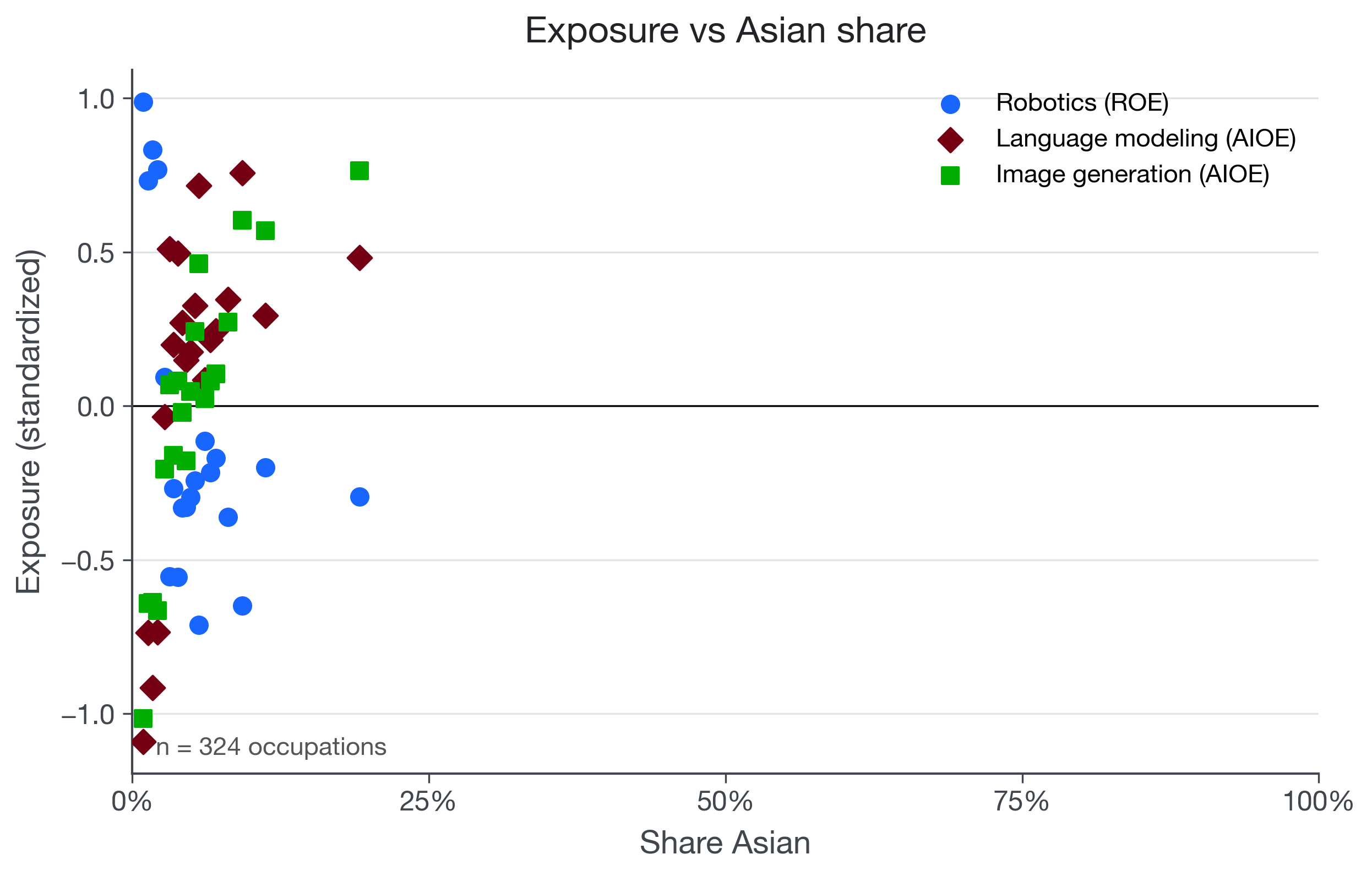}
    \caption{Asian share}
    \label{fig:roe-aioe-asian-share}
  \end{subfigure}\hfill
  \begin{subfigure}[t]{0.48\textwidth}
    \centering
    \includegraphics[width=\linewidth]{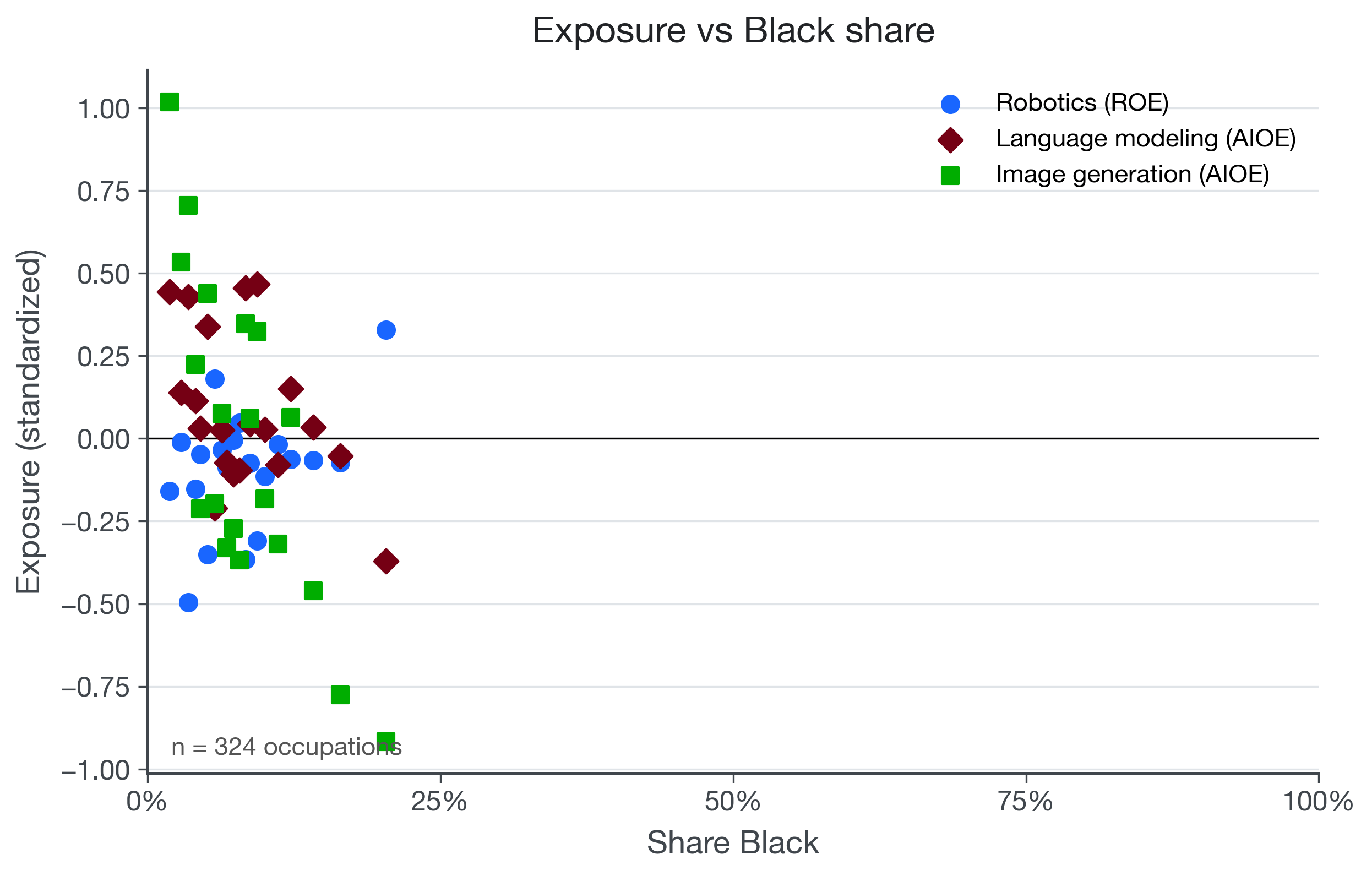}
    \caption{Black share}
    \label{fig:roe-aioe-black-share}
  \end{subfigure}
  \caption{Robotics ROE, language-modeling AIOE, and image-generation AIOE versus demographic shares.}
  \label{fig:roe-aioe-correlates-demo}
\end{figure}

Figure~\ref{fig:roe-aioe-correlates-demo} plots the three exposure series against demographic shares on the same bins.

\subsection{Which capabilities matter most for exposure?}
\label{sec:results-domains}

Our estimates so far describe how exposed occupations, industries and geographies are to robotics as a whole.
This section asks how much each capability contributes to that exposure.

We remove one capability $c$ at a time from the sum in Equation~\eqref{eq:ability-exposure}, recompute exposure from the remaining 15, and re-standardize across occupations, which gives the index that would obtain without it.
We then compare the original ROE ranking with the newly computed one, using the mean absolute change in an occupation's exposure percentile and churn in the most exposed occupations at three cut-offs.

We find that Physical Collaboration \& Assistance, Terrain-Based Locomotion, Grasping \& Handling and Task \& Functional Reasoning have the largest effect, and Tactile \& Force Sensing the smallest (Table~\ref{tab:capability-removal}).
The results also indicate that no single capability changes the ranking significantly on its own, and that robotics exposure depends rather on the breadth of capabilities.
Still, since capabilities within a domain are close substitutes, we repeat the analysis at the domain level in Appendix~\ref{app:domain-removal}, where the spread between the most and the least important is considerably larger.
Manipulation and Locomotion \& Mobility dominate in that analysis.

This analysis matters because if a capability accounts for much of that exposure and is at a low stage of technical maturity today, that is where a marginal improvement would open up the most new automation.
However, the exercise ranks capabilities by how much work each one reaches, not by how close any is to reaching it.
We also treat the capabilities as separable, so the index cannot represent a task that requires several capabilities at once, and removing one at a time ignores that technical progress in one depends on progress in the others.

% Auto-generated by 06 Capability Decomposition/decompose_roe_by_domain.py
\begin{table}[H]
  \centering
  \small
  \setlength{\tabcolsep}{5pt}
  \caption{Effect of removing each robotics capability from the exposure index. Removal drops the capability's row from the relatedness matrix and re-standardizes. $|\Delta$ pct$|$ is the mean movement of an occupation in the exposure ranking, in percentile points. Churn is the share of the most exposed 10\%, 25\% and 50\% of occupations no longer in that group after removal.}
  \label{tab:capability-removal}
  \begin{tabular}{@{}>{\raggedright\arraybackslash}p{0.30\linewidth}>{\raggedright\arraybackslash}p{0.255\linewidth}cccc@{}}
    \toprule
    Capability & Domain & $|\Delta$ pct$|$ & 10\% & 25\% & 50\% \\
    \midrule
  Physical Collaboration \& Assistance & Human--Robot Interaction & 2.5 & 12\% & 6\% & 4\% \\
  Terrain-Based Locomotion & Locomotion \& Mobility & 1.9 & 12\% & 6\% & 3\% \\
  Grasping \& Handling & Manipulation & 1.5 & 6\% & 3\% & 3\% \\
  Task \& Functional Reasoning & Cognition \& Planning & 1.5 & 8\% & 2\% & 2\% \\
  Tool Use \& Process Interaction & Manipulation & 1.2 & 8\% & 3\% & 2\% \\
  System Health \& Self-Management & System Maintenance \& Autonomy & 1.2 & 5\% & 2\% & 2\% \\
  Aerial \& Aquatic Locomotion & Locomotion \& Mobility & 1.2 & 8\% & 4\% & 2\% \\
  Dexterous \& In-Hand Manipulation & Manipulation & 1.2 & 5\% & 2\% & 2\% \\
  Visual \& Environmental Perception & Perception \& Sensing & 1.2 & 8\% & 2\% & 2\% \\
  Path \& Trajectory Planning & Navigation \& Spatial Awareness & 1.1 & 4\% & 1\% & 1\% \\
  Mapping \& Localization & Navigation \& Spatial Awareness & 1.1 & 4\% & 1\% & 1\% \\
  Autonomy \& Mission Execution & System Maintenance \& Autonomy & 0.9 & 5\% & 1\% & 1\% \\
  Expressive \& Communicative Interaction & Human--Robot Interaction & 0.7 & 5\% & 1\% & 1\% \\
  Complex Object Manipulation & Manipulation & 0.7 & 4\% & 2\% & 2\% \\
  Collective Coordination & Multi-Agent \& Collective Behavior & 0.6 & 3\% & 1\% & 1\% \\
  Tactile \& Force Sensing & Perception \& Sensing & 0.4 & 3\% & 1\% & 1\% \\
    \bottomrule
  \end{tabular}
\end{table}

\section{Discussion}
\label{sec:discussion}

\subsection{Interpretation}
\label{sec:discussion-interpretation}

This paper adapts the abilities-based exposure framework from AI to physical automation.
The resulting indices ROE, RIE, and RGE describe where work could be affected if technical robotics capabilities were available, not where robots are installed today or where automation is currently economical. As emphasized in the introduction, ROE should therefore be read as an upper bound on technical relatedness, not a forecast of displacement.

Occupational patterns line up with the physical content of the capability taxonomy.
Standardized ROE is highest in job families whose O{*}NET profiles emphasize manual handling, movement in unstructured settings, and routine physical operations such as construction, farming, transportation, and production-related work and lowest in families centered on abstract reasoning, interpersonal coordination, and creative judgment. Employment weighting shifts levels within families but preserves this broad ordering.
No single capability drives this pattern: removing any one of the 16 leaves the occupational ranking almost unchanged (Section~\ref{sec:results-domains}).
Whole domains do matter, because capabilities within a domain substitute for one another, in particular Manipulation and Locomotion \& Mobility (Appendix~\ref{app:domain-removal}).

Descriptive correlates reinforce that robotics exposure affects different occupations, industries, and demographic groups than AI exposure.
Binned means of standardized ROE fall with median wage, required education, and creative-ability weight, whereas language-modeling AIOE tends to rise on those same axes in our overlay figures \citep{felten2024generative}.
Robotics exposure is therefore skewed toward occupations with lower measured formal requirements and less creative-ability content in O{*}NET, relative to AI exposure on the matched sample.

At the industry level, employment-weighted RIE (Table~\ref{tab:rie-ind-rankings}) reallocates the same occupational scores through May OEWS occupation--industry employment.
Rankings rise where high-ROE work is concentrated in the industry mix such as logging, truck transportation, mining, construction and specialty trade contracting, crop and textile production, warehousing, and related transportation support and fall where employment is dominated by finance, insurance, professional and business services, and education.

County-level RGE and published AIGE \citep{felten2024generative} show that robotics exposure tracks goods movement, building, and production-weighted activity, whereas generative-AI geographic exposure tracks cognitive and professional services concentration.

\subsection{Limitations}
\label{sec:discussion-limitations}

Two limitations bound how the indices should be used.
First, relatedness scores come from LLM judgments over a fixed prompt and four evaluation personas (Appendix~\ref{app:validation-personas}).
Persona-level matrices agree closely and yield nearly identical occupation rankings, but they remain model- and prompt-dependent; we report validation checks in the appendix rather than treating scores as ground truth.
\citet{lyu2026right} show that LLM judges can reproduce human orderings while placing scores at the wrong level, which our own survey reproduces (Appendix~\ref{app:validation-survey}): ROE should therefore be read as a ranking, not as an absolute level of exposure.

Second, the robotics taxonomy reflects 78 curated sources and structured consolidation choices (Section~\ref{sec:method-taxonomy}).
Different source sets or condensation rules would shift capability selection and definition, and thus ROE, even holding O{*}NET fixed.

\subsection{Future Work}
\label{sec:discussion-extensions}

Several extensions would sharpen the economic interpretation of these indices.
First, the taxonomy distinguishes capabilities and form factors; repeating the exposure pipeline by form factor (for example industrial arms, mobile platforms, or humanoids) would show which embodiments align most strongly with U.S.\ occupational and industry structure.
Second, relatedness exposure can be combined with measures of technical readiness (hardware, software, and model performance), cost competitiveness relative to incumbent production, and regulatory or safety constraints, following the decomposition emphasized in Section~\ref{sec:method-formulas}.
Third, linking exposure to firm- or establishment-level panels on technology adoption would help separate where work is technically automatable from where firms actually deploy robotics.

\section{Conclusion}
\label{sec:conclusion}

We introduce occupational, industry, and geographic exposure measures for robotics built in parallel to published AI exposure indices for occupations, industries, and places.
A source-grounded capability taxonomy and LLM-scored relatedness matrices link robotics domains to O{*}NET work abilities; aggregation yields ROE for 774 six-digit occupations, employment-weighted RIE for NAICS industries, and county-level RGE.

Four findings stand out.
First, robotics exposure concentrates in physically intensive job families and industry mixes, not in the cognitive and creative profiles where generative-AI exposure is highest.
Second, descriptive correlates place higher ROE in occupations with lower median wages, lower measured education requirements, and lower creative-ability weights on O{*}NET, patterns that differ from generative-AIOE overlays on the same covariates.
Third, on occupations with harmonized SOC codes, ROE and AIOE are strongly negatively related and imply different geographic hotspots when mapped to workers through industry structure.
Fourth, no single robotics capability carries the index, though whole domains do: removing any one of the 16 capabilities leaves the occupational ranking almost unchanged, while dropping all of manipulation moves it substantially.

These measures are intended to help researchers and policymakers characterize where physical automation could matter for work and to inform research and policy on robotics alongside existing AI exposure tools.

\section{Acknowledgments}
\label{sec:acknowledgments}

We thank Bridget Jiang for implementing the exposure measure and the initial analyses, and Julia Wang for her work on validation, both as part of the MIT Undergraduate Research Opportunities Program (UROP).

\bibliographystyle{plainnat}
\bibliography{references}

\clearpage
\appendix

\section{Appendix}
\label{app:main}

\subsection{Robotics capability taxonomy: sources and LLM prompt}
\label{app:taxonomy-sources}

This appendix documents the source base and language-model prompts used to build the robotics capability taxonomy (Section~\ref{sec:method-taxonomy}).

The full list of 78 documents used in source collection is maintained in an
\href{https://docs.google.com/spreadsheets/d/1KUM-T7XB89sgD2sWI2NM3PVa_UCIIEvg9C0LVM5r4vU/edit?usp=sharing}{online taxonomy source spreadsheet}.

The following per-source extraction prompt (prompt 1) was applied to each source PDF individually.

\begin{promptverbatim}
You will be given a single article, report, or similar document about robotics. Your task is to extract how that source categorizes robotics --- based strictly on what the source actually says, not on your own knowledge of robotics.
Extract categories along up to three dimensions:
Form factor --- what the robot physically is (e.g., industrial arm, mobile robot, humanoid, drone, quadruped)
Capability --- what the robot does (e.g., locomotion, manipulation, inspection, navigation)
Sector --- the industry or application area it's used in (e.g., agriculture, healthcare, manufacturing, logistics)
Strictness rules (these are the most important part):
Only extract categories explicitly stated in the source. Do not add, infer, generalize, or fill in a category that seems like it logically belongs, however obvious it feels.
If the source does not address one of the three dimensions, leave that dimension empty for this source --- do not guess.
Extract a term only if the source uses it as a category or type of robot/robotics (a way of grouping or classifying), not merely a passing mention. If a source says "robots are used in hospitals" that is a sector category (healthcare); if it just mentions a hospital in an anecdote without treating it as a class of application, do not extract it.
Use the source's own wording, lightly normalized (e.g., singular, lowercase) --- do not translate a specific term into a broader one. If the source says "collaborative robot," record "collaborative robot," not "industrial robot."
If one term could fit two dimensions, place it where the source frames it. If the framing is genuinely ambiguous, place it in the single best-fit dimension and note the ambiguity in brackets. Do not list the same term in two dimensions.
Output format --- output exactly one row, matching this structure, for the source provided:
| Source Name | Source Link | Capabilities | Form Factors | Applications |
Within the Capabilities, Form Factors, and Applications cells, separate multiple categories with a semicolon and a space (; ).
If a dimension has no categories, leave that cell empty (do not write "none" or "N/A").
Output only the row (and a header row). Do not add commentary, explanation, or your own categorization suggestions.
If nothing in any of the three dimensions is present in the source, say so in one line instead of emitting a row.
\end{promptverbatim}

Only the capability lens feeds the main index: its output defines the 16 domains in Table~\ref{tab:capabilities}.
We disregarded the form-factor and application-sector lenses for ROE construction.

The following condensation prompt (prompt 2; application sector lens) consolidates raw application-sector extractions into domains and use-cases. Analogous prompts were applied to the capability and form-factor lenses, with JSON field names adjusted to match each lens.

\begin{promptverbatim}
You are consolidating a taxonomy of [CATEGORY] extracted from 78 robotics review papers. The raw list mixes broad sectors ("healthcare") with narrow use-cases ("throat swab sampling"), plus synonyms and some non-application entries.
Your job: produce a clean, two-level taxonomy of application domains for tagging and analysis across papers.
RAW ITEMS:
{items}
RULES:
1. Two levels: ~10-15 top-level DOMAINS (economic sectors / broad settings), each with canonical USE-CASES underneath. No deeper than two levels. Likely domains: Manufacturing/Industrial; Logistics/Warehousing; Healthcare/Medical; Surgery; Agriculture; Construction; Domestic/Service; Defense/Military; Search & Rescue/Disaster; Space; Marine/Underwater; Inspection & Maintenance; Energy/Utilities; Retail/Hospitality; Science/Exploration.
2. Consistent granularity. Collapse hyper-specific procedures into their use-case or domain (e.g., "throat swab sampling", "venous blood collection" -> clinical nursing tasks under Healthcare). Keep a use-case only if papers would meaningfully distinguish it.
3. Distinguish DOMAIN (where/what sector) from the capability or form factor used --- drop anything that is actually a capability or embodiment.
4. Some medical sub-areas are large enough to be their own domain (e.g., Surgery, with its many specialties as use-cases). Use judgment; note it.
5. Merge synonyms to one canonical term; list merged raw terms for auditability.
6. Drop non-application entries (URLs, bracketed editorial notes, pure capability/form terms); list with one-line reasons. Every remaining term maps to exactly one node.
OUTPUT as JSON:
{
  "taxonomy": [
    { "domain": "...", "definition": "one line",
      "use_cases": [
        { "name": "...", "definition": "one line", "merged_raw_terms": ["...","..."] }
      ] }
  ],
  "dropped": [ { "term": "...", "reason": "..." } ],
  "notes": "ambiguous items and judgment calls for human review"
}
\end{promptverbatim}

\subsection{O{*}NET work abilities}
\label{app:onet-abilities}

The relatedness matrix and ROE index use the 52 O{*}NET~30.3 work abilities.
They span four content-model categories: cognitive, physical, psychomotor, and sensory.

\medskip
% Auto-generated by 100 Paper Scripts/build_onet_abilities_table.py
\begingroup
\footnotesize
\setlength{\tabcolsep}{4pt}
\setlength{\LTpre}{0.35em}
\setlength{\LTpost}{0.35em}
\setlength{\LTleft}{0pt}
\setlength{\LTright}{0pt}
\renewcommand{\arraystretch}{1.12}
\begin{longtable}{@{}>{\raggedright\arraybackslash}p{0.21\textwidth}>{\raggedright\arraybackslash}p{0.10\textwidth}r>{\raggedright\arraybackslash}p{0.58\textwidth}@{}}
  \caption{O{*}NET 30.3 work abilities used in the relatedness matrix and ROE index (52 abilities). Sorted by ability-level relatedness $R_a$, the sum of normalized relatedness across the 16 robotics capabilities (range 0--16). Definitions follow the O{*}NET Content Model.} \label{tab:onet-abilities} \\
  \toprule
  Ability & Category & $R_a$ & Definition \\
  \midrule
  \endfirsthead
  \multicolumn{4}{l}{\textit{(continued)}} \\
  \toprule
  Ability & Category & $R_a$ & Definition \\
  \midrule
  \endhead
  \midrule
  \multicolumn{4}{r}{\textit{(continued on next page)}} \\
  \endfoot
  \bottomrule
  \endlastfoot
  Spatial Orientation & Cognitive & 11.75 & The ability to know your location in relation to the environment or to know where other objects are in relation to you. \\
  Speed of Closure & Cognitive & 8.44 & The ability to quickly make sense of, combine, and organize information into meaningful patterns. \\
  Depth Perception & Sensory & 8.35 & The ability to judge which of several objects is closer or farther away from you, or to judge the distance between you and an object. \\
  Problem Sensitivity & Cognitive & 7.80 & The ability to tell when something is wrong or is likely to go wrong. It does not involve solving the problem, only recognizing that there is a problem. \\
  Control Precision & Psychomotor & 7.39 & The ability to quickly and repeatedly adjust the controls of a machine or a vehicle to exact positions. \\
  Response Orientation & Psychomotor & 7.12 & The ability to choose quickly between two or more movements in response to two or more different signals (lights, sounds, pictures). It includes the speed with which the correct response is started with the hand, foot, or other body part. \\
  Manual Dexterity & Psychomotor & 7.00 & The ability to quickly move your hand, your hand together with your arm, or your two hands to grasp, manipulate, or assemble objects. \\
  Arm-Hand Steadiness & Psychomotor & 6.76 & The ability to keep your hand and arm steady while moving your arm or while holding your arm and hand in one position. \\
  Rate Control & Psychomotor & 6.74 & The ability to time your movements or the movement of a piece of equipment in anticipation of changes in the speed and/or direction of a moving object or scene. \\
  Information Ordering & Cognitive & 6.50 & The ability to arrange things or actions in a certain order or pattern according to a specific rule or set of rules (e.g., patterns of numbers, letters, words, pictures, mathematical operations). \\
  Gross Body Coordination & Physical & 6.32 & The ability to coordinate the movement of your arms, legs, and torso together when the whole body is in motion. \\
  Category Flexibility & Cognitive & 6.04 & The ability to generate or use different sets of rules for combining or grouping things in different ways. \\
  Deductive Reasoning & Cognitive & 5.75 & The ability to apply general rules to specific problems to produce answers that make sense. \\
  Visualization & Cognitive & 5.40 & The ability to imagine how something will look after it is moved around or when its parts are moved or rearranged. \\
  Multilimb Coordination & Psychomotor & 5.20 & The ability to coordinate two or more limbs (for example, two arms, two legs, or one leg and one arm) while sitting, standing, or lying down. It does not involve performing the activities while the whole body is in motion. \\
  Inductive Reasoning & Cognitive & 5.16 & The ability to combine pieces of information to form general rules or conclusions (includes finding a relationship among seemingly unrelated events). \\
  Dynamic Strength & Physical & 5.09 & The ability to exert muscle force repeatedly or continuously over time. This involves muscular endurance and resistance to muscle fatigue. \\
  Speed of Limb Movement & Psychomotor & 4.99 & The ability to quickly move the arms and legs. \\
  Dynamic Flexibility & Physical & 4.94 & The ability to quickly and repeatedly bend, stretch, twist, or reach out with your body, arms, and/or legs. \\
  Extent Flexibility & Physical & 4.89 & The ability to bend, stretch, twist, or reach with your body, arms, and/or legs. \\
  Reaction Time & Psychomotor & 4.76 & The ability to quickly respond (with the hand, finger, or foot) to a signal (sound, light, picture) when it appears. \\
  Finger Dexterity & Psychomotor & 4.51 & The ability to make precisely coordinated movements of the fingers of one or both hands to grasp, manipulate, or assemble very small objects. \\
  Perceptual Speed & Cognitive & 4.39 & The ability to quickly and accurately compare similarities and differences among sets of letters, numbers, objects, pictures, or patterns. The things to be compared may be presented at the same time or one after the other. This ability also includes comparing a presented object with a remembered object. \\
  Flexibility of Closure & Cognitive & 4.31 & The ability to identify or detect a known pattern (a figure, object, word, or sound) that is hidden in other distracting material. \\
  Selective Attention & Cognitive & 4.31 & The ability to concentrate on a task over a period of time without being distracted. \\
  Near Vision & Sensory & 4.14 & The ability to see details at close range (within a few feet of the observer). \\
  Far Vision & Sensory & 4.01 & The ability to see details at a distance. \\
  Gross Body Equilibrium & Physical & 3.98 & The ability to keep or regain your body balance or stay upright when in an unstable position. \\
  Mathematical Reasoning & Cognitive & 3.98 & The ability to choose the right mathematical methods or formulas to solve a problem. \\
  Time Sharing & Cognitive & 3.89 & The ability to shift back and forth between two or more activities or sources of information (such as speech, sounds, touch, or other sources). \\
  Wrist-Finger Speed & Psychomotor & 3.86 & The ability to make fast, simple, repeated movements of the fingers, hands, and wrists. \\
  Peripheral Vision & Sensory & 3.85 & The ability to see objects or movement of objects to one's side when the eTrue are looking ahead. \\
  Static Strength & Physical & 3.51 & The ability to exert maximum muscle force to lift, push, pull, or carry objects. \\
  Visual Color Determination & Sensory & 3.12 & The ability to match or detect differences between colors, including shades of color and brightness. \\
  Glare Sensitivity & Sensory & 2.84 & The ability to see objects in the presence of a glare or bright lighting. \\
  Memorization & Cognitive & 2.69 & The ability to remember information such as words, numbers, pictures, and procedures. \\
  Originality & Cognitive & 2.17 & The ability to come up with unusual or clever ideas about a given topic or situation, or to develop creative ways to solve a problem. \\
  Stamina & Physical & 2.08 & The ability to exert yourself physically over long periods of time without getting winded or out of breath. \\
  Trunk Strength & Physical & 2.08 & The ability to use your abdominal and lower back muscles to support part of the body repeatedly or continuously over time without "giving out" or fatiguing. \\
  Oral Comprehension & Cognitive & 2.07 & The ability to listen to and understand information and ideas presented through spoken words and sentences. \\
  Night Vision & Sensory & 2.04 & The ability to see under low-light conditions. \\
  Explosive Strength & Physical & 1.79 & The ability to use short bursts of muscle force to propel oneself (as in jumping or sprinting), or to throw an object. \\
  Speech Recognition & Sensory & 1.69 & The ability to identify and understand the speech of another person. \\
  Number Facility & Cognitive & 1.57 & The ability to add, subtract, multiply, or divide quickly and correctly. \\
  Hearing Sensitivity & Sensory & 1.55 & The ability to detect or tell the differences between sounds that vary in pitch and loudness. \\
  Oral Expression & Cognitive & 1.54 & The ability to communicate information and ideas in speaking so others will understand. \\
  Sound Localization & Sensory & 1.52 & The ability to tell the direction from which a sound originated. \\
  Auditory Attention & Sensory & 1.50 & The ability to focus on a single source of sound in the presence of other distracting sounds. \\
  Speech Clarity & Sensory & 1.25 & The ability to speak clearly so others can understand you. \\
  Fluency of Ideas & Cognitive & 1.21 & The ability to come up with a number of ideas about a topic (the number of ideas is important, not their quality, correctness, or creativity). \\
  Written Comprehension & Cognitive & 1.06 & The ability to read and understand information and ideas presented in writing. \\
  Written Expression & Cognitive & 0.31 & The ability to communicate information and ideas in writing so others will understand. \\
\end{longtable}
\endgroup

\medskip

\subsection{Relatedness matrix: LLM prompt and personas}
\label{app:matrix-prompt}

For O{*}NET-based scoring, the template labels the work side as an occupational requirement; each row uses the O{*}NET ability name, category, and definition from Table~\ref{tab:onet-abilities}. The system prompt below was filled per capability--ability pair with the capability fields from Table~\ref{tab:capabilities}.

\begin{promptverbatim}
You are answering from the perspective of a {persona}.
How you see the world:
{persona_description}
Your task is to assess the RELATEDNESS between a robotics capability and an occupational requirement.
Robotics capability:
Name: {capability_name}
Definition: {capability_definition}
Example terms: {capability_examples}
Occupational requirement:
Category: {requirement_category}
Requirement: {requirement_text}
Core question:
To what extent would exercising this robotics capability involve directly engaging with or responding to this occupational requirement?
Important framing:
- Score RELATEDNESS, not current technical feasibility.
- Assume the robot possesses the stated capability.
- Do not judge whether present-day robots can perform it at a human level.
- Do not consider cost, maturity, adoption, reliability, regulation, or deployment practicality.
- Do not estimate automation or worker replacement.
- Focus only on the functional relationship between the capability and requirement.
- Give a low score when the connection is incidental or based only on both occurring in the same workplace.
Scale (integer 0-4):
0 = Unrelated: no functional connection.
1 = Weakly related: touches the requirement only indirectly or rarely.
2 = Moderately related: engages the requirement in some but not most uses.
3 = Clearly related: engaging this requirement is a normal part of exercising the capability.
4 = Strongly related: exercising the capability directly and centrally involves this requirement.
Calibration examples (capability vs requirement -> score):
- "Grasping & Handling" vs "Fine manipulation" -> 4
- "Grasping & Handling" vs "Exposure to extreme cold" -> 0
- "Terrain-Based Locomotion" vs "Reaching overhead" -> 1
- "Visual & Environmental Perception" vs "Near vision" -> 4
Return only valid JSON, with the reason BEFORE the score:
{
  "reason": "<one or two concise sentences>",
  "score": <integer from 0 to 4>
}
\end{promptverbatim}

\begin{table}[H]
  \centering
  \small
  \caption{Evaluation personas for the relatedness matrix.}
  \label{tab:matrix-personas}
  \begin{tabular}{@{}>{\raggedright\arraybackslash}p{0.28\linewidth}>{\raggedright\arraybackslash}p{0.66\linewidth}@{}}
    \toprule
    Persona & Description \\
    \midrule
    Robotics engineer &
    Hardware and control: actuators, sensors, end effectors, and physical variables the capability must sense or regulate. \\
    Occupational therapist &
    Bodily demands of work: muscles, joints, senses, postures, and physical strain on the worker. \\
    Industrial engineer &
    Production processes, material flow, task sequences, and physical operations in a job's workflow. \\
    Workplace safety and ergonomics inspector &
    Physical hazards and load: environmental exposure, force, repetition, and strenuous or risky job elements. \\
    \bottomrule
  \end{tabular}
\end{table}

\subsection{Validation}
\label{app:validation}

\subsubsection{Persona agreement}
\label{app:validation-personas}

\begin{figure}[H]
  \centering
  \begin{subfigure}[t]{0.48\textwidth}
    \centering
    \includegraphics[width=\linewidth]{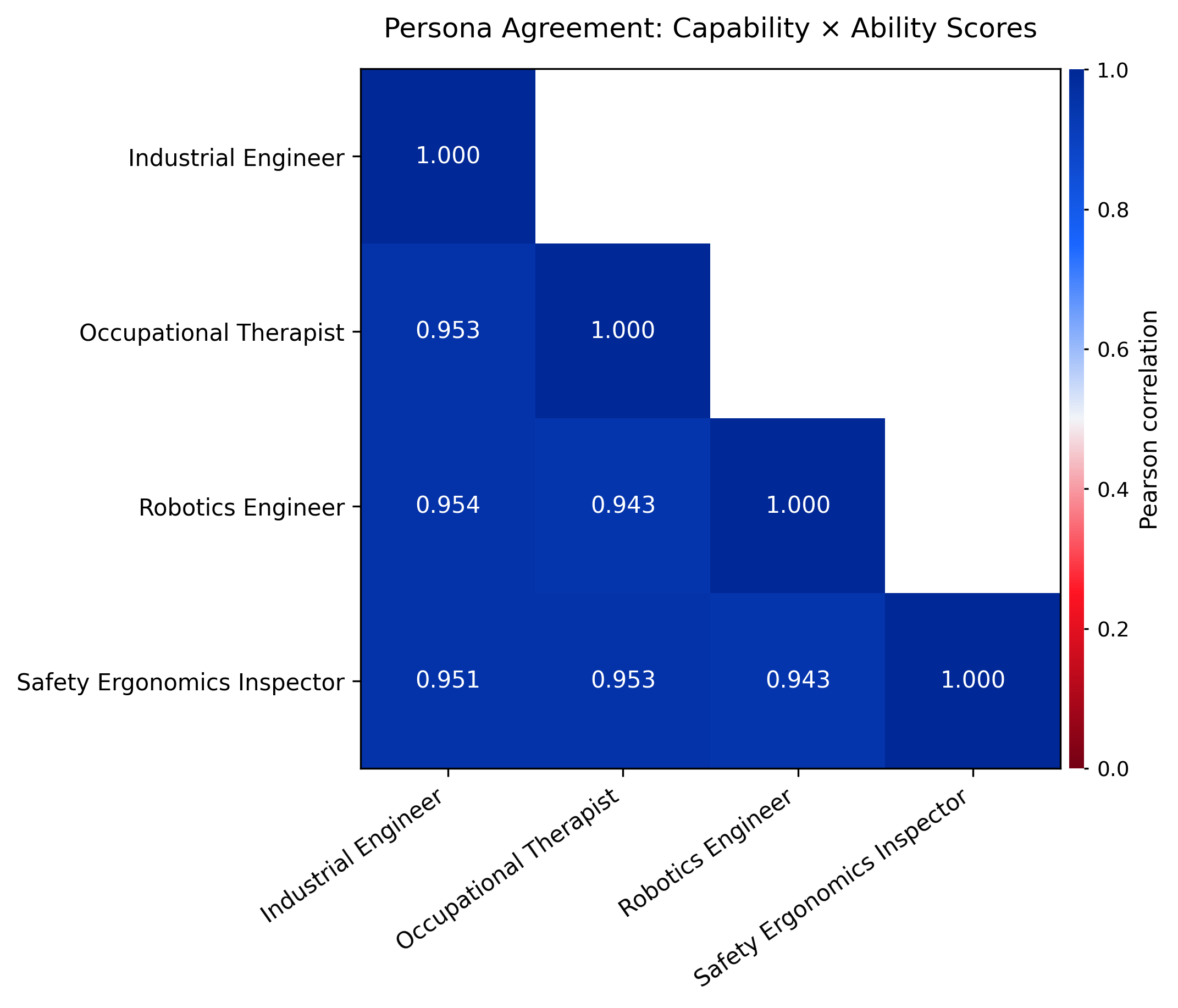}
    \caption{Capability--ability matrix scores}
    \label{fig:persona-matrix-corr}
  \end{subfigure}\hfill
  \begin{subfigure}[t]{0.48\textwidth}
    \centering
    \includegraphics[width=\linewidth]{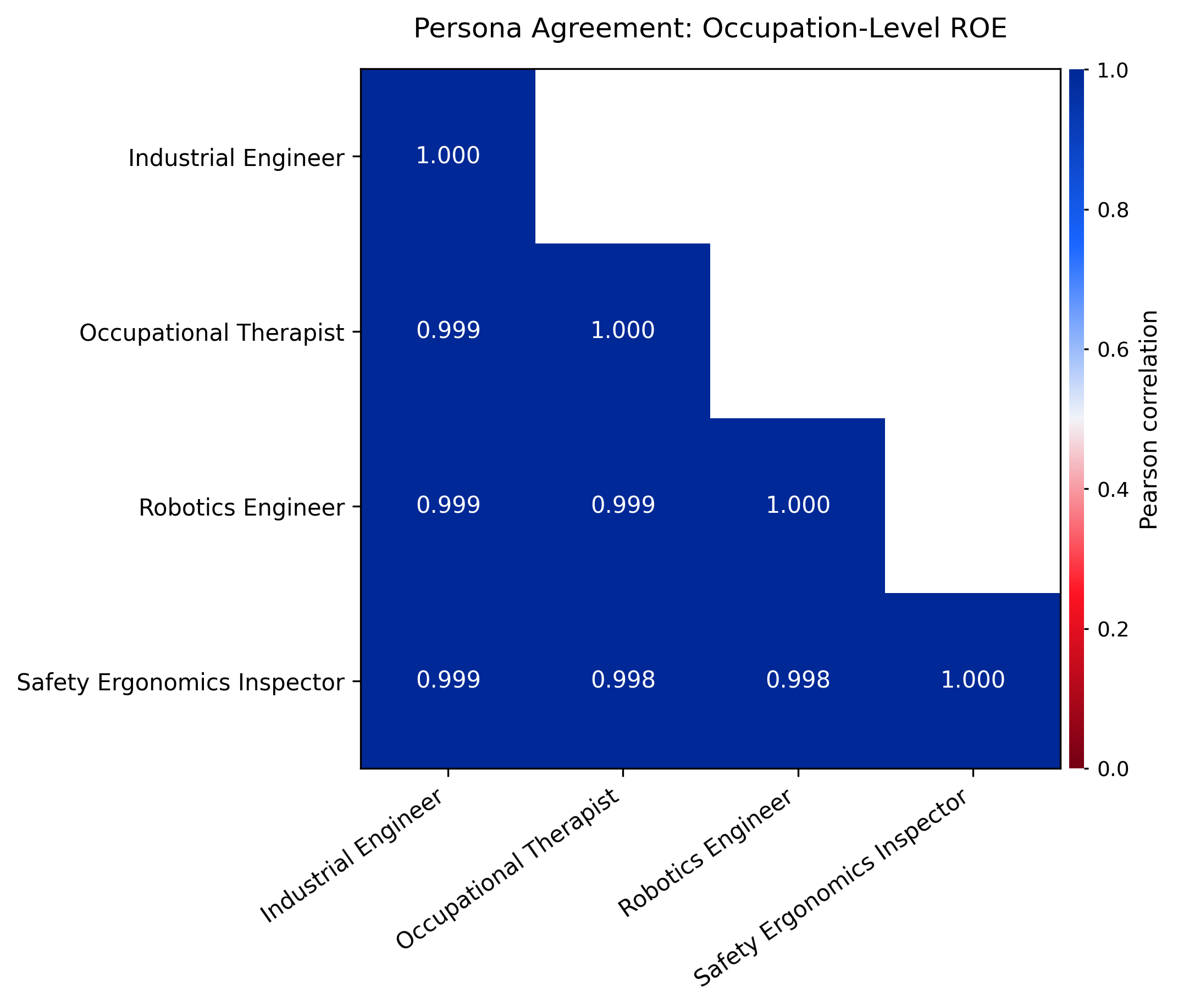}
    \caption{Occupation-level ROE}
    \label{fig:persona-occ-corr}
  \end{subfigure}
  \caption{Pearson correlations between relatedness scores from the four evaluation personas.
  Panel~(a) correlates raw matrix cells; panel~(b) correlates ROE built from each persona's matrix using the main-sample aggregation.}
  \label{fig:persona-agreement}
\end{figure}

We scored the full robotics-capability-by-O{*}NET-ability matrix separately under each evaluation persona in Table~\ref{tab:matrix-personas}.
For each persona pair we computed Pearson correlations across all matrix cells (832 scores per persona: 16 capabilities $\times$ 52 abilities) and, separately, across occupation-level ROE after applying the same Felten-style aggregation used in the main index.
Figure~\ref{fig:persona-matrix-corr} reports cell-level agreement; Figure~\ref{fig:persona-occ-corr} reports occupation-level agreement.
Pairwise cell correlations range from about 0.94 to 0.95; occupation-level ROE correlations exceed 0.998 for every pair.
No persona is clearly separated from the others on this structure: perspective shifts average ratings slightly but leaves rankings nearly unchanged.

\subsubsection{Second LLM replication (Qwen3-14B)}
\label{app:validation-qwen}

\begin{figure}[H]
  \centering
  \begin{subfigure}[t]{0.32\textwidth}
    \centering
    \includegraphics[width=\linewidth]{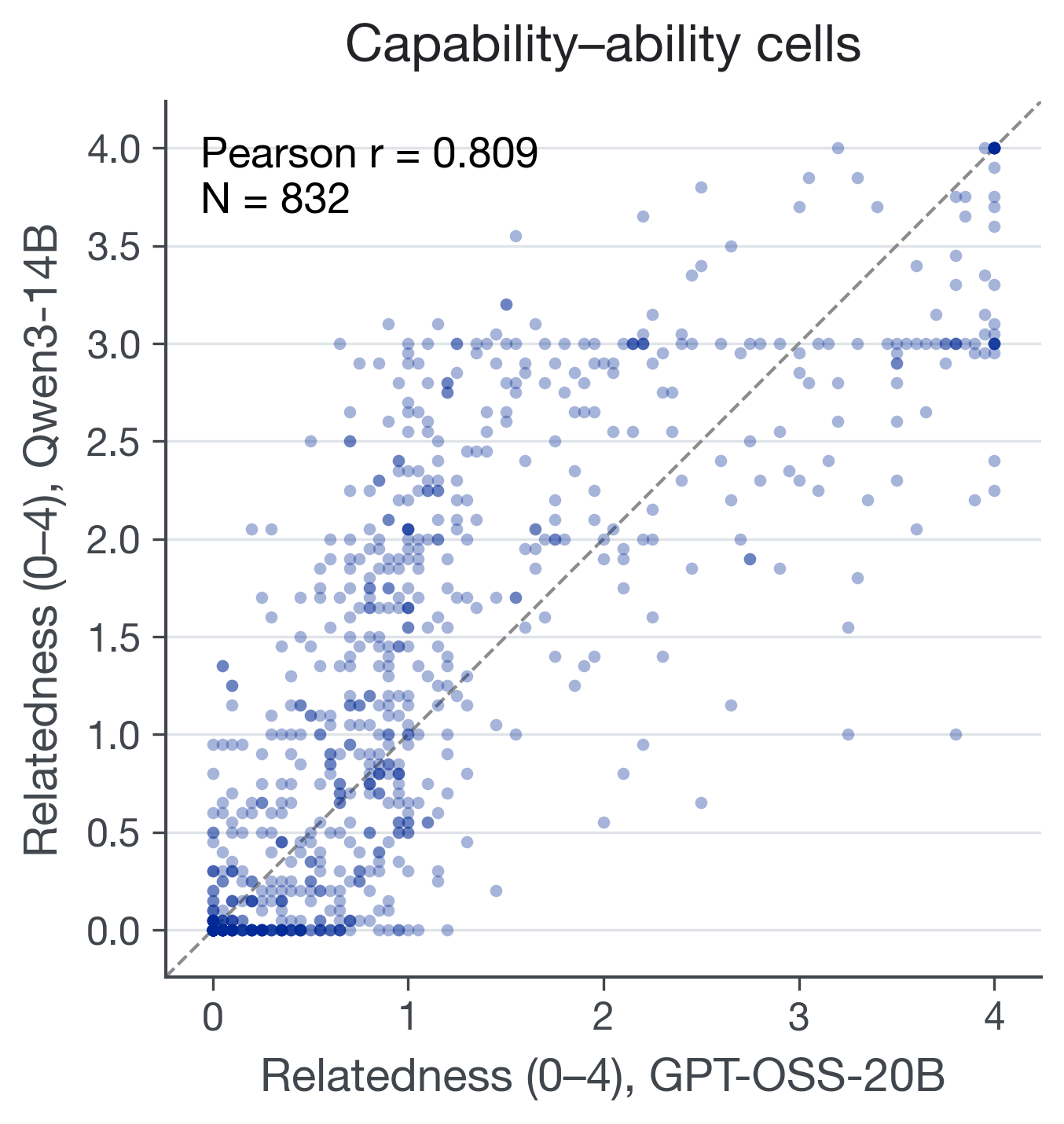}
    \caption{Matrix cells}
    \label{fig:qwen-replication-cells}
  \end{subfigure}\hfill
  \begin{subfigure}[t]{0.32\textwidth}
    \centering
    \includegraphics[width=\linewidth]{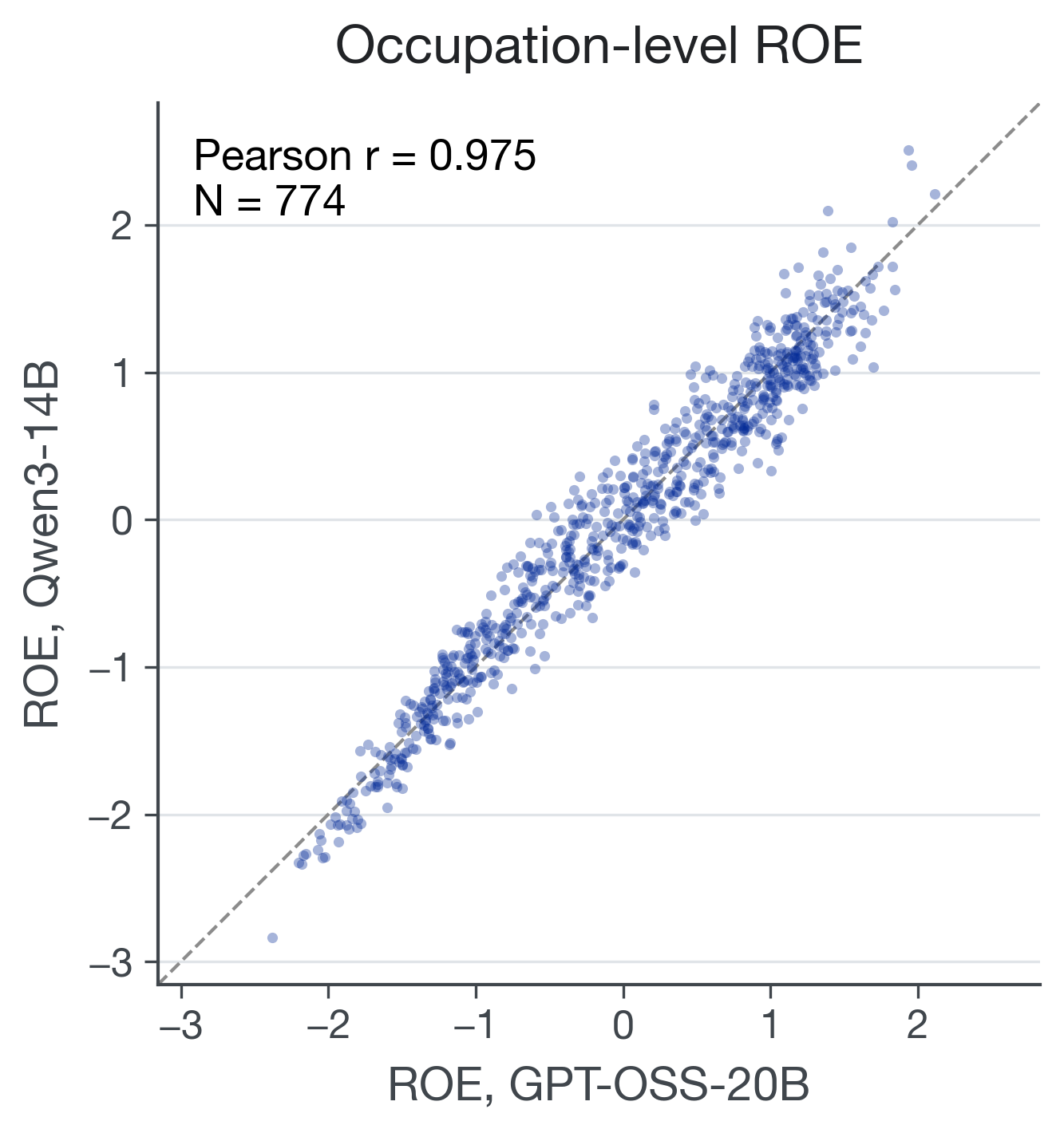}
    \caption{Occupation-level ROE}
    \label{fig:qwen-replication-roe}
  \end{subfigure}\hfill
  \begin{subfigure}[t]{0.32\textwidth}
    \centering
    \includegraphics[width=\linewidth]{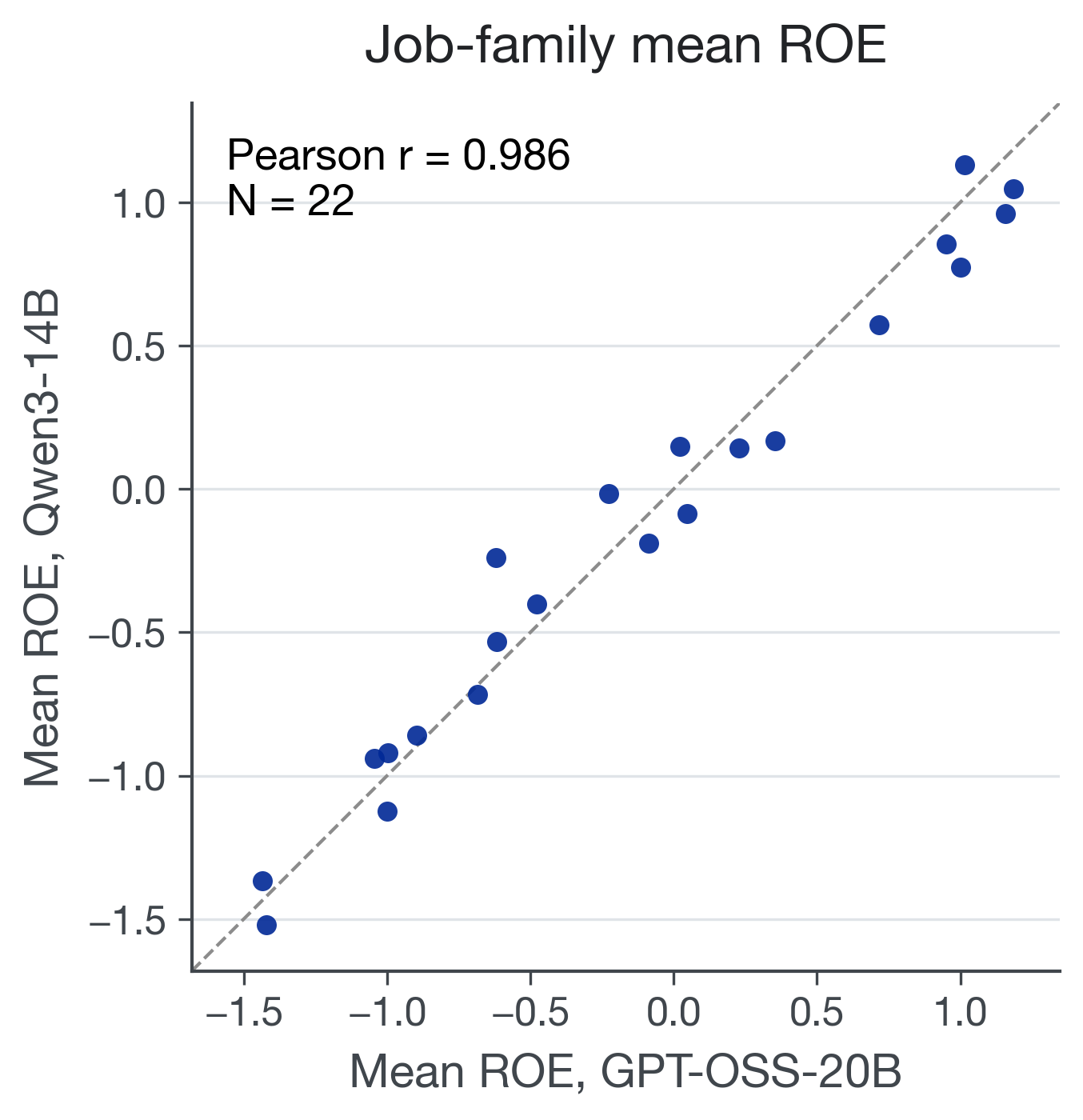}
    \caption{Job-family mean ROE}
    \label{fig:qwen-replication-family}
  \end{subfigure}
  \caption{GPT-OSS-20B versus Qwen3-14B.
  Panel~(a) compares the mean relatedness rating (0--4, averaged over personas and samples) for the 832 capability--ability pairs; panel~(b) compares standardized ROE for 774 occupations built from each model's matrix; panel~(c) compares unweighted mean standardized ROE across the 22 job families.
  The dashed line marks equality.}
  \label{fig:qwen-replication}
\end{figure}

To check that ROE does not hinge on one model, we re-scored the full capability-by-ability matrix with Qwen3-14B, using the same prompt (Appendix~\ref{app:matrix-prompt}), the same four personas, and five samples per persona, and then rebuilt ROE with the same aggregation as the main index.
Each individual response is an integer rating from 0 to 4; a cell's value is the mean of its 20 ratings (4 personas $\times$ 5 samples), so non-integer values such as 2.5 arise when the ratings for a pair are split, for example evenly between 2 and 3.
We compare the two models at three levels: the 832 matrix cells, occupation-level ROE for 774 occupations, and mean ROE for the 22 job families.
Figure~\ref{fig:qwen-replication} plots each Qwen3-14B value against its GPT-OSS-20B counterpart.
Agreement is moderate at the cell level (Pearson $r = 0.81$) and very high after aggregation: $r = 0.98$ for occupation-level ROE and $r = 0.99$ for job-family means.
Qwen3-14B rates pairs as somewhat more related on average (mean rating 1.23 versus 1.05 on the 0--4 scale), but individual cell disagreements largely cancel once scores are weighted by each occupation's ability profile, so the ranking of occupations and job families is nearly unchanged.

\begin{figure}[H]
  \centering
  \includegraphics[width=0.6\linewidth]{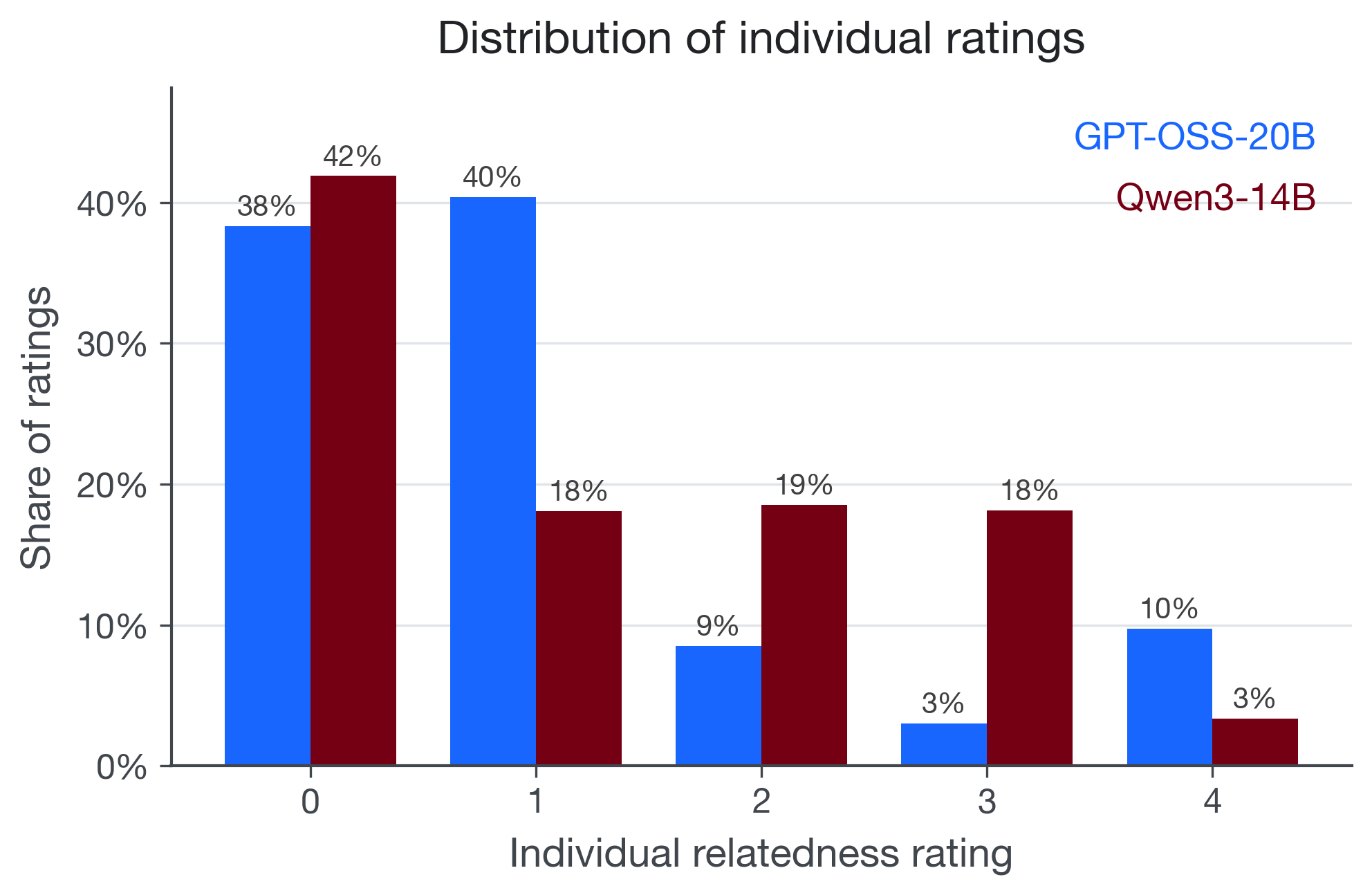}
  \caption{Distribution of individual relatedness ratings (0--4) for GPT-OSS-20B and Qwen3-14B, pooled over all 832 capability--ability pairs, four personas, and five samples (16{,}640 ratings per model).}
  \label{fig:qwen-rating-dist}
\end{figure}

Much of the cell-level disagreement reflects scale use rather than ranking (Figure~\ref{fig:qwen-rating-dist}): GPT-OSS-20B rarely gives a 3 and jumps to 4 for clearly related pairs, whereas Qwen3-14B rarely gives a 4, which produces the band at 3 in Figure~\ref{fig:qwen-replication-cells} but has little effect on ROE.

\subsubsection{Human survey versus LLM relatedness matrix}
\label{app:validation-survey}

To test whether the LLM agrees with people, we fielded a small survey on a subset of the matrix.
We drew 50 of the 832 capability--ability pairs at random, stratified by how much the model's own 20 ratings disagreed: 10 pairs rated unanimously, 20 with low spread ($0 < \text{SD} < 0.5$ rating points), and 20 with high spread ($\text{SD} \geq 0.5$).
Respondents rated each pair on the same 0--4 scale, with the same definitions given to the model and no sight of its scores.
Seven respondents each rated all 50 pairs.
We also split the same pairs post hoc by the level of the LLM mean (below 1, 1--3, 3 or above).

\begin{figure}[H]
  \centering
  \includegraphics[width=\linewidth]{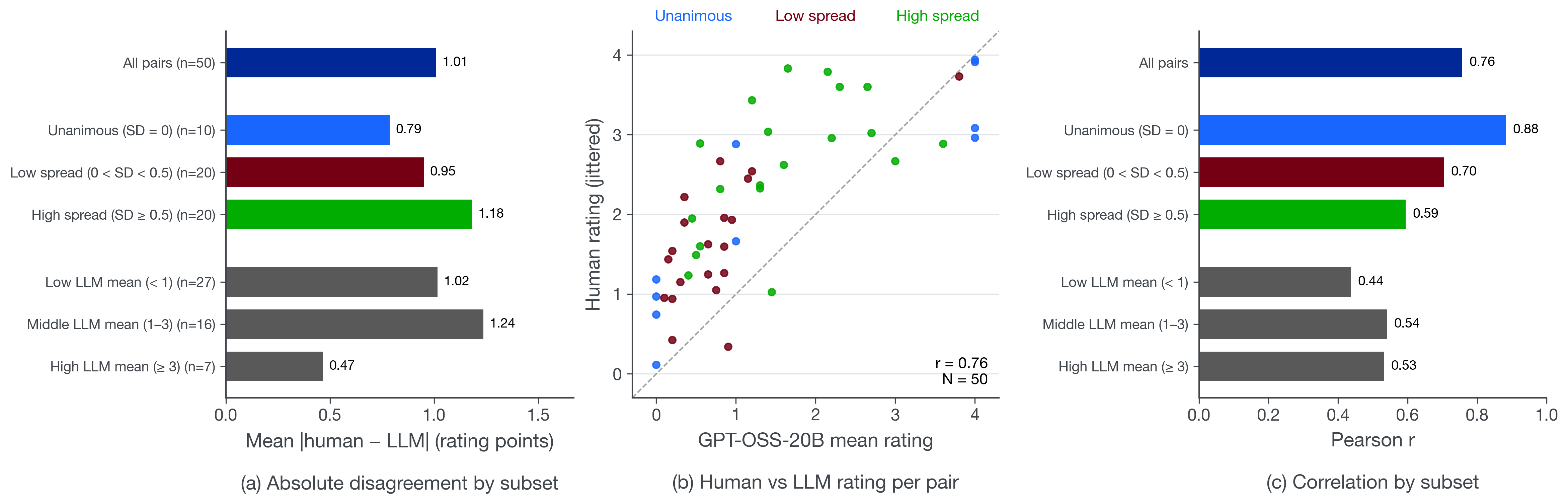}
  \caption{Human ratings versus GPT-OSS-20B relatedness scores for the 50 surveyed capability--ability pairs.
  Panel~(a) reports the mean absolute difference between the human mean and the LLM mean, by the spread of the 20 LLM ratings (colored) and by the level of the LLM mean (gray);
  panel~(b) plots each pair against the 45-degree line, colored by LLM spread, with human ratings jittered vertically to separate ties;
  panel~(c) reports the Pearson correlation between human and LLM means for the same subsets.}
  \label{fig:human-vs-llm}
\end{figure}

With seven raters the survey is indicative rather than inferential.
We find that humans rate pairs as more related for 82\% of pairs, so most pairs in Figure~\ref{fig:human-vs-llm}(b) sit above the 45-degree line.
The gap reverses only for the seven pairs where the LLM mean is at least 3, on which humans rate $0.47$ points lower.
As for the ordering, the correlation is high across all 50 pairs, at $r = 0.76$ and Spearman $\rho = 0.81$.
\citet{lyu2026right} find the same for LLM judges in occupational settings, where judges preserve ordering but differ from human input on absolute quantities.
Because ROE is standardized across occupations, a common level offset leaves the ranking largely unchanged.

\subsection{Robotics exposure versus other AI exposure indices}
\label{app:roe-other-ai}

We compare ROE to two previously published robot exposure scores.
\citet{webb2019impact} focuses on industrial robots and measures the overlap between the text of robot patents and O{*}NET task descriptions and aggregates it to occupations.
\citet{legateyang2026robots} follow a task-based approach similar to \citet{eloundou2024gpts}, where an LLM rates whether robots demonstrated today can perform each physical O{*}NET task, and task ratings are aggregated to occupations by time share.

For the joint comparison, we harmonize all measures once to six-digit 2018 SOC codes, following the Economic Innovation Group \href{https://github.com/EIG-Research/AI-unemployment}{replication repository}, which bundles the Felten, Eloundou, and Webb releases.
Webb's scores are reported for 1990 Census occupations, which we map to 2018 SOC through the crosswalks documented in that repository; the Anthropic index is taken from the occupation-level \href{https://huggingface.co/datasets/Anthropic/EconomicIndex/resolve/main/robot_exposure/robot_exposure_occupations.csv}{data release} of \citet{legateyang2026robots}, which uses the same O{*}NET-SOC 2019 codes as ROE.
The correlation matrix uses the occupations covered by all measures ($N=332$); Webb's 1990 occupation classification is the binding constraint.
Because the Anthropic index covers all 774 ROE occupations, we additionally compare ROE and the Anthropic index alone on this larger sample.

\begin{figure}[H]
  \centering
  \includegraphics[width=0.85\linewidth]{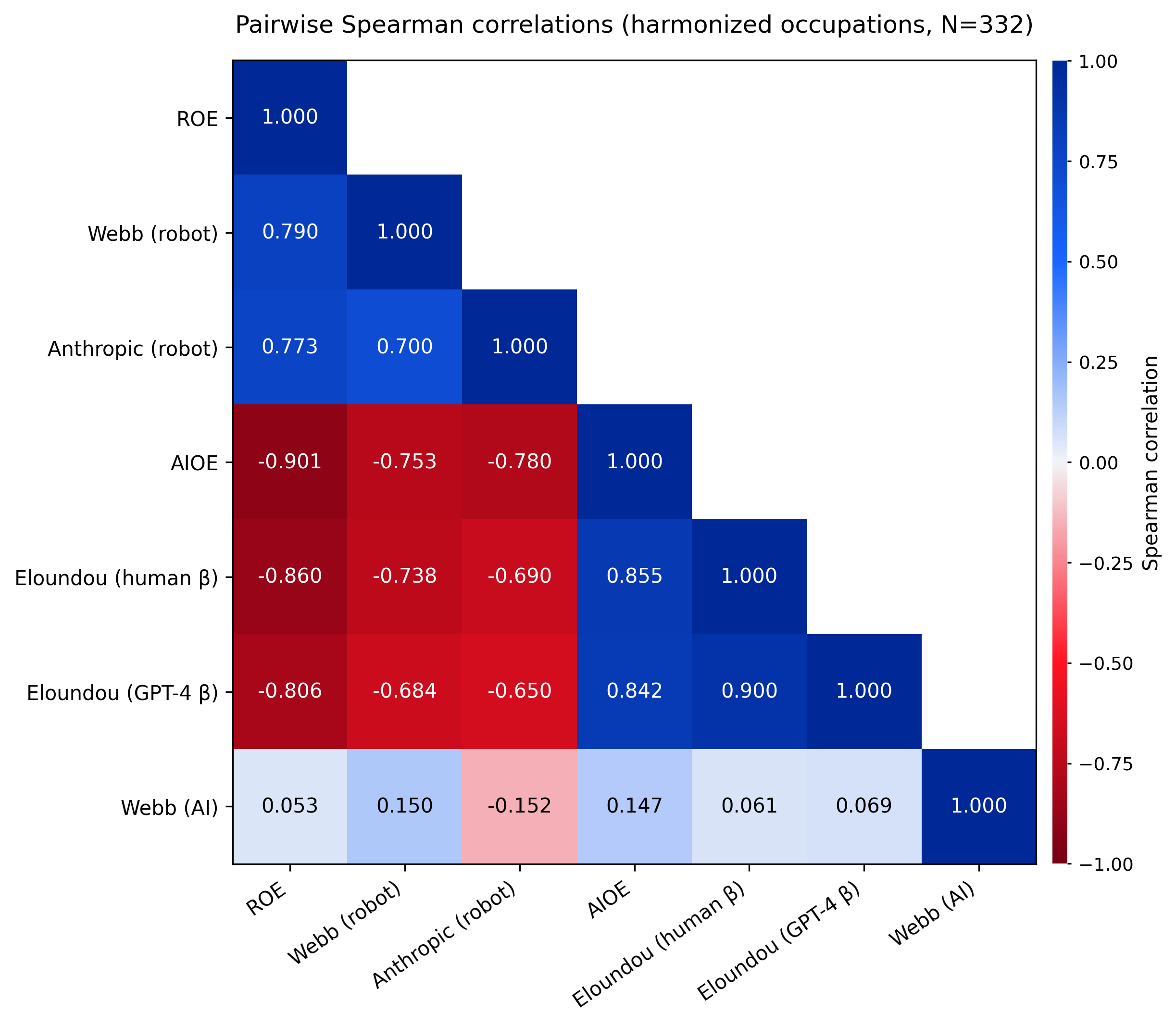}
  \caption{Spearman correlation matrix for harmonized occupation-level exposure ($N=332$ six-digit SOC codes).}
  \label{fig:exposure-correlation-matrix}
\end{figure}

\begin{figure}[H]
  \centering
  \begin{subfigure}[t]{0.48\textwidth}
    \centering
    \includegraphics[width=\linewidth]{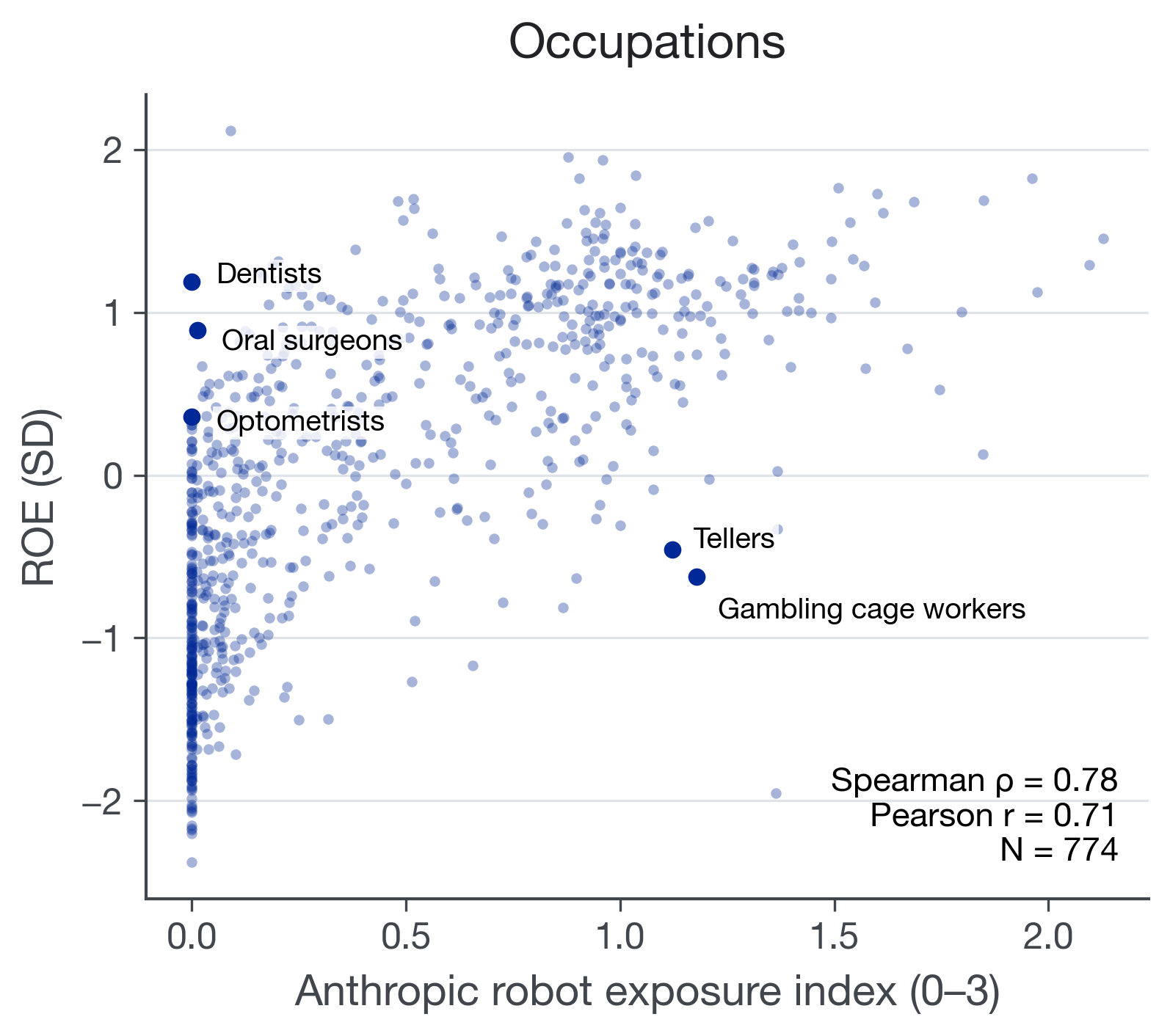}
    \caption{Occupations}
    \label{fig:roe-vs-anthropic-occ}
  \end{subfigure}\hfill
  \begin{subfigure}[t]{0.48\textwidth}
    \centering
    \includegraphics[width=\linewidth]{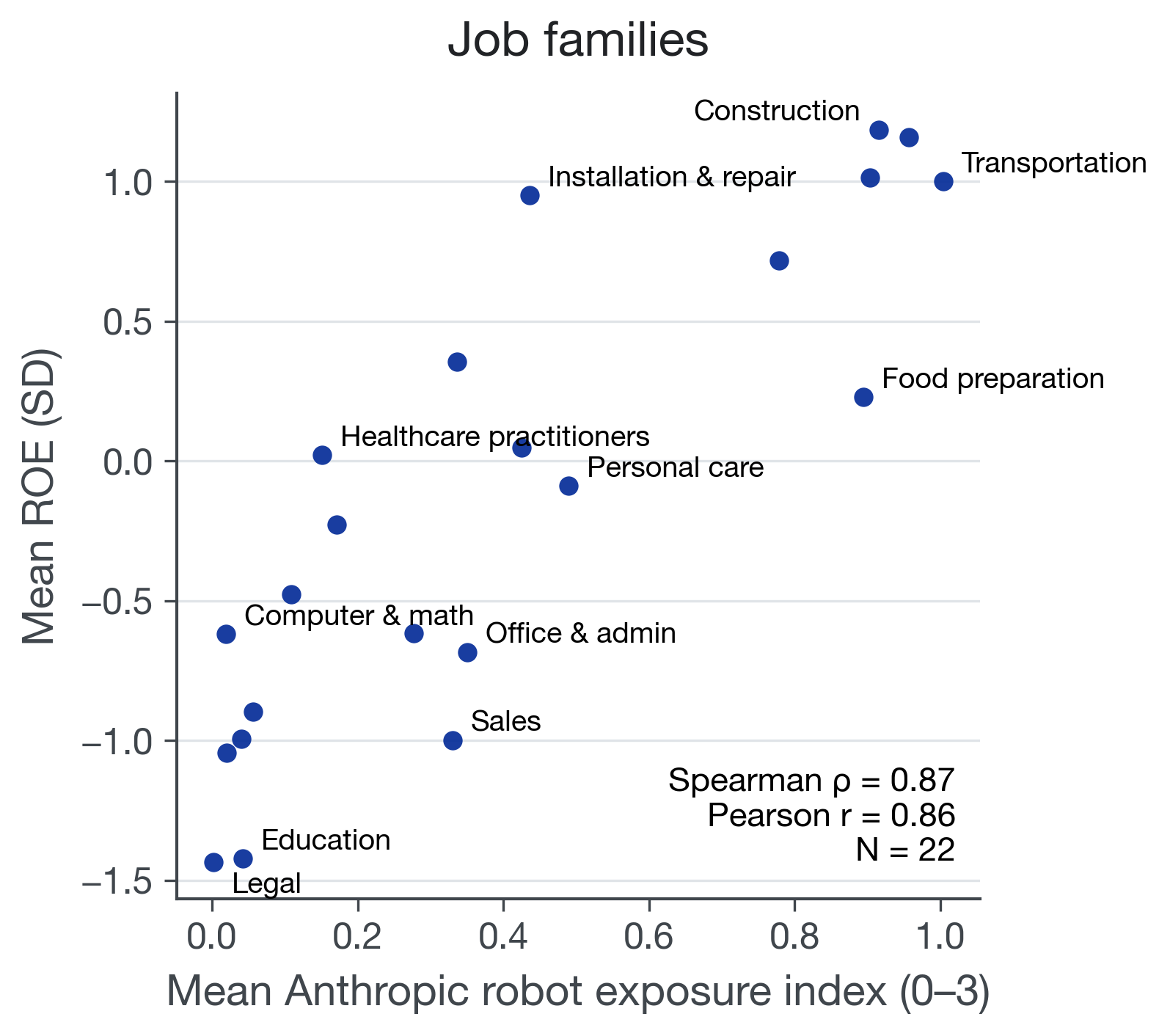}
    \caption{Job families}
    \label{fig:roe-vs-anthropic-family}
  \end{subfigure}\\[0.75em]
  \begin{subfigure}[t]{0.48\textwidth}
    \centering
    \includegraphics[width=\linewidth]{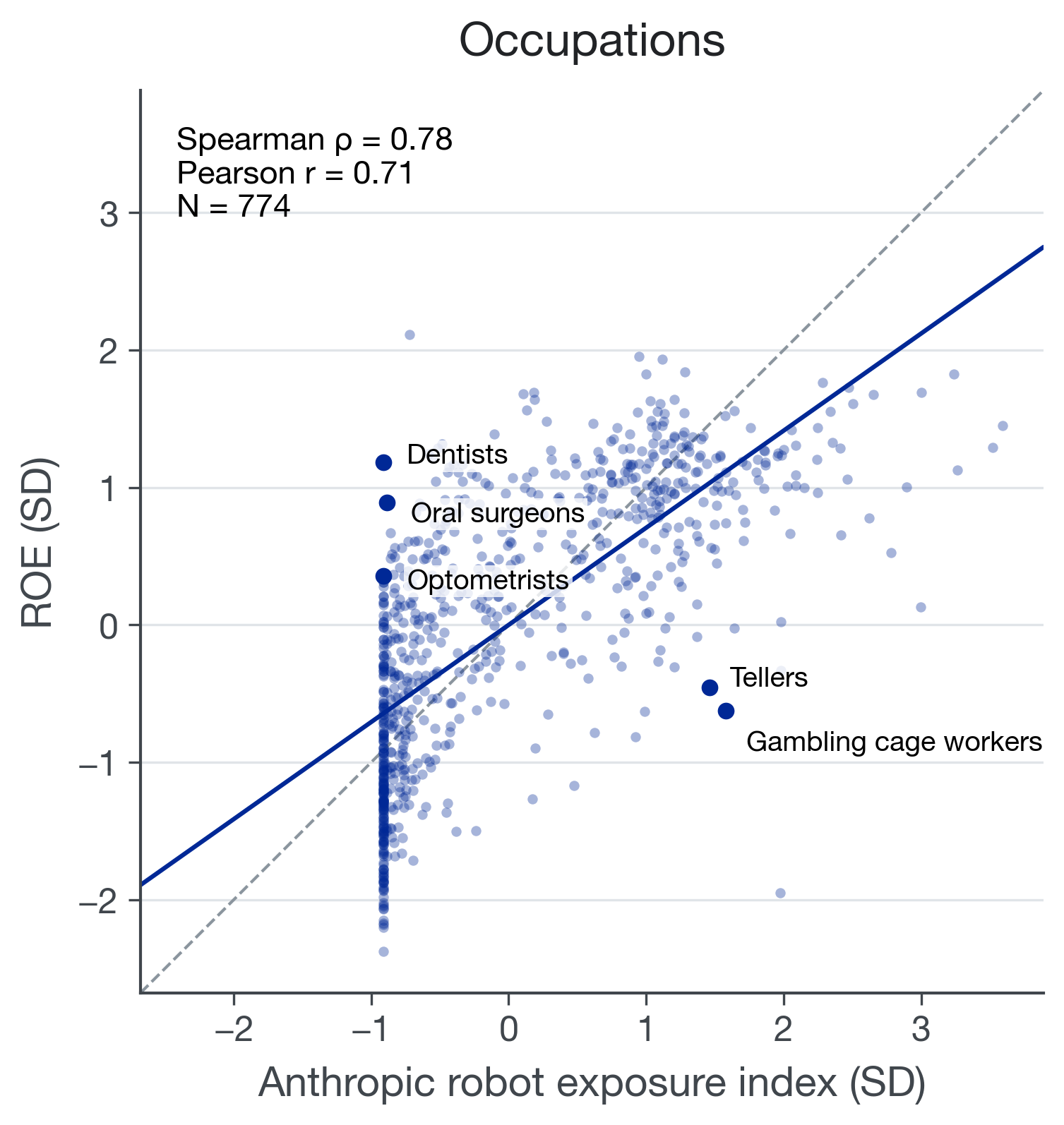}
    \caption{Occupations, both axes standardized}
    \label{fig:roe-vs-anthropic-occ-sd}
  \end{subfigure}\hfill
  \begin{subfigure}[t]{0.48\textwidth}
    \centering
    \includegraphics[width=\linewidth]{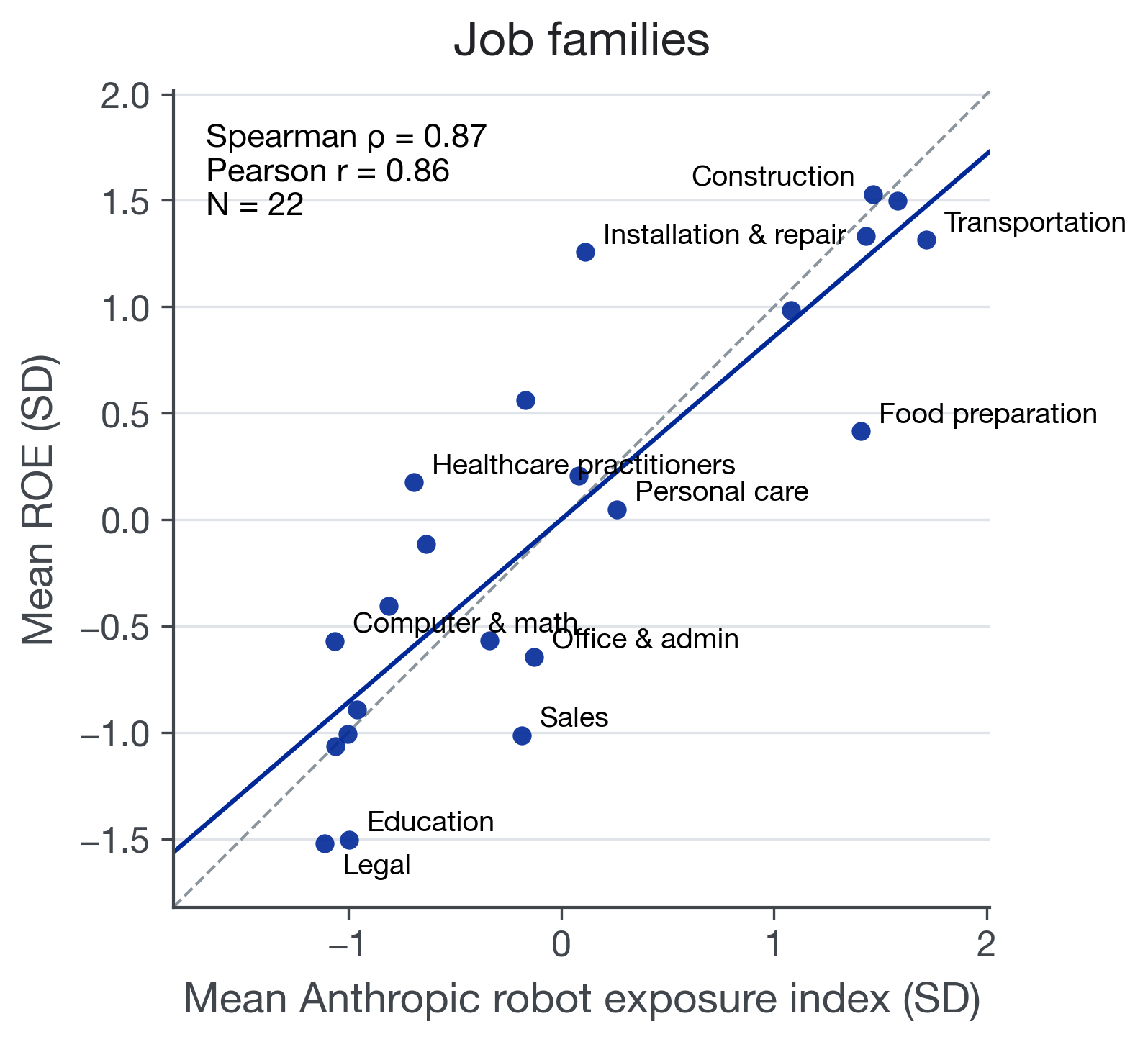}
    \caption{Job families, both axes standardized}
    \label{fig:roe-vs-anthropic-family-sd}
  \end{subfigure}
  \caption{ROE versus the Anthropic robot exposure index \citep{legateyang2026robots}.
  Panel~(a) plots standardized ROE against the Anthropic index (0--3) for all 774 ROE occupations; panel~(b) plots unweighted means across occupations for the 22 job families.
  Panels~(c) and~(d) standardize both measures across the plotted units; the solid line is the least-squares fit, whose slope equals the Pearson correlation, and the dashed line marks equality.}
  \label{fig:roe-vs-anthropic}
\end{figure}

ROE correlates strongly with both robot exposure scores: $\rho = 0.79$ with Webb and $\rho = 0.77$ with the Anthropic index, which correlate at $\rho = 0.70$ with each other (Figure~\ref{fig:exposure-correlation-matrix}).
On all 774 occupations, ROE and the Anthropic index correlate at $\rho = 0.78$ (Pearson $r = 0.71$), and at $\rho = 0.87$ across the 22 job families (Figure~\ref{fig:roe-vs-anthropic}).
The largest disagreements reflect the Anthropic index's focus on robots demonstrated today rather than capability relatedness: precision clinical occupations such as dentists, oral surgeons, and optometrists score high on ROE but near zero on the Anthropic index, whereas cash-handling occupations such as tellers and gambling cage workers score higher on the Anthropic index.

We also compare ROE to three AI exposure measures: AIOE \citep{felten2024generative}, the human- and GPT-4-labeled exposure scores of \citet{eloundou2024gpts}, and Webb's AI exposure, which applies his patent method to AI patents \citep{webb2019impact}.
ROE correlates negatively with the language-model measures (Figure~\ref{fig:exposure-correlation-matrix}).
Occupations most exposed to robotics are thus largely those least exposed to language models.

\subsection{Capability domains and exposure}
\label{app:domain-removal}

Section~\ref{sec:results-domains} removes one capability at a time and finds that no single capability carries the index.
Capabilities within a domain are close substitutes, so this understates what a whole area of robotics contributes: drop grasping and the three other manipulation capabilities still load on the same manual abilities.
Here we repeat the exercise for the eight top-level domains of Table~\ref{tab:capabilities}, removing all of a domain's capabilities together.

Domains contain between one and four capabilities, so under the published ROE a domain weighs more the more finely the taxonomy subdivides it.
We therefore report two aggregation rules.
The first sums the capabilities in a domain, which reproduces the published index.
The second averages them, so every domain counts the same however many capabilities it holds; the two indices correlate at Pearson $r = 0.94$ and Spearman $\rho = 0.95$ across occupations.
Table~\ref{tab:aggregation-example} works the two rules through for a single ability.

% Auto-generated by 06 Capability Decomposition/decompose_roe_by_domain.py
\begin{table}[H]
  \centering
  \small
  \caption{The two aggregation rules applied to one O{*}NET ability, Manual Dexterity.}
  \label{tab:aggregation-example}
  \begin{tabular}{@{}>{\raggedright\arraybackslash}p{0.34\linewidth}l rr@{}}
    \toprule
    Domain & $X_{ca}$ per capability & Summed & Averaged \\
    \midrule
    Manipulation & 1.00, 1.00, 1.00, 1.00 & 4.00 & 1.00 \\
    Human--Robot Interaction & 0.20, 0.97 & 1.18 & 0.59 \\
    Perception \& Sensing & 0.38, 0.17 & 0.55 & 0.28 \\
    System Maintenance \& Autonomy & 0.26, 0.17 & 0.44 & 0.22 \\
    Navigation \& Spatial Awareness & 0.03, 0.24 & 0.26 & 0.13 \\
    Locomotion \& Mobility & 0.12, 0.10 & 0.23 & 0.11 \\
    Multi-Agent \& Collective Behavior & 0.19 & 0.19 & 0.19 \\
    Cognition \& Planning & 0.16 & 0.16 & 0.16 \\
    \midrule
    Ability relatedness $R_a$ & & 7.00 & 2.67 \\
    \bottomrule
  \end{tabular}
\end{table}

Removing a domain moves the ranking much further than removing any one capability (Table~\ref{tab:domain-removal}).
Under the published index, manipulation moves the average occupation 11.6 percentile points, against 2.5 points for the largest single capability.
The effect is more than the sum of its parts: manipulation's four capabilities removed individually give 1.5, 1.2, 1.2 and 0.7 points, so the domain contributes 2.5 times what its capabilities contribute separately.
Equal weighting raises every domain and leaves the leading three in place, but narrows the gap at the top: manipulation and locomotion and mobility become comparable at 17.5 and 16.9 points.

% Auto-generated by 06 Capability Decomposition/decompose_roe_by_domain.py
\begin{table}[H]
  \centering
  \small
  \setlength{\tabcolsep}{5pt}
  \caption{Effect of removing each robotics capability domain from the exposure index. Removal drops all of the domain's capabilities and re-standardizes. The upper panel sums the capabilities in a domain, which reproduces the published ROE; the lower averages them, so every domain counts the same however many capabilities it holds. $|\Delta$ pct$|$ is the mean movement of an occupation in the exposure ranking, in percentile points. Churn is the share of the most exposed 10\%, 25\% and 50\% of occupations no longer in that group after removal.}
  \label{tab:domain-removal}
  \begin{tabular}{@{}>{\raggedright\arraybackslash}p{0.36\linewidth}cccc@{}}
    \toprule
    Domain & $|\Delta$ pct$|$ & 10\% & 25\% & 50\% \\
    \midrule
  \multicolumn{5}{@{}l}{\textit{Capabilities summed within a domain (published ROE)}} \\
  Manipulation & 11.6 & 36\% & 27\% & 16\% \\
  Locomotion \& Mobility & 3.8 & 24\% & 11\% & 5\% \\
  Human--Robot Interaction & 2.9 & 14\% & 6\% & 5\% \\
  Navigation \& Spatial Awareness & 2.1 & 9\% & 3\% & 3\% \\
  System Maintenance \& Autonomy & 2.0 & 9\% & 2\% & 3\% \\
  Cognition \& Planning & 1.5 & 8\% & 2\% & 2\% \\
  Perception \& Sensing & 1.3 & 6\% & 2\% & 2\% \\
  Multi-Agent \& Collective Behavior & 0.6 & 3\% & 1\% & 1\% \\
  \addlinespace
  \multicolumn{5}{@{}l}{\textit{Capabilities averaged within a domain}} \\
  Manipulation & 17.5 & 55\% & 48\% & 26\% \\
  Locomotion \& Mobility & 16.9 & 60\% & 51\% & 24\% \\
  Human--Robot Interaction & 8.9 & 32\% & 23\% & 12\% \\
  Navigation \& Spatial Awareness & 4.1 & 13\% & 11\% & 7\% \\
  System Maintenance \& Autonomy & 4.6 & 14\% & 12\% & 8\% \\
  Cognition \& Planning & 6.6 & 23\% & 15\% & 10\% \\
  Perception \& Sensing & 2.1 & 8\% & 7\% & 3\% \\
  Multi-Agent \& Collective Behavior & 2.1 & 8\% & 5\% & 4\% \\
    \bottomrule
  \end{tabular}
\end{table}

\subsection{ROE with and without the capability taxonomy}
\label{app:taxonomy-sensitivity}

ROE has two inputs: the relatedness score of each O{*}NET ability, which comes from the capability taxonomy and LLM scoring (Equation~\ref{eq:ability-exposure}), and the importance and level of each ability in each occupation, which come from O{*}NET.
To assess the contribution of the taxonomy, we construct an O{*}NET-only version of ROE.
It applies the same aggregation (Equations~\ref{eq:roe-raw} and~\ref{eq:roe-std}) but replaces the relatedness scores with a simple classification: each of the 19 physical and psychomotor abilities receives a score of one, and all other abilities a score of zero.
Both indices are standardized across the 774 occupations.

\begin{figure}[H]
  \centering
  \includegraphics[width=0.65\linewidth]{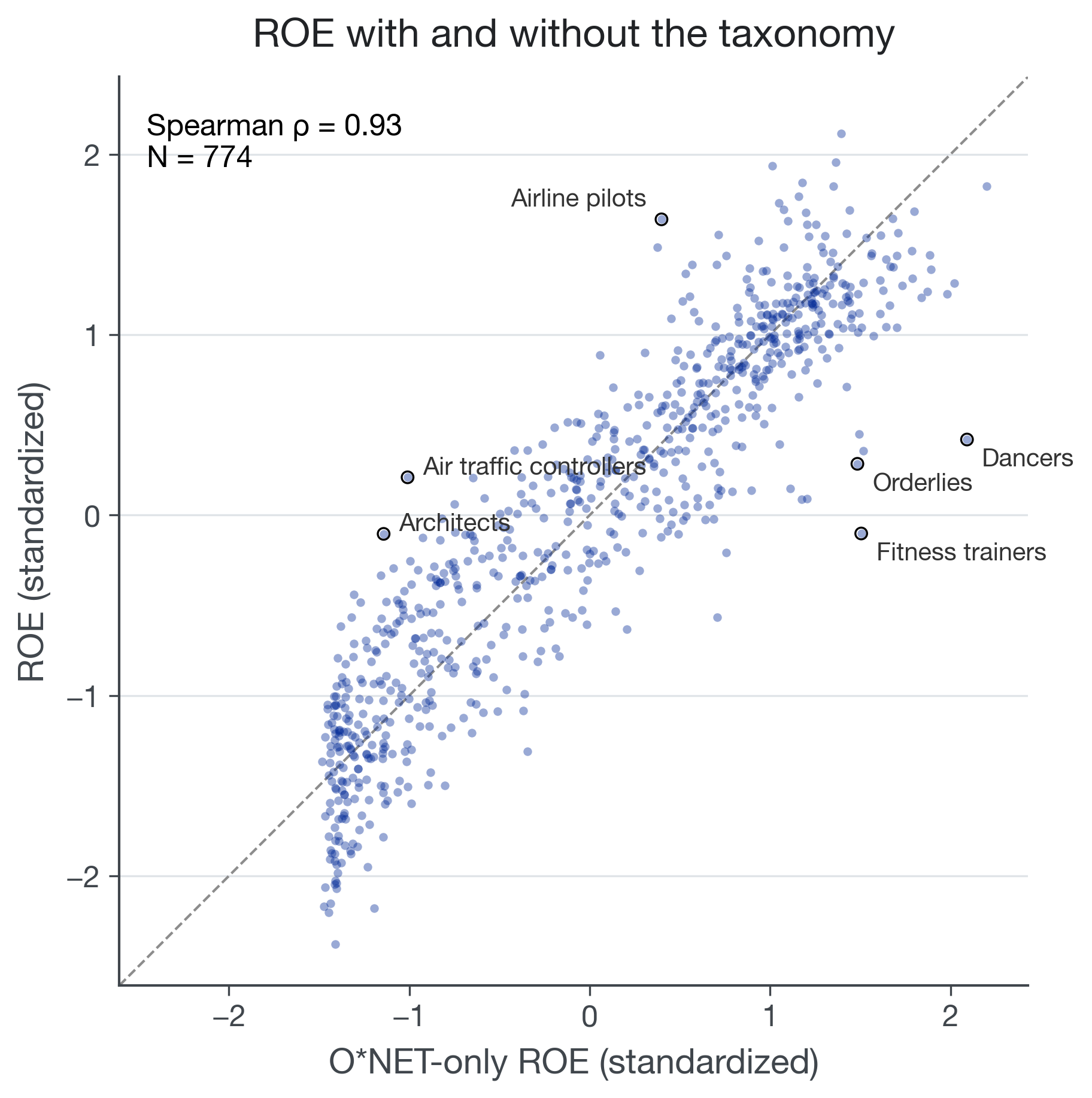}
  \caption{ROE versus an O{*}NET-only version of ROE in which the 19 physical and psychomotor abilities receive a relatedness score of one and all other abilities zero ($N=774$; both standardized). The dashed line marks equality; labeled occupations are among those that deviate most.}
  \label{fig:roe-onet-only}
\end{figure}

Figure~\ref{fig:roe-onet-only} compares the two indices; their Spearman correlation is 0.93.
The occupation ranking is therefore largely determined by the O{*}NET ability profiles, which primarily separate physically from cognitively demanding occupations.
For the same reason, ROE is insensitive to individual capabilities: removing any one of the 16 capabilities from the taxonomy changes the ranking only marginally ($\rho \geq 0.993$).

The two indices differ where the taxonomy departs from the physical--cognitive classification.
Spatial Orientation, a cognitive ability in O{*}NET, receives the highest relatedness score and Depth Perception, a sensory ability, the third highest, whereas Stamina, Trunk Strength, and Explosive Strength rank near the bottom (Table~\ref{tab:onet-abilities}).
As a result, airline pilots and air traffic controllers, whose work relies on spatial orientation and depth perception, score higher on ROE than on the O{*}NET-only index, while dancers and fitness trainers, whose work relies on strength and stamina, score lower.

\subsection{Additional figures}
\label{app:additional-figures}

\begin{figure}[H]
  \centering
  \begin{subfigure}[t]{0.55\textwidth}
    \centering
    \includegraphics[width=\linewidth]{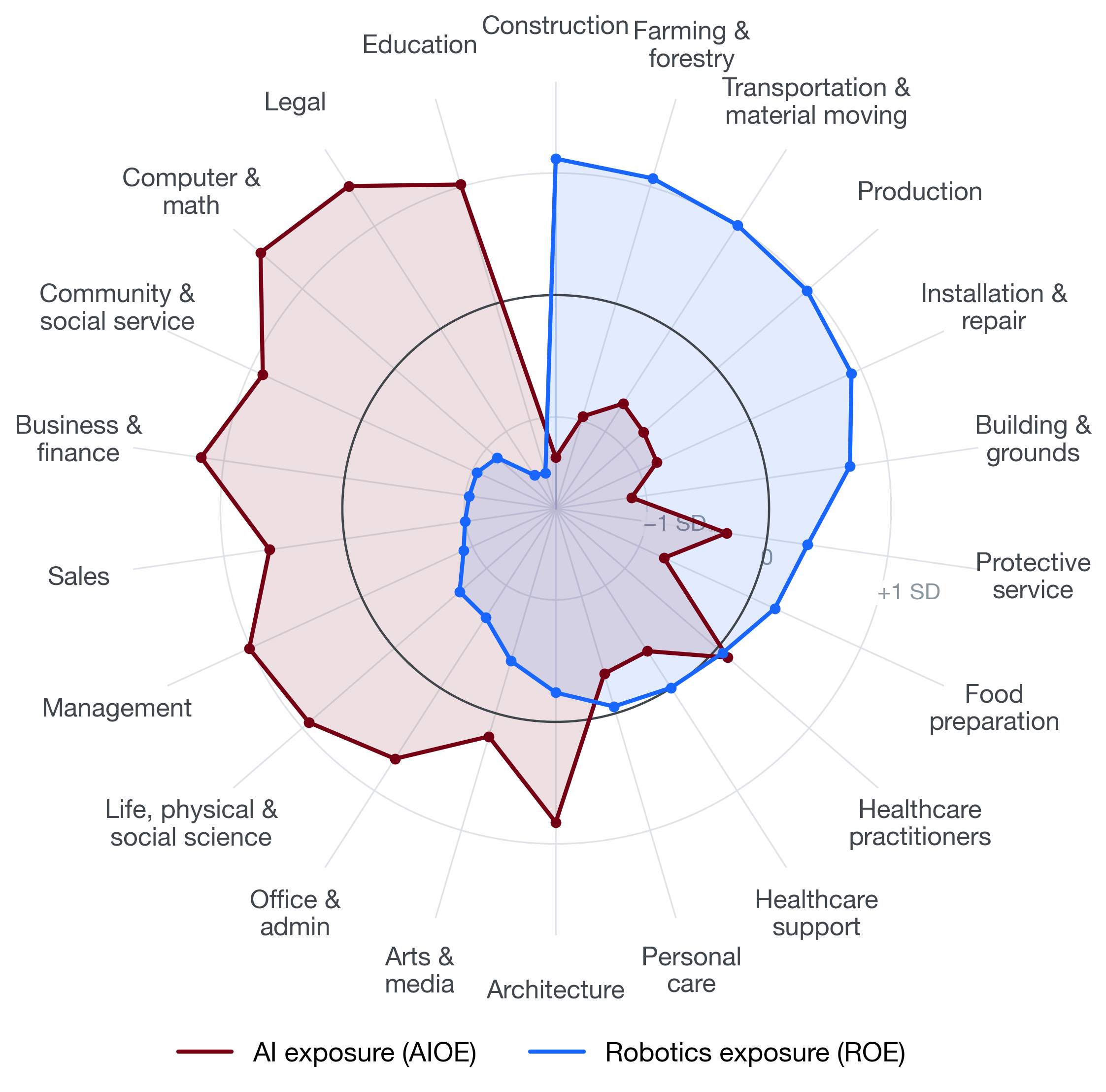}
    \caption{Unweighted across occupations}
    \label{fig:roe-aioe-job-family-radar-unw}
  \end{subfigure}\\[0.75em]
  \begin{subfigure}[t]{0.55\textwidth}
    \centering
    \includegraphics[width=\linewidth]{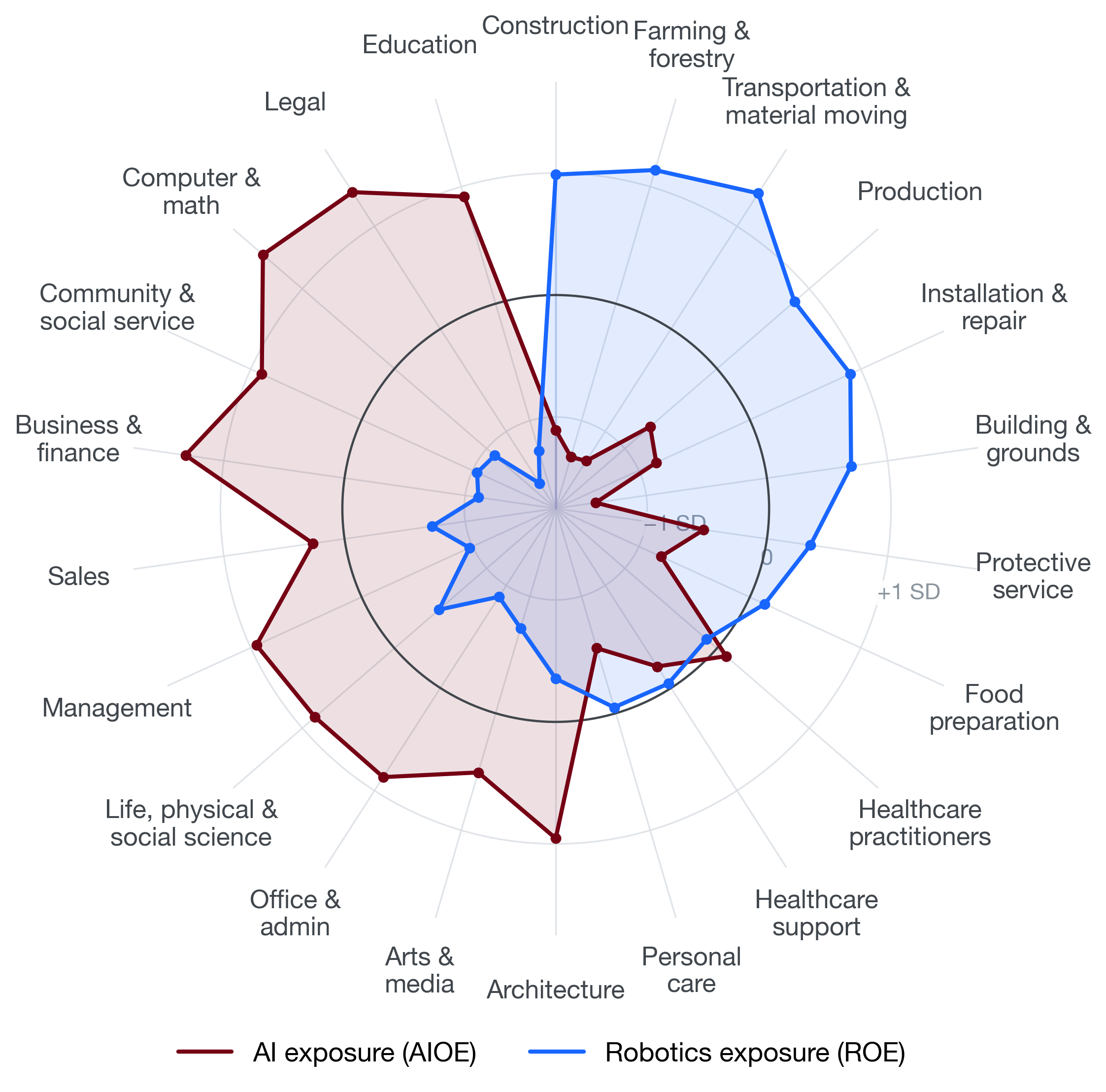}
    \caption{Employment-weighted}
    \label{fig:roe-aioe-job-family-radar-emp}
  \end{subfigure}
  \caption{Robotics (ROE) and AI (AIOE) exposure by job family, radar view of Figure~\ref{fig:roe-aioe-job-family-bars}.
  Each spoke is a job family, ordered clockwise by unweighted mean ROE; the radius is mean standardized exposure, and the dark ring marks zero.}
  \label{fig:roe-aioe-job-family-radar}
\end{figure}

\begin{figure}[H]
  \centering
  \includegraphics[width=0.86\linewidth]{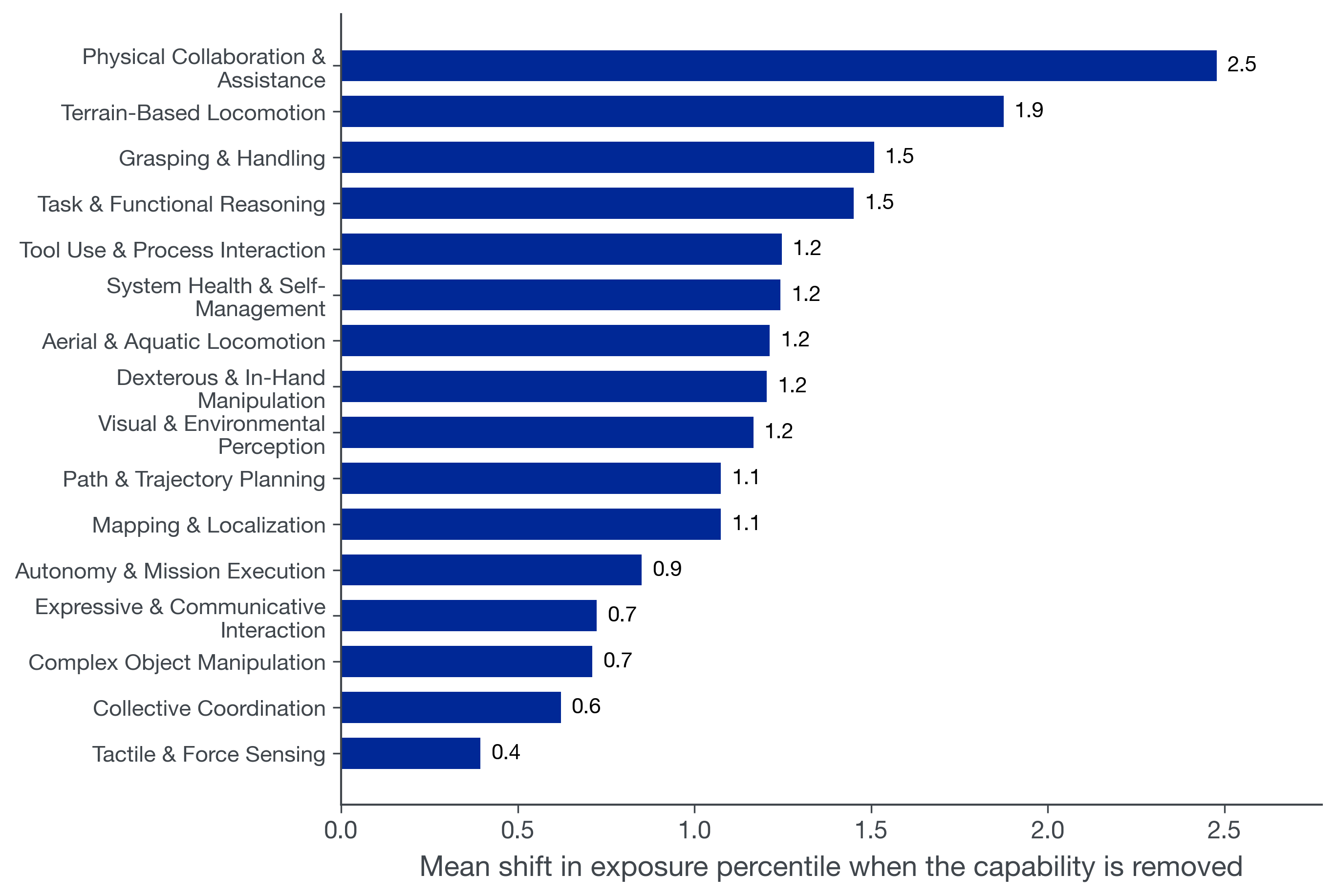}
  \caption{Effect of removing each robotics capability from the exposure index, across 774 occupations, as reported in Table~\ref{tab:capability-removal}.
  Each bar drops the capability from the relatedness matrix, re-standardizes, and shows the mean absolute change in an occupation's exposure percentile.}
  \label{fig:roe-capability-removal}
\end{figure}

\end{document}